\documentclass[%
 aip,
 amsmath,amssymb,
 reprint,%
]{revtex4-1}

\usepackage{graphicx}
\usepackage{dcolumn}
\usepackage{bm}

\usepackage[utf8]{inputenc}
\usepackage[T1]{fontenc}
\usepackage{mathptmx}
\usepackage{etoolbox}

\usepackage[font=small,singlelinecheck=true,justification=raggedright]{caption}
\usepackage{subcaption}
\usepackage{multirow}
\usepackage{hyperref}
\usepackage[version=4]{mhchem}
\usepackage{pifont}
\usepackage{algorithm}
\usepackage{algpseudocode}
\usepackage{setspace}
\usepackage[dvipsnames]{xcolor}
\newcommand{\cmark}{\ding{51}}
\newcommand{\xmark}{\ding{55}}
\newcommand{\gcheck}{{\color{ForestGreen}\cmark}}
\newcommand{\rcross}{{\color{red}\xmark}}

\makeatletter
\def\@email#1#2{%
 \endgroup
 \patchcmd{\titleblock@produce}
  {\frontmatter@RRAPformat}
  {\frontmatter@RRAPformat{\produce@RRAP{*#1\href{mailto:#2}{#2}}}\frontmatter@RRAPformat}
  {}{}
}%
\makeatother
\begin{document}

\preprint{AIP/123-QED}

\title[Implicit Neural Representations for Chemical Reaction Paths]{Implicit Neural Representations for Chemical Reaction Paths}

\author{Kalyan Ramakrishnan}
 \email{kalyanr@robots.ox.ac.uk}
\affiliation{
University of Oxford, Oxford, United Kingdom
}
\author{Lars L. Schaaf}
\affiliation{ 
Engineering Laboratory, University of Cambridge, Cambridge CB2 1PZ, United Kingdom
}
\affiliation{ 
Cavendish Laboratory, University of Cambridge, Cambridge CB3 0HE, United Kingdom
}
\author{Chen Lin}
\affiliation{
University of Oxford, Oxford, United Kingdom
}
\author{Guangrun Wang}
\affiliation{
University of Oxford, Oxford, United Kingdom
}
\author{Philip Torr}
\homepage{https://torrvision.com/}
\affiliation{
University of Oxford, Oxford, United Kingdom
}


\begin{abstract}
We show that neural networks can be optimized to represent minimum energy paths as continuous functions, offering a flexible alternative to discrete path-search methods such as Nudged Elastic Band (NEB). Our approach parameterizes reaction paths with a network trained on a loss function that discards tangential energy gradients and enables instant estimation of the transition state. We first validate the method on two-dimensional potentials and then demonstrate its advantages over NEB on challenging atomistic systems where (i) poor initial guesses yield unphysical paths, (ii) multiple competing paths exist, or (iii) the reaction follows a complex multi-step mechanism. Results highlight the versatility of the method: for instance, a simple adjustment to the sampling strategy during optimization can help escape local-minimum solutions. Finally, in a low-dimensional setting, we demonstrate that a single neural network can learn from existing paths and generalize to unseen systems, showing promise for a universal reaction path representation.
\end{abstract}

\maketitle

\noindent Journal of Chemical Physics. DOI: \href{https://doi.org/10.1063/5.0267023}{10.1063/5.0267023}.
\vspace{-8pt}

\section{Introduction}
\label{sec:intro}
Understanding the mechanisms of chemical reactions and predicting their rates are central goals in computational chemistry, with wide-ranging applications in catalysis, biochemistry, and materials science\cite{wales,jensen}. A key concept is the minimum energy path (MEP), which traces a valley on the potential energy surface, connecting given stable configurations $A$ and $B$. The MEP approximates the most probable route between the reactants and products of a chemical reaction\cite{irc}. The highest-energy configuration along this path, the transition state (TS), defines the energy barrier of the reaction and largely determines its kinetics via transition state theory\cite{eyring,evans}. Consequently, identifying MEPs and transition states is an essential step toward guiding the design of semiconductors\cite{ge_si}, catalysts\cite{solid,electronic,dft_surface, rhodes202517o}, and drugs\cite{enzyme}.

A well-established computational technique for MEP search is the Nudged Elastic Band (NEB)\cite{neb}, which optimizes a discrete chain of configurations (called images) between the endpoints $A$ and $B$. The images are connected by springs and repeatedly \emph{nudged} in directions perpendicular to the chain to minimize energy and converge on the MEP. A subsequent modification, known as climbing-image NEB\cite{cl-neb}, additionally drives the highest-energy image \emph{up} the energy surface to estimate the TS simultaneously. While these methods remain widely adopted, their reliance on a discrete representation of the underlying continuous reaction path can lead to difficulties\cite{imptan,dneb}. NEB can produce unphysical configurations in the absence of a good initial guess\cite{gsm,idpp}, can converge to suboptimal solutions during optimization\cite{neb_paths}, and may fail to capture complex reaction paths\cite{dneb}.

Inspired by the recent success of Implicit Neural Representations (INRs)\cite{nerf,siren,deepsdf,occupancy} in modeling continuous functions, we parameterize reaction paths with a neural network representing a smooth function of a reaction coordinate. An INR is a neural network that parameterizes an implicitly defined function\cite{siren}, in our case, the MEP, defined by the constraint that the energy gradient remains aligned with the path tangent. The continuous formulation naturally supports automatic differentiation\cite{autodiff,pytorch}, enabling precise computation of path tangents, curvatures, and parameter gradients while enjoying the expressivity of neural networks. This allows us to incorporate physical priors into the training algorithm, such as the nudging and climbing mechanisms from NEB, without requiring a discrete representation. It also allows flexibility in the loss function and sampling strategy for optimization. 

We demonstrate the capabilities of our implicit neural representation for MEP and TS search through experiments on systems ranging from simple two-dimensional potentials to challenging material and molecular reactions for which the conventional approach of NEB fails. Results show that the INR reliably locates transition states in these settings, overcoming the limitations of NEB described earlier. Finally, we show that, unlike existing methods that require a fresh optimization for each new system, an INR can learn from existing paths to instantly approximate MEPs for unseen systems, opening the door for a universal reaction path representation.

\section{Related Work}
\label{sec:related}
We are not the first to utilize continuous functions for MEP or TS search. Below, we summarize some relevant work.

\paragraph{Discrete representations.}
Several methods\cite{string,gsm,improved_string,gmam} optimize a discrete representation of the reaction path but repeatedly fit a continuous curve to estimate tangents or redistribute points along the path. Ref.~\onlinecite{gsm} fits cubic splines and approximates the TS via interpolation. Ref.~\onlinecite{improved_string} also fits cubic splines but runs a separate optimization for TS search once an approximate MEP is found and does not discard tangential energy gradients. Ref.~\onlinecite{gmam} fits piecewise-linear curves within a variational formulation but does not focus on TS search. In contrast to these methods, we directly update a continuous neural network representation of the reaction path, combining MEP and TS search in a single optimization.

\paragraph{Continuous representations.}
Other methods\cite{ssm,curve,vrpo,doob} explicitly work with a continuous representation of the reaction path. Ref.~\onlinecite{ssm} represents the path with a cubic spline but only optimizes it for TS search and consequentially the path itself has no physical meaning. Ref.~\onlinecite{curve} updates a B-spline curve representation and approximates the TS with the highest-energy sample observed on the path. While the curve may be fit to an improved initial guess\cite{idpp}, the method has not been compared against NEB and has primarily demonstrated success on systems requiring very few optimization steps. Ref.~\onlinecite{vrpo} adopts a variational framework and represents the path with a linear combination of basis functions. Despite involving a non-linear, constrained optimization, the method does not, in practice, find MEPs for even two-dimensional potentials\cite{vrpo}, limiting its applicability for path-search in atomistic settings. 
Refs.~\onlinecite{doob,raja2025action} use neural networks to model a distribution over stochastic trajectories connecting the end states rather than optimizing for MEP or TS search.

To our knowledge, we are the first to use neural networks to represent MEPs and estimate transition states. Unlike prior work, we combine the nudging and climbing mechanisms in the same optimization. Moreover, all previous methods lack the ability to condition the path representation on arbitrary end states for potential generalization across systems.

\section{Method}
\label{sec:method}
\subsection{Overview}
For an $N$-atom system, let $\vec{A}\in\mathbb{R}^{3N}$ and $\vec{B}\in\mathbb{R}^{3N}$ represent the coordinates of all atoms in the initial and final states, respectively. We assume that the reaction path $\vec{x}(t)\in\mathbb{R}^{3N}$, with $t\in[0, 1]$, takes the form
\begin{equation}
    \vec{x}_\theta(t) = \vec{b}(t) + t(1-t)\ \vec{g}_\theta(t),
\end{equation}
where $\vec{b}(t)$ is a \emph{base} path and $\vec{g}_\theta: [0, 1] \rightarrow \mathbb{R}^{3N}$ is a neural network with learnable parameters $\theta$. Note that $\vec{x}_\theta(0) = \vec{A}$ and $\vec{x}_\theta(1) = \vec{B}$ are satisfied by construction as long as $\vec{b}_\theta(0) = \vec{A}$ and $\vec{b}_\theta(1) = \vec{B}$. 
Typically, we set the base path as the linear interpolation from $\vec{A}$ to $\vec{B}$,
\begin{equation}
\vec{b}(t) = (1-t)\vec{A} + t\vec{B}.
\end{equation}
In general, given $n+1$ points $\{\vec{x}_0, \vec{x}_1, \cdots, \vec{x}_{n}\}$ sampled from a reference path, we may set $\vec{b}(t)$ as the Lagrange interpolating polynomial,
\begin{equation}
\vec{b}(t) = \sum_{j=0}^{n} \left[ \prod_{k\neq j}^n \left(\frac{t-t_k}{t_j-t_k}\right) \right] \vec{x}_j,
\end{equation}
with $t_j = j/n$ for $j\in\{0, 1, \cdots, n\}$.

In either case, the path $\vec{x}_\theta(t)$ is differentiable with respect to both $t$ and $\theta$, allowing us to compute exact tangents, curvatures, and parameter gradients. For instance, we can differentiate through the potential energy $U(\vec{x})$ of the system as follows, assuming access to its gradient $\nabla_x U(\vec{x})$:
\begin{equation}
    \frac{\partial U(\vec{x}_\theta(t))}{\partial\theta} = \nabla_x U(\vec{x}_\theta(t))^T \frac{\partial\vec{x}_\theta(t)}{\partial\theta}.
\end{equation}

One objective is to tune the parameters of the network to find the minimum energy path (MEP), which follows a valley on the potential energy surface, connecting the local minima $\vec{A}$ and $\vec{B}$. This requires that the energy gradient $\nabla_x U(\vec{x}_\theta(t))$ has no component orthogonal to the path tangent $\vec{x}_\theta^{'}(t) = {d\vec{x}_\theta(t)}/{dt}$ over $t\in[0, 1]$. More importantly, we are interested in the transition state (TS), the highest-energy point on the MEP from $\vec{A}$ to $\vec{B}$. The energy at this point (relative to the starting point) is the barrier of the reaction. We approach this problem by sampling points along the path and minimizing a loss function over the parameters $\theta$, as described next.

\begin{figure}[t]
  \centering
  \begin{subfigure}[b]{0.487\linewidth}
    \centering
    \includegraphics[width=\linewidth]{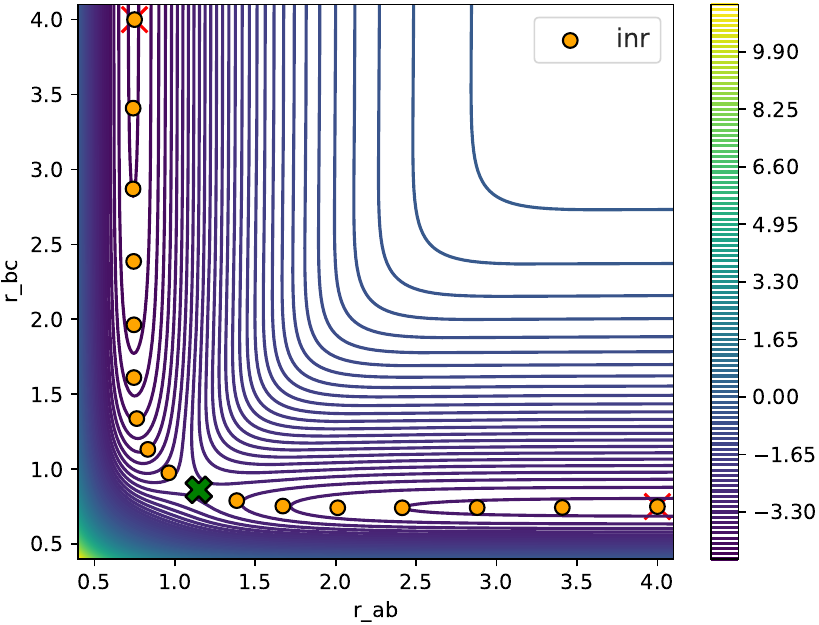}
    \vspace{-15pt}
    \caption{LEPS}
    \label{fig:leps_inr}
  \end{subfigure}
  \begin{subfigure}[b]{0.498\linewidth}
    \centering
    \includegraphics[width=\linewidth]{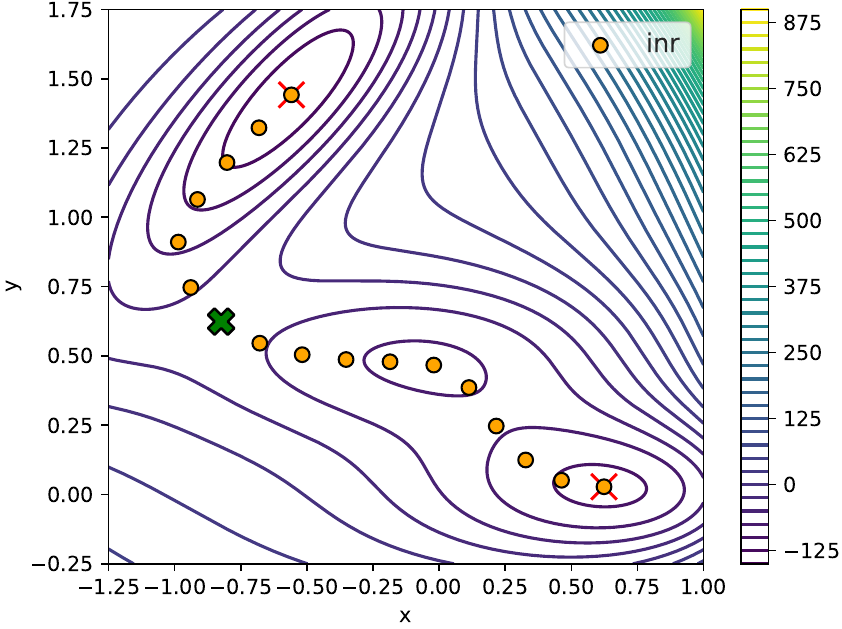}
    \vspace{-15pt}
    \caption{MB}
    \label{fig:mb_inr}
  \end{subfigure}
  \\[0.5mm]
  \begin{subfigure}[b]{0.49\linewidth}
    \centering
    \includegraphics[width=\linewidth]{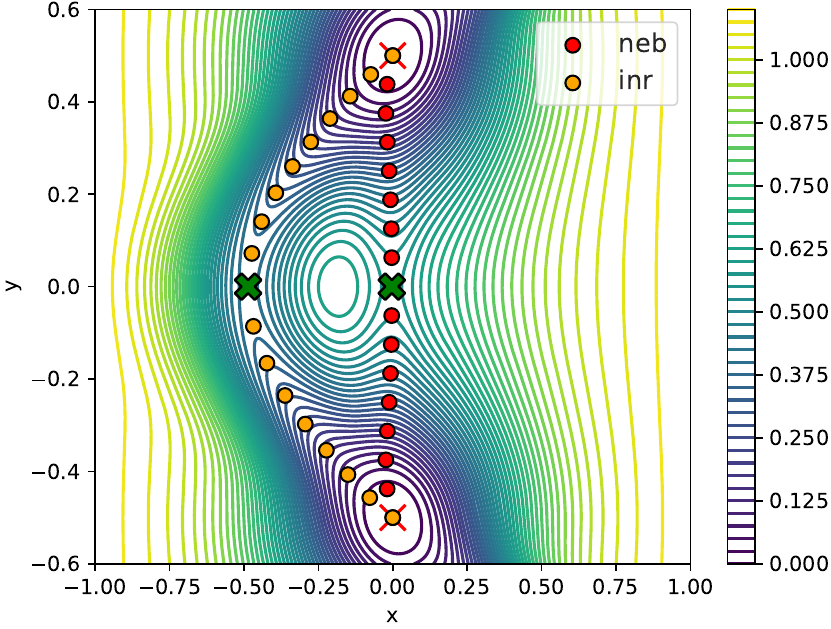}
    \vspace{-15pt}
    \caption{Sine-1}
    \label{fig:larsx1_neb+inr}
  \end{subfigure}
  \begin{subfigure}[b]{0.498\linewidth}
    \centering
    \includegraphics[width=\linewidth]{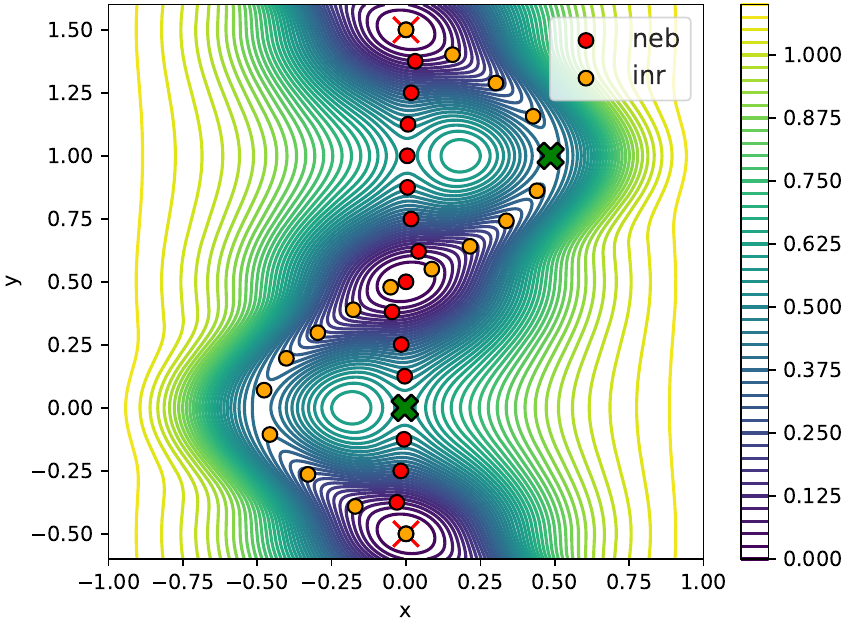}
    \vspace{-15pt}
    \caption{Sine-2 (INR-GS)}
    \label{fig:larsx2_neb+inr}
  \end{subfigure}
  \vspace{-20pt}
  \caption{INR paths (orange) for four 2D potentials (a--d). Panels c and d show incorrect, higher-barrier paths (red) predicted by NEB.}
  \label{fig:inr_paths_2d}
\end{figure}

\subsection{Loss function}
\label{subsec:loss}
A simple choice for the loss function is the expected energy along the path with an additional term to encourage a uniform \emph{speed}, e.g.,
\begin{equation}
\label{eq:naive}
    L(\theta) = \mathbb{E}_t\left[U(\vec{x}_\theta(t))\right] + \lambda_s\text{Var}_t \left[\|\vec{x}_\theta^{'}(t)\|\right],
\end{equation}
where $\mathbb{E}_t$ and $\text{Var}_t$ are the expectation and variance over some chosen distributions $q_u(t)$ and $q_s(t)$, respectively, and $\lambda_s>0$ is a hyperparameter. A similar loss function was adopted previously\cite{curve} to optimize B-spline curves, using the uniform distribution $u(0, 1)$ for both. However, since the gradient $\nabla_\theta L(\theta)$ also includes components of $\nabla_x U(\vec{x}_\theta(t))$ \emph{tangential} to the path, this loss encourages sampled points to depart from higher-energy regions on it, relying entirely on the second term to reduce this effect (see Appendix~\ref{app:sec:hyper} for details).

Our loss function is inspired by the Nudged Elastic Band (NEB)\cite{neb}, which optimizes a discrete set of points by nudging each point along the projection of $-\nabla_x U(\vec{x})$ \emph{orthogonal} to the (estimated) path tangent. Specifically, we remove the tangential components using the \emph{stop gradient} operator, which behaves as an identity function that has a zero gradient during automatic differentiation. We define the modified energy,
\begin{equation}
\label{eq:stop}
    \tilde U(\vec{x}_\theta(t)) := U(\vec{x}_\theta(t)) - \text{Stop}[\nabla_x U(\vec{x}_\theta(t))_{\parallel}]^T \vec{x}_\theta(t),
\end{equation}
where
\begin{equation}
\nabla_x U(\vec{x}_\theta(t))_{\parallel} = \left(\frac{\nabla_x U(\vec{x}_\theta(t))^T \vec{x}_\theta^{'}(t)}{\|\vec{x}_\theta^{'}(t)\|^2}\right) \vec{x}_\theta^{'}(t)
\end{equation}
is the projection of $\nabla_x U(\vec{x}_\theta(t))$ tangential to the path. Using $\tilde U(\vec{x}_\theta(t))$ in Eq.~\ref{eq:naive} gives the updated loss function,
\begin{equation}
\label{eq:updated}
    \tilde L(\theta) = \mathbb{E}_t\left[\tilde U(\vec{x}_\theta(t))\right] + \lambda_s\text{Var}_t \left[\|\vec{x}_\theta^{'}(t)\|\right].
\end{equation}
As intended, the loss gradient $\nabla_\theta \tilde L(\theta)$ now discards components of $\nabla_x U(\vec{x}_\theta(t))$ tangential to the path since
\begin{align}
\label{eq:discard}
\nabla_x \tilde U(\vec{x}_\theta(t)) &= \nabla_x U(\vec{x}_\theta(t)) - \nabla_x U(\vec{x}_\theta(t))_{\parallel}\nonumber\\
&= \nabla_x U(\vec{x}_\theta(t))_{\perp}
\end{align}
is the projection of $\nabla_x U(\vec{x}_\theta(t))$ orthogonal to the path\textcolor{black}{, producing nudging updates during optimization in our continuous setting. We theoretically justify the use of the modified energy $\tilde U$ (vs.~$U$) in Appendix~\ref{app:sec:theory}.}

The stop gradient is a standard feature in popular deep learning libraries\cite{pytorch}, making it straightforward to use in a loss function for neural network training. Additional loss terms to exploit prior knowledge about the reaction or explicitly promote alignment with the energy gradient may further help, but they are beyond the scope of this paper.

\textcolor{black}{
Another advantage of our formulation is that it does not evaluate the energy Hessian, $H_{pq} = \frac{\partial^2 U(\vec{x})}{\partial x_p \partial x_q}$, which is in general more expensive than the gradient by a factor proportional to the system size \cite{analytic1,analytic2,nohess,hess}.
}

\vspace{-18pt}
\textcolor{black}{
\subsection{Improving TS localization}
\label{subsec:climb}
While the loss $\tilde L(\theta)$ can be numerically minimized (by sampling from $q_u$ and $q_s$) to estimate the MEP, locating the TS would still require a continuous search along the resulting path. Moreover, such an optimization would only passively improve the TS estimate (via improving the path). To avoid this, analogous to the climbing-image NEB\cite{cl-neb}, we modify the loss so that at each update, the highest-energy sample is pushed \emph{up} the energy surface, tangential to the path. Specifically, given samples $\{t_i\}$ from $q_u$, if $t_* = \mathop{\arg\max}_{t_i} U(\vec{x}_\theta(t_i))$, our climbing loss is
\begin{equation}
\label{eq:cl}
    \tilde L_\text{Cl}(\theta) = \tilde L(\theta; \lambda_s) - \lambda_\text{Cl}\ \text{Stop}[\nabla_x U(\vec{x}_\theta(t_*))_{\parallel}]^T \vec{x}_\theta(t_*),
\end{equation}
where $\lambda_\text{Cl}>0$ is a hyperparameter. The intuition for this choice is that $\nabla_\theta \tilde L_\text{Cl}(\theta)$ essentially removes $\nabla_x U(\vec{x}_\theta(t_*))_{\parallel}$ twice, first zeroing it (see Eq.~\ref{eq:discard}) and then flipping its direction (due to the second term in Eq.~\ref{eq:cl}), thereby producing a climbing effect. After numerically minimizing $\tilde L_\text{Cl}(\theta)$, we locate the TS at $\vec{x}_\theta(t_*)$, obtaining both the MEP and TS from a single optimization.
}

\vspace{-30pt}
\textcolor{black}{
\subsection{Sampling strategies}
\label{subsec:sampling}
Typically, we let $q_u = q_s = u(0, 1)$, and compute the mean and variance in Eq.~\ref{eq:updated} with $n+1$ equidistant (``uniform'') samples $t_i = i/n$, $i\in\{0, 1, \cdots, n\}$. While this is a simple default, more complex sampling strategies may better suit the system or potential at hand, such as dynamically adapting $q_u$ or varying the number of samples used.
}

\textcolor{black}{
We focus on one such strategy in this paper inspired by the growing string method (GSM)\cite{gsm}, which, conceptually, optimizes a path by gradually growing it from the endpoints ($\vec{A}, \vec{B}$) until the two ends merge. The idea is to update the path starting from where the solution is known/easier to find (near the ends) and gradually expand toward the center. In practice, GSM maintains a discrete, increasing set of points at each end and optimizes both sets iteratively with the following steps: (i) nudging the points (as in NEB), (ii) fitting a cubic spline, (iii) adding new points when appropriate, and (iv) repositioning points along the spline for a uniform spacing within each end. In contrast, our approach provides a simple and elegant implementation of this concept by just modifying the sampling strategy for Eq.~\ref{eq:updated}. Specifically, for the energy loss (first term), rather than sampling uniformly from $t \in [0, 1]$, we could sample from the \emph{ends} by setting
\begin{equation}
    q_u(t) = \begin{cases}
        1/\alpha_k & \text{if}\ \ t \in [0, \frac{\alpha_k}{2}] \cup [1 - \frac{\alpha_k}{2}, 1]\\
        0 & \text{if}\ \ t \in (\frac{\alpha_k}{2}, 1 - \frac{\alpha_k}{2}),
    \end{cases}
\end{equation}
where $\alpha_k \in (0, 1]$ is gradually increased with each iteration $k$. The simplest choice is to linearly expand the sampled region toward the center with $\alpha_k = \frac{k}{N}$, where $N$ is the maximum number of iterations over which the loss function is optimized. Empirically, we find that growing sampling can improve the optimization trajectory to avoid local minimum paths, as we show in Sec.~\ref{sec:exp}. Other sampling strategies are beyond the scope of this paper.
}

\vspace{\baselineskip}
\noindent
\textcolor{black}{
We summarize our overall training procedure in Algorithm~\ref{alg:inr} and provide Python code on GitHub\footnote{See \url{https://github.com/Kalyan0821/INR-Path} for the Python code of our method.}.
}

\begin{figure*}[t]
\begin{minipage}{0.9\linewidth}
\vspace{-12pt}
\begin{algorithm}[H]
\textcolor{black}{
\caption{Training procedure for MEP and TS search.}
\vspace{-5pt}
\label{alg:inr}
\begin{spacing}{1.2}
\begin{algorithmic}[1]
\For{$k = 1$ to $n_\text{iters}$}
    \If{\emph{uniform sampling}}
        \State $\{t_i\} \gets \text{Uniform}\left([0, 1],\ n_\text{samples}\right)$ \Comment{$n_\text{samples}$ equidistant inputs in $[0, 1]$}
    \ElsIf{\emph{growing sampling}}
        \State $f_k \gets k / n_\text{iters}$
        \State $\{t^{(l)}_i\} \gets \text{Uniform}\left([0, f_k/2],\ n_\text{samples} / 2\right)$
        \State $\{t^{(r)}_i\} \gets \text{Uniform}\left([1-f_k/2, 1],\ n_\text{samples} / 2\right)$
        \State $\{t_i\} \gets \{t^{(l)}_i\} \cup \{t^{(r)}_i\}$ \Comment{$n_\text{samples}$ inputs from the ends}
    \EndIf
    \State $\{\vec{x}_\theta (t_i)\},\ \{\vec{x}\ '_\theta (t_i)\} \gets \text{INR}_\theta \{t_i\}$
    \State $\{U(\vec{x}_\theta (t_i))\},\ \{\nabla_x U(\vec{x}_\theta (t_i))\} \gets \text{Potential} \{\vec{x}_\theta (t_i)\}$ 
    \State $\{\tilde{U}(\vec{x}_\theta (t_i))\} \gets \{\ U(\vec{x}_\theta (t_i)) - \text{Stop} [\nabla_x U(\vec{x}_\theta(t_i))_{\parallel}]^T \vec{x}_\theta(t_i)\ \}$ \Comment{modified energies}
    \State $t_* \gets \mathop{\arg\max}_{t_i} U(\vec{x}_\theta(t_i))$
    \State $\text{Loss}(\theta) \gets \text{mean}_i\{\tilde{U}(\vec{x}_\theta (t_i))\} + \lambda_s\ \text{var}_i \{\|\vec{x}\ '_\theta (t_i)\|\} - \lambda_\text{Cl}\ \text{Stop}[\nabla_x U(\vec{x}_\theta(t_*))_{\parallel}]^T \vec{x}_\theta(t_*)$
    \State Update $\theta$ based on $\frac{\partial \text{Loss}(\theta)}{\partial \theta}$ 
\EndFor
\State $\text{MEP} \gets \vec{x}_\theta (\cdot)$
\State $\text{TS} \gets \vec{x}_\theta(t_*)$
\end{algorithmic}
\end{spacing}
}
\end{algorithm}
\end{minipage}
\end{figure*}

\section{Experiments}
\label{sec:exp}
In this section, we first validate our method (INR) on two-dimensional systems and then focus on challenging high-dimensional, atomistic systems for which the traditional approach of NEB fails. We show that our method performs reliably in such settings.

\subsection{Implementation details}
\label{subsec:implementation}
\vspace{-8pt}
We use the Atomic Simulation Library (ASE)\cite{ase} for the climbing NEBs, with the Fast Inertial Relaxation Engine (FIRE)\cite{fire} optimizer, $15$ intermediate ($17$ total) images, and spring constant $k=0.1$. We consistently observed faster convergence with FIRE than with other ASE optimizers. For the INRs, we use a three-layer feedforward neural network having $256$ hidden features with the $\tanh$ activation and a linear output layer. To be consistent with the NEBs, we sample $17$ inputs for training and inference, although a different number may be used for inference. By default, we set $\lambda_s = 0$ and $\lambda_\text{Cl} = 1.0$ for uniform sampling (see Appendix~\ref{app:sec:hyper} for details). For growing sampling (Sec.~\ref{subsec:sampling}), we use $\lambda_s = 0.1$ (and $\lambda_\text{Cl} = 0$) to maintain a balanced distribution of points throughout as the sampling region expands from the ends. The networks are trained with the Adam optimizer\cite{adam} and a default learning rate of $1e{-3}$. We use linear interpolation as the INR base path and NEB initialization and optimize both methods for a maximum of $500$ iterations. The end states are first relaxed to equilibrium using L-BFGS\cite{lbfgs} for atomistic systems. The potential energy function is a variant of MACE\cite{mace} specific to the system type, enabling fast evaluation of energies and gradients. To assess the practical cost of locating the TS, we wait until the root mean square loss gradient reaches $\text{RMS}\left(\nabla_\theta L(\theta)\right) < 1e{-3}$ (for uniform sampling). We then tightly refine the TS estimate $\vec{x}(t_*)$ using Sella\cite{sella}, a saddle point optimizer, until the largest per-atom $\nabla_x U(\vec{x}(t_*))$ norm reaches $F_{\max} < 5e{-4}$, for a maximum of $500$ iterations. For growing sampling, we always train the network up to a pre-defined iteration count (e.g., $200$) since the sampled region expands according to this number. On average, refinement needed $<3\%$ of the method's energy evaluations to converge.

\begin{figure}[t]
  \centering
  \begin{subfigure}[b]{0.43\linewidth}
    \centering
    \includegraphics[width=\linewidth]{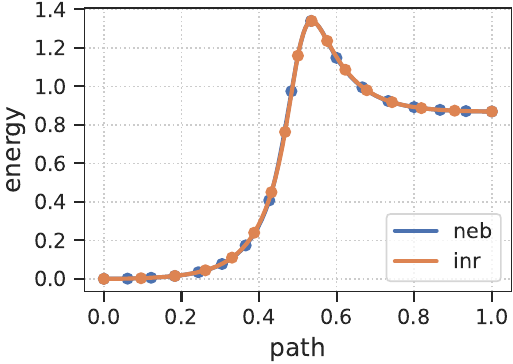}
    \vspace{-15pt}
    \caption{LEPS}
    \label{fig:leps_E}
  \end{subfigure}
  \begin{subfigure}[b]{0.425\linewidth}
    \centering
    \includegraphics[width=\linewidth]{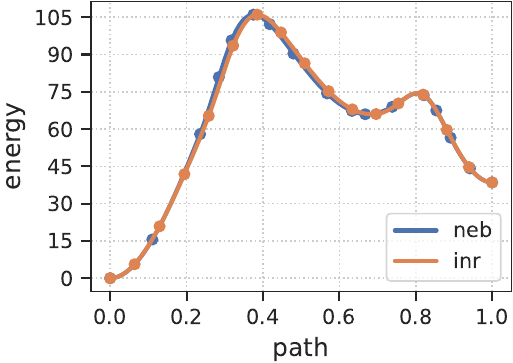}
    \vspace{-15pt}
    \caption{MB}
    \label{fig:mb_E}
  \end{subfigure}
  \\[0.5mm]
  \begin{subfigure}[b]{0.43\linewidth}
    \centering
    \includegraphics[width=\linewidth]{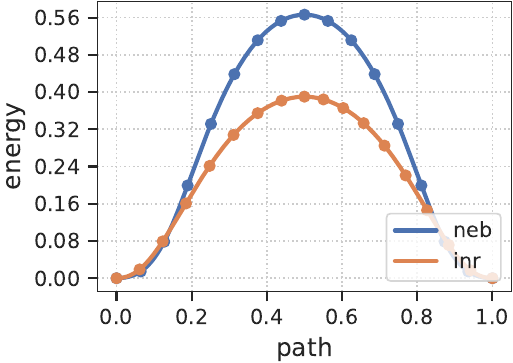}
    \vspace{-15pt}
    \caption{Sine-1}
    \label{fig:larsx1_E}
  \end{subfigure}
  \begin{subfigure}[b]{0.425\linewidth}
    \centering
    \includegraphics[width=\linewidth]{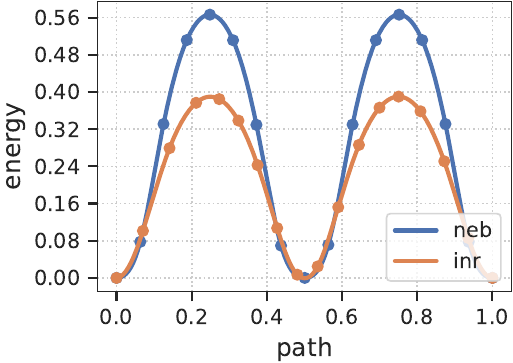}
    \vspace{-15pt}
    \caption{Sine-2 (INR-GS)}
    \label{fig:larsx2_E}
  \end{subfigure}
  \vspace{-8pt}
  \caption{Energy profiles for the 2D potentials. NEB (blue) and INR (orange) find identical paths for standard systems (a--b). NEB gets stuck in higher-barrier solutions for some other systems (c--d).}
  \label{fig:inr_profiles_2d}
\vspace{-10pt}
\end{figure}

\subsection{Two-dimensional systems}
\label{subsec:2d}
\paragraph{\textbf{Standard potentials.}}
Figs.~\ref{fig:leps_inr} and \ref{fig:leps_E} show the path produced by the INR and the energy profiles of both methods for the standard London--Eyring--Polanyi--Sato (LEPS)\cite{leps,neb} potential. Both methods converge to the same path and correct energy barrier of $1.34$. In Fig.~\ref{fig:leps_progress} (Appendix~\ref{app:sec:2d}), we plot the TS energy error $|U(\vec{x}(t_*)) - U(\vec{x}_\text{TS})|$ vs.~iterations, showing that the methods approach the TS at a similar rate. We also show results for the popular M\"uller--Brown (MB)\cite{mb,ssm,gsm} potential in Figs.~\ref{fig:mb_inr} and \ref{fig:mb_E}, where once again the INR matches the NEB path. The exact forms of the potentials, along with the initial and final minima, are provided in Appendix~\ref{app:sec:2d}.

\paragraph{\textbf{Global optimization.}}
We design a new potential energy surface to evaluate TS search methods on systems having multiple competing paths but only one with the lowest barrier. This surface tests the ability of methods to escape local minimum energy paths in favor of the global solution. The surface is periodic in the $y$ direction, with adjacent pairs of minima connected by two MEPs: (i) the nearly straight-line path connecting the minima and (ii) a curved path with a smaller barrier. The potential takes the following form:
\begin{align}
U_\text{Sine}(x, y) = &\left(1 - e^{-6x^2 - \cos^2(\pi y)}\right) \times \nonumber\\
& \left(1 - 0.1e^{-100x^2} - 0.6e^{-20(x+0.5\cos(\pi y))^2}\right) + 0.1x^2.
\end{align}
The local minima are at $\left\{\left(0, y_n\right) \mid y_n = n - \frac{1}{2},  n\in\mathbb{Z}\right\}$, and the period along the $y$ direction is $2$.

Figs.~\ref{fig:larsx1_neb+inr} and \ref{fig:larsx1_E} show NEB and INR results for the endpoints $(0, y_0)\rightarrow (0, y_1)$. We call this system sine-1. While NEB gets stuck in the direct path with a barrier of $0.566$, the INR identifies the curved path with a smaller barrier of $0.39$. For the more complex sine-2 system, having endpoints $(0, y_0)\rightarrow (0, y_2)$, neither method finds the curved path. However, by applying the growing sampling idea from Sec.~\ref{subsec:sampling}, the INR (named INR-GS) finds the lower-barrier curved path, as shown in Figs.~\ref{fig:larsx2_neb+inr} and \ref{fig:larsx2_E}. We also present the evolution of this path over iterations and results for the sine-3 system in Appendix~\ref{app:sec:larsxn}, where similar behavior is observed. 

\subsection{\textbf{\ce{In2O3}}}
\begin{figure}[h]
\vspace{-4pt}
  \centering
  \begin{subfigure}[b]{0.22\linewidth}
    \centering
    \includegraphics[width=\linewidth]{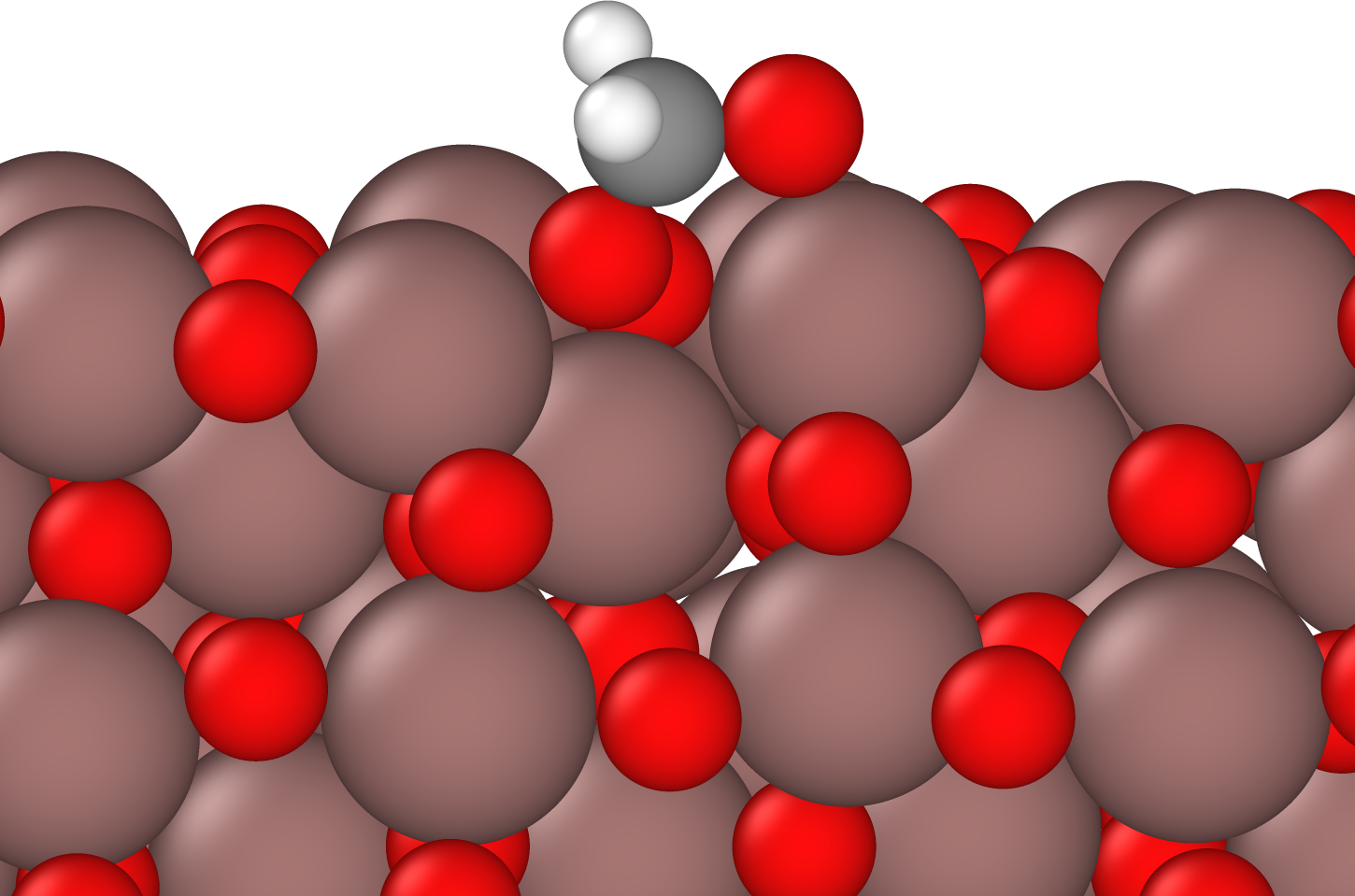}
    \vspace{-12pt}
    \caption{Initial}
    \label{fig:lars_a}
  \end{subfigure}\hspace{0.001\linewidth}
  \begin{subfigure}[b]{0.22\linewidth}
    \centering
    \includegraphics[width=\linewidth]{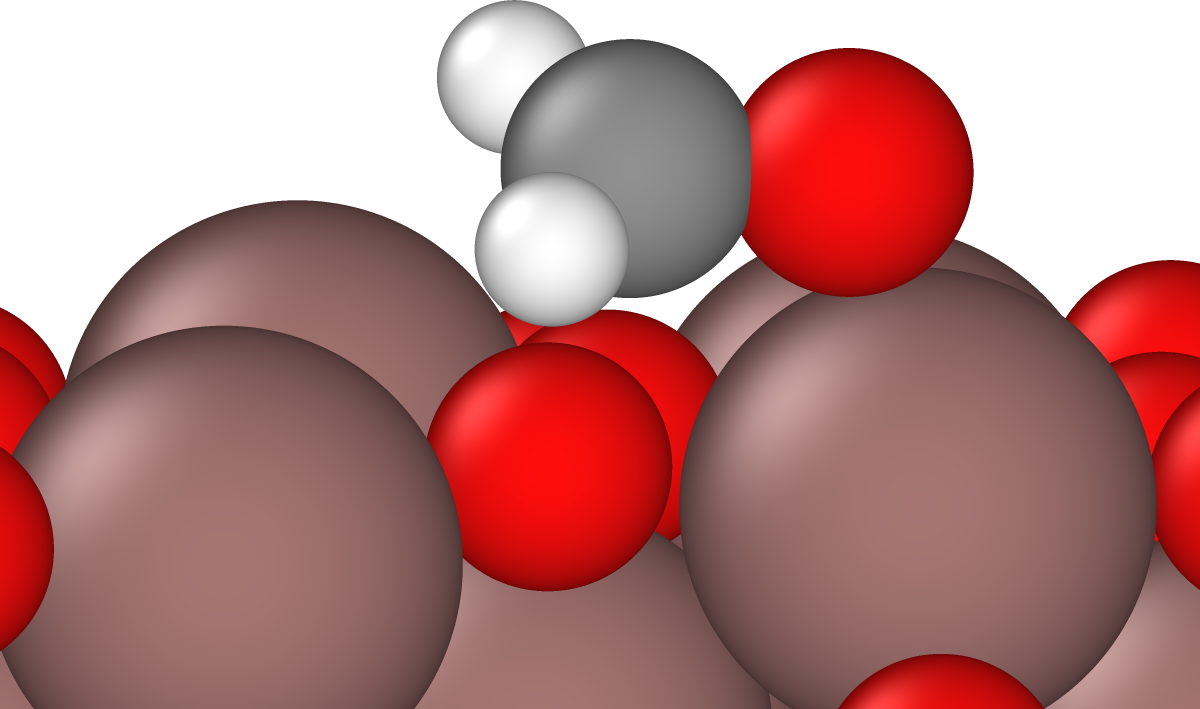}
    \vspace{-12pt}
    \caption{TS}
    \label{fig:lars_sella_ts}
  \end{subfigure}\hspace{0.001\linewidth}
  \begin{subfigure}[b]{0.22\linewidth}
    \centering
    \includegraphics[width=\linewidth]{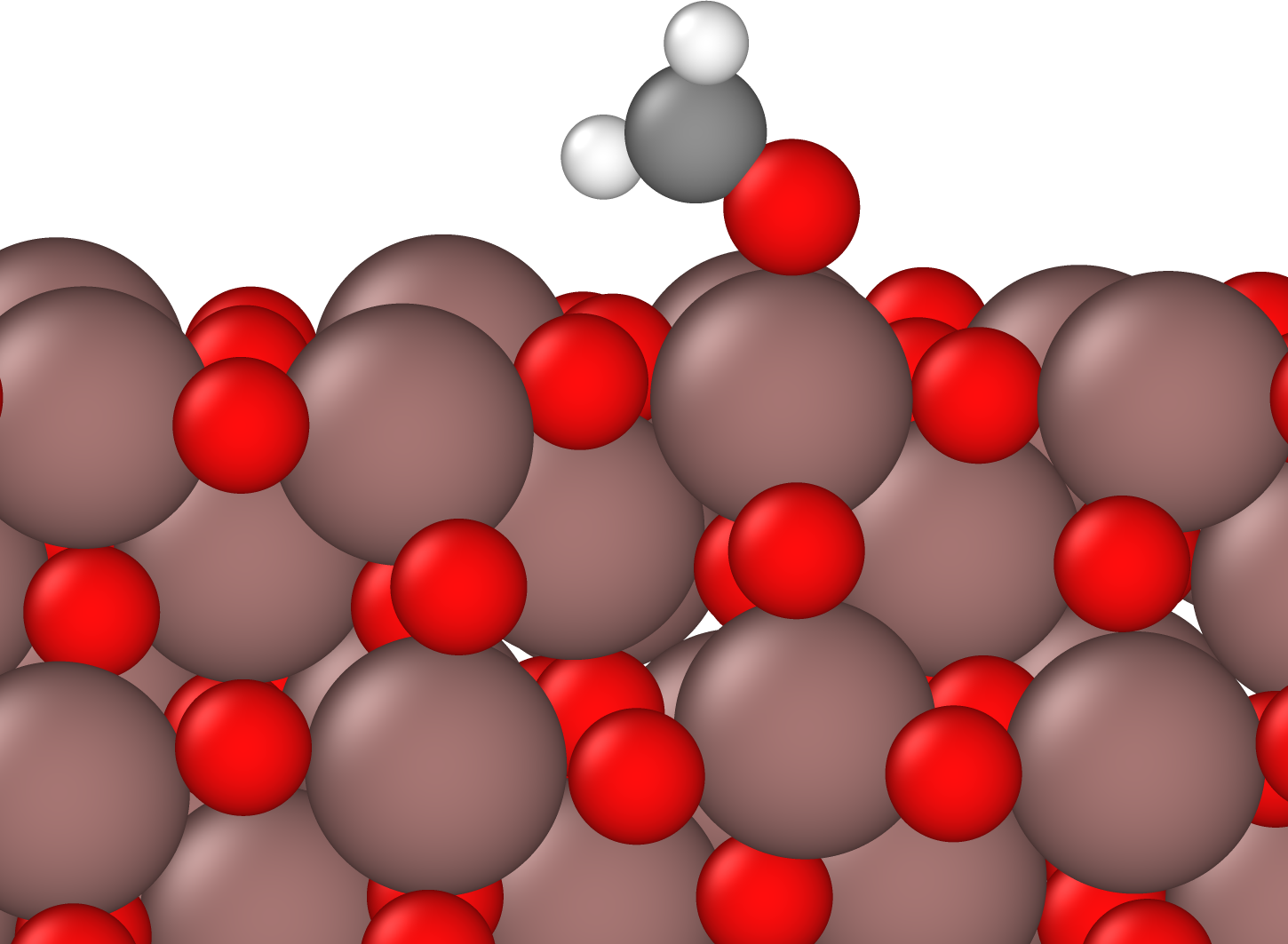}
    \vspace{-12pt}
    \caption{Final}
    \label{fig:lars_b}
  \end{subfigure}
  \\[0.5mm]
  \begin{subfigure}[b]{0.3\linewidth}
    \centering
    \includegraphics[width=\linewidth]{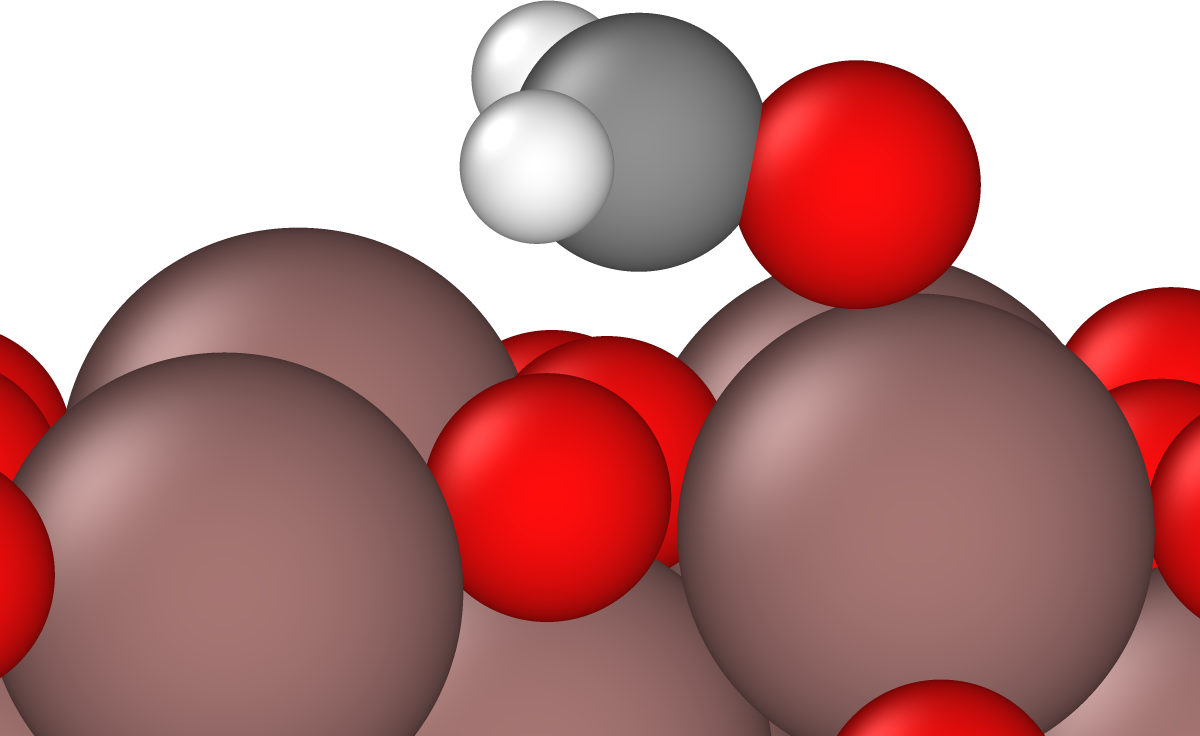}
    \vspace{-12pt}
    \caption{NEB TS \rcross}
    \label{fig:lars_neb_ts}
  \end{subfigure}\hspace{0.005\linewidth}
  \begin{subfigure}[b]{0.3\linewidth}
    \centering
    \includegraphics[width=\linewidth]{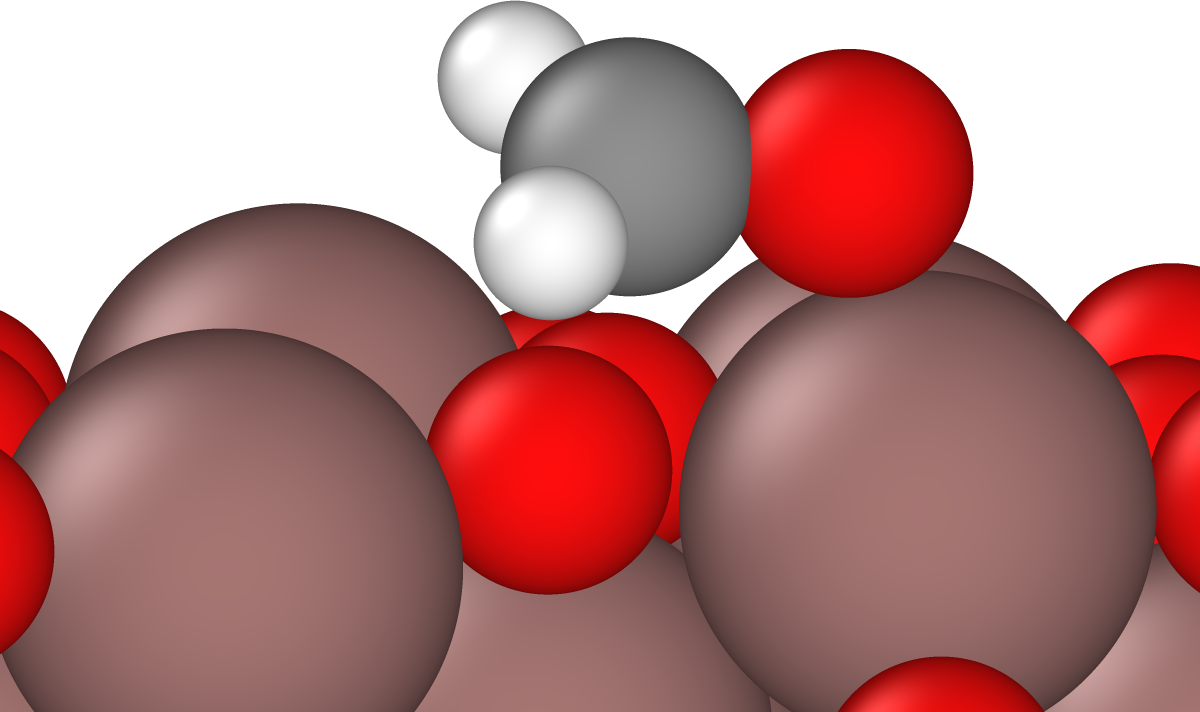}
    \vspace{-12pt}
    \caption{INR TS \gcheck}
    \label{fig:lars_inr_ts}
  \end{subfigure}
  \vspace{-8pt}
  \caption{Conversion of a dioxymethylene species into formaldehyde on an \ce{In2O3} surface. Compared to NEB, the INR identifies a more accurate TS in the same number of iterations.}
  \label{fig:lars}
\end{figure}
We begin by testing the methods on a basic chemical reaction step. Here, we consider the conversion of an adsorbed dioxymethylene (\ce{H2CO2}*) species into formaldehyde (\ce{H2CO}*) on an \ce{In2O3} surface with an oxygen-vacant active site. This is a key step in the hydrogenation of carbon dioxide to methanol over an indium oxide catalyst\cite{dang,lars}. Figs.~\ref{fig:lars_a} and \ref{fig:lars_b} depict the initial and final states of this reaction step. The system is periodic in the $x$ and $y$ directions, with the bottom two (of four) \ce{In2O3} layers fixed, and is modeled with the MACE-MP\cite{mace_mp} potential. 

Fig.~\ref{fig:lars_neb_ts} and \ref{fig:lars_inr_ts} show the estimated TS from both methods, and Fig.~\ref{fig:lars_sella_ts} shows the reference TS obtained by refining either estimate with Sella. Visually, the INR estimates a more accurate TS with energy $1.2364\ \text{eV}$, close to the refined value $1.2362\ \text{eV}$ (Table~\ref{tab:numbers}). The TS energy error plot in Fig.~\ref{fig:lars_E_error} further shows that the INR approaches the TS faster than NEB. In Fig.~\ref{fig:lars_E}, we plot the energy profiles of both methods and for NEB initialized with the INR path. The INR also finds a better approximation of the MEP, aligning closely with the NEB-refined path. However, the advantage for this system is marginal in practice. Both methods find the correct TS within a similar number of energy evaluations when refined after around $50$ iterations, as summarized in Table~\ref{tab:numbers}.

\subsection{\textbf{Alanine dipeptide rotation}}
\begin{figure}[h]
\vspace{-4pt}
  \centering
  \begin{subfigure}[b]{0.22\linewidth}
    \centering
    \includegraphics[width=\linewidth]{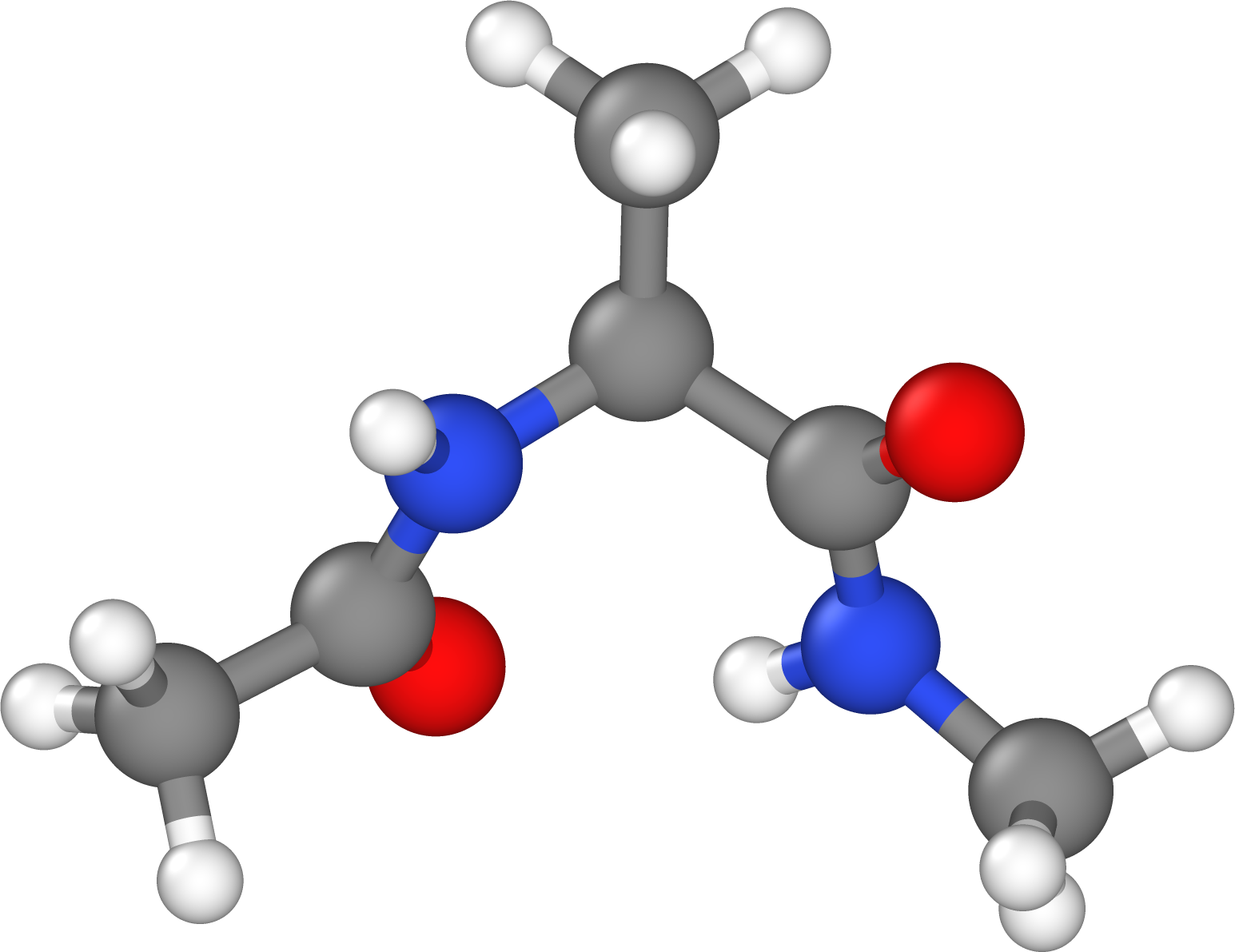}
    \vspace{-12pt}
    \caption{Initial}
    \label{fig:aladi_a}
  \end{subfigure}\hspace{0.003\linewidth}
  \begin{subfigure}[b]{0.22\linewidth}
    \centering
    \includegraphics[width=\linewidth]{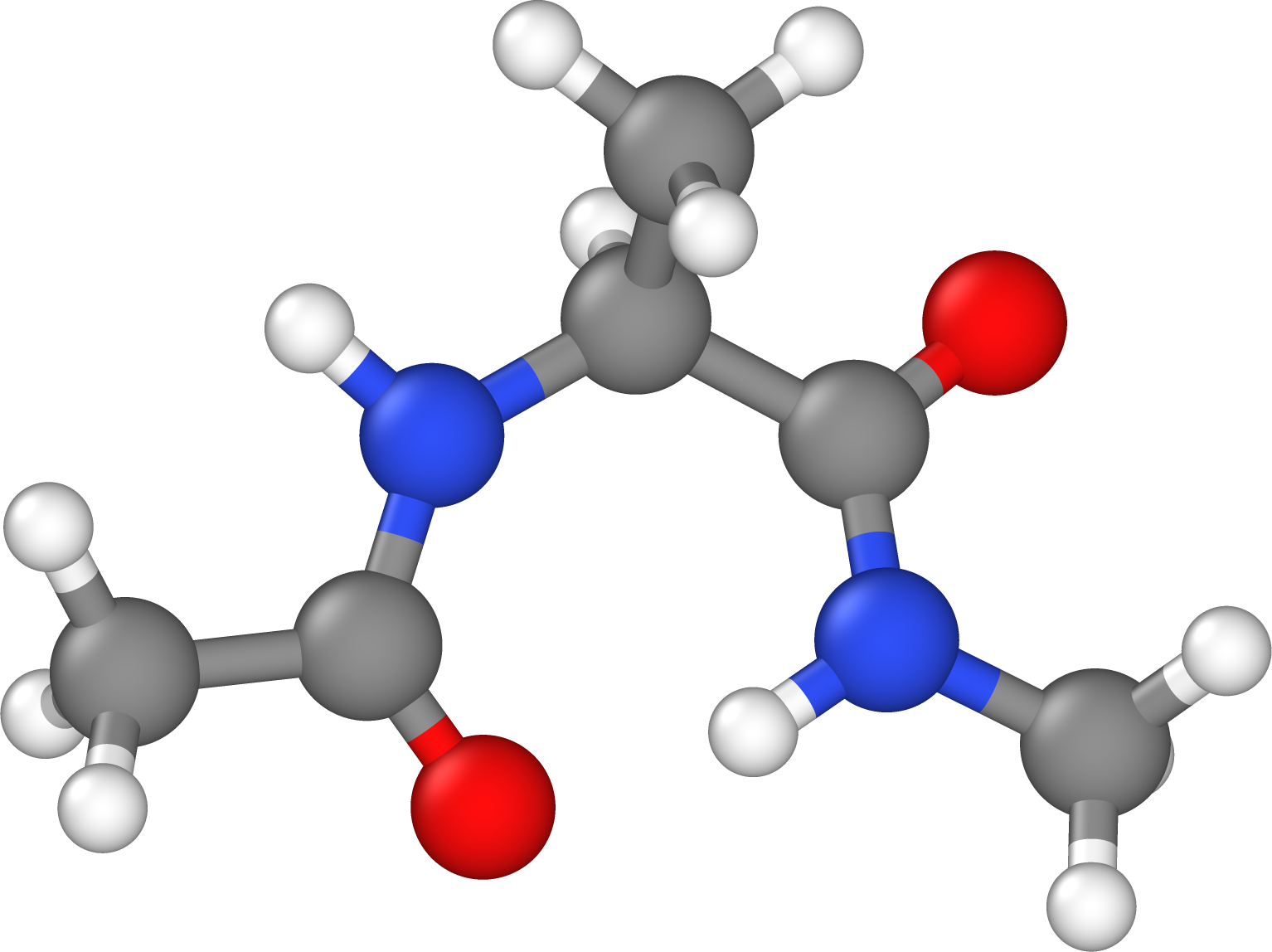}
    \vspace{-12pt}
    \caption{TS}
    \label{fig:aladi_sella_ts}
  \end{subfigure}\hspace{0.003\linewidth}
  \begin{subfigure}[b]{0.22\linewidth}
    \centering
    \includegraphics[width=\linewidth]{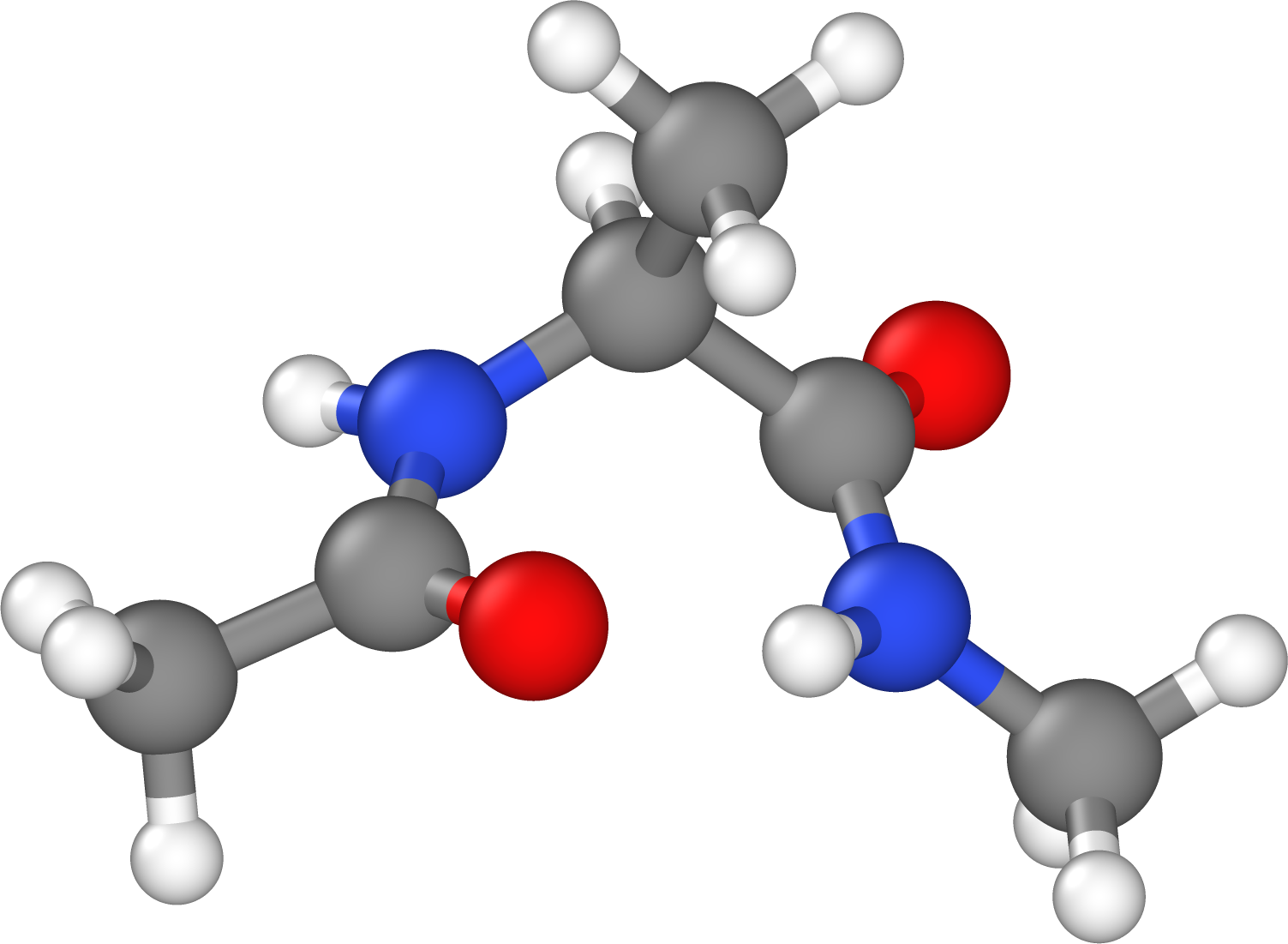}
    \vspace{-12pt}
    \caption{Final}
    \label{fig:aladi_b}
  \end{subfigure}
  \\[0.5mm]
  \begin{subfigure}[b]{0.33\linewidth}
    \centering
    \includegraphics[width=\linewidth]{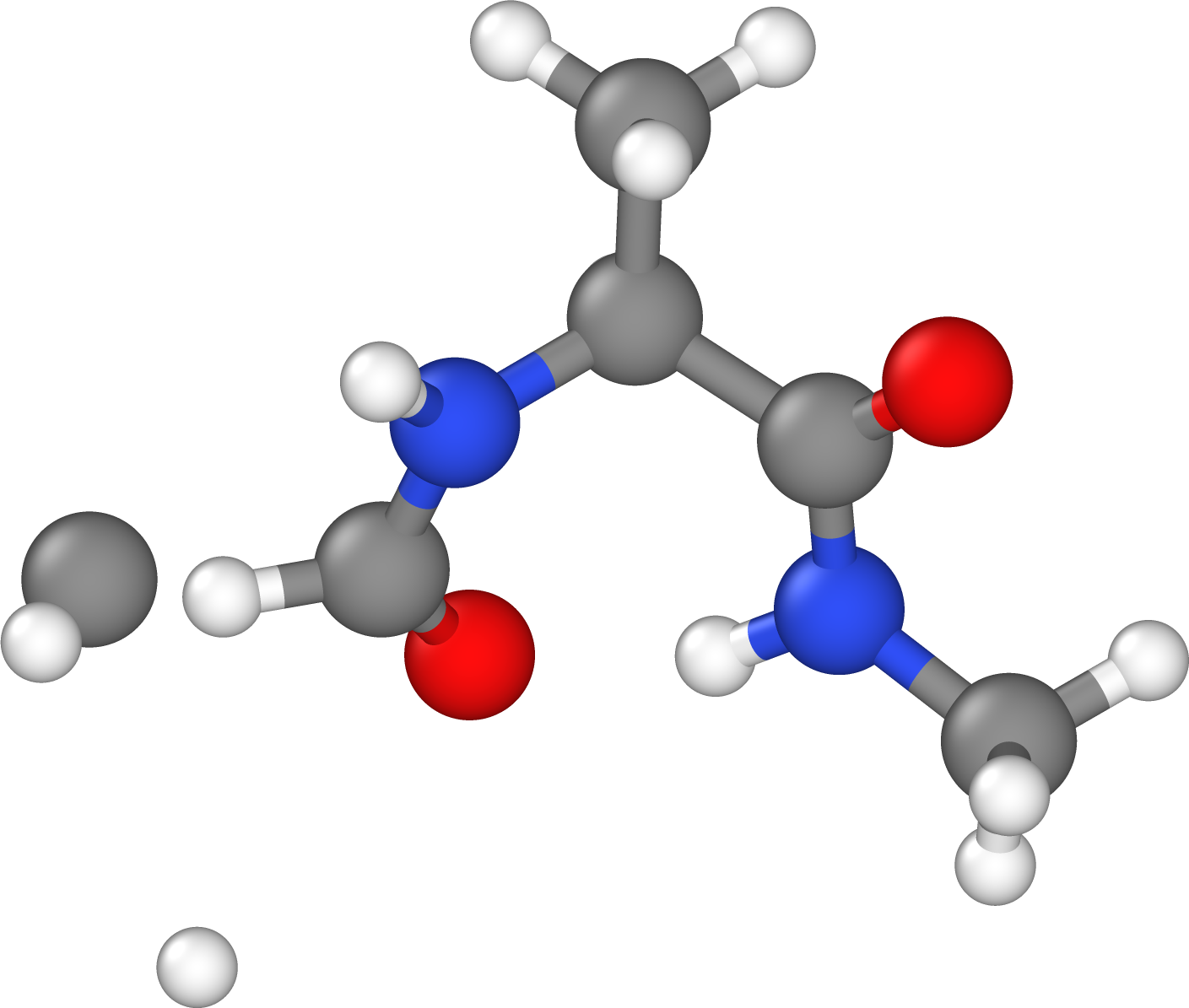}
    \vspace{-12pt}
    \caption{NEB TS \rcross}
    \label{fig:aladi_neb_ts}
  \end{subfigure}\hspace{0.008\linewidth}
  \begin{subfigure}[b]{0.33\linewidth}
    \centering
    \includegraphics[width=\linewidth]{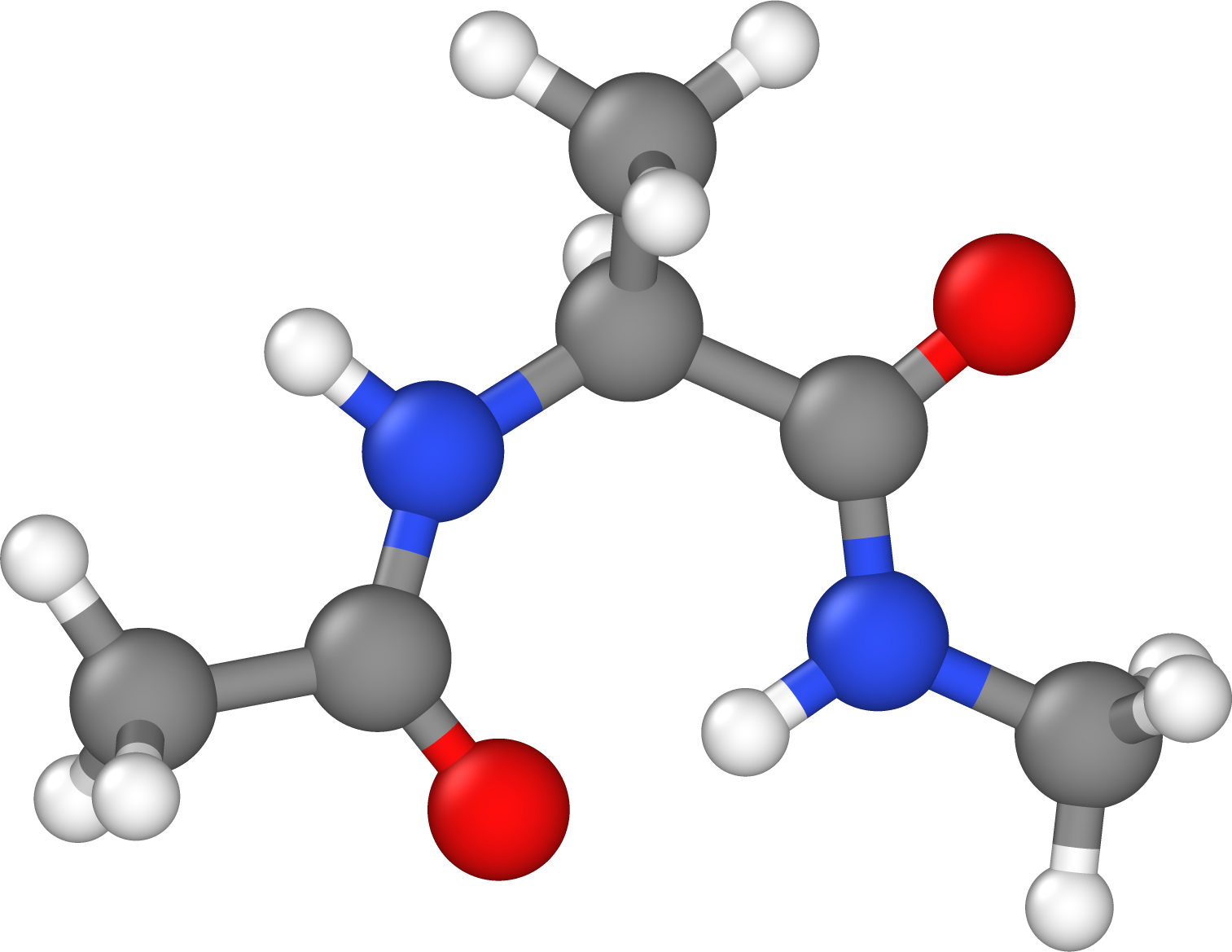}
    \vspace{-12pt}
    \caption{INR TS \gcheck}
    \label{fig:aladi_inr_ts}
  \end{subfigure}\hspace{0.01\linewidth}
  \vspace{-8pt}
  \caption{Dihedral rotation of alanine dipeptide. NEB produces an unphysical TS. The INR preserves bond integrity and estimates the TS reliably. Both methods start from a linear interpolation path.
  }
  \label{fig:aladi}
\end{figure}
A key limitation of NEB is the need for a reasonable initial guess for the reaction path. In the worst case, a poor guess may lead to unphysical atomic configurations, causing the method, or the potential energy calculation, to diverge\cite{gsm,idpp}. A well-known example is the $22$-atom alanine dipeptide (\ce{C6H12N2O2}) reaction, which involves two simultaneous dihedral rotations along the path between the initial and final states shown in Figs.~\ref{fig:aladi_a} and \ref{fig:aladi_b}. Here, the system is modeled with the MACE-OFF\cite{mace_off} potential. 

A linear interpolation of the Cartesian coordinates introduces steric clashes, leading to large energy gradients. Consequently, NEB predicts a TS with broken bonds, as shown in Fig.~\ref{fig:aladi_neb_ts}, and a discontinuous path, as evident from the energy profile in Fig.~\ref{fig:aladi_E}. Sella refinement fails to recover from this unphysical state (Table \ref{tab:numbers}). In contrast, the INR produces a physically meaningful path, preserving atomic connectivity through the dihedral rotations. The TS is shown in Fig.~\ref{fig:aladi_inr_ts} and has energy $0.4544\ \text{eV}$, close to the refined value $0.4070\ \text{eV}$. The error curve in Fig.~\ref{fig:aladi_E_error} reveals that the INR quickly recovers from the high energy, unnatural, initial guess and finds a reasonable estimate of the TS in around $100$ iterations. Thus, unlike NEB, the INR does not rely on a specialized initialization; instead, it leverages its inherent smoothness to learn the reaction path robustly.

\subsection{\textbf{Methylidyne diffusion}}
\begin{figure}[h]
\vspace{-6pt}
  \centering
  \begin{subfigure}[b]{0.19\linewidth}
    \centering
    \includegraphics[width=\linewidth]{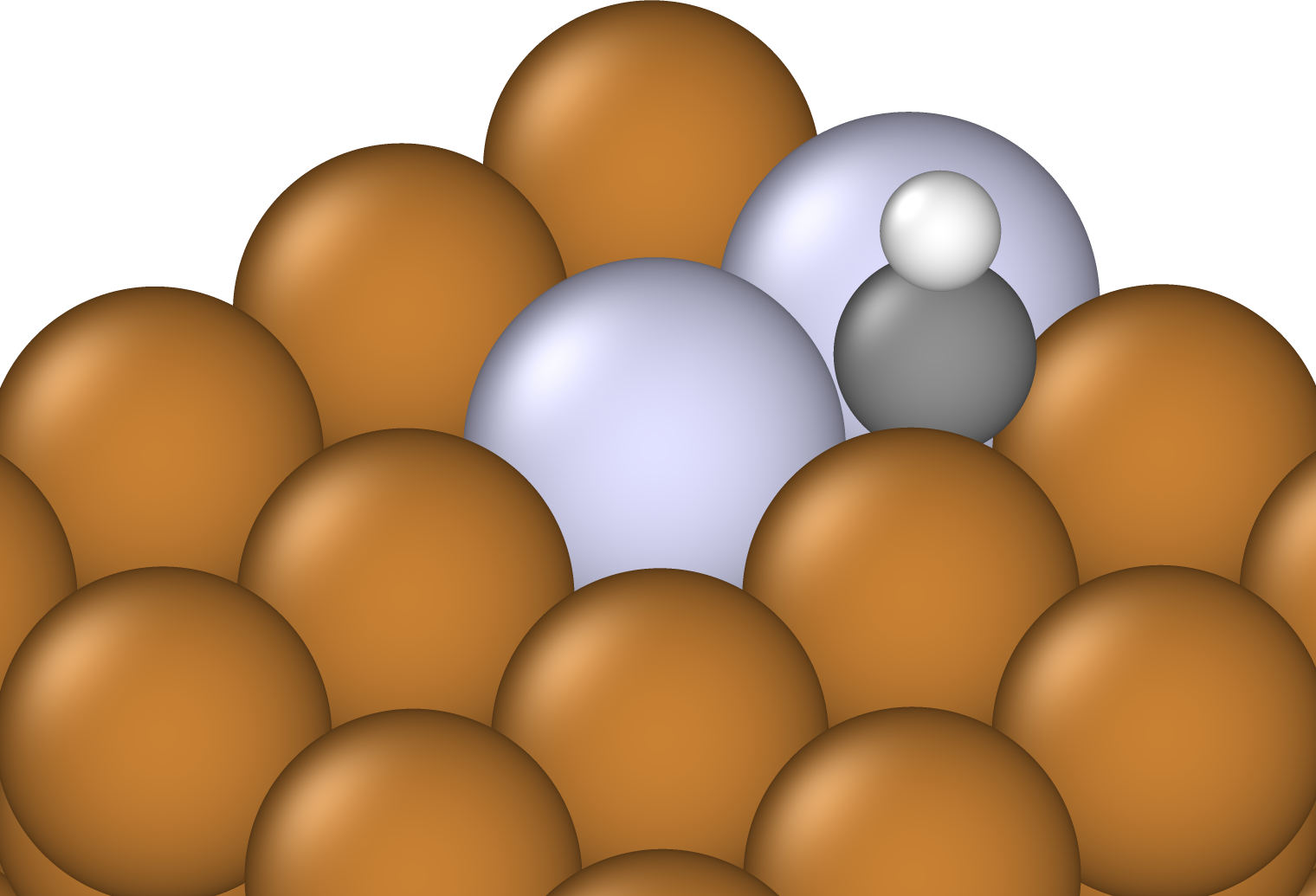}
    \vspace{-12pt}
    \caption{Initial}
    \label{fig:sine_a}
  \end{subfigure}\hspace{0.001\linewidth}
  \begin{subfigure}[b]{0.19\linewidth}
    \centering
    \includegraphics[width=\linewidth]{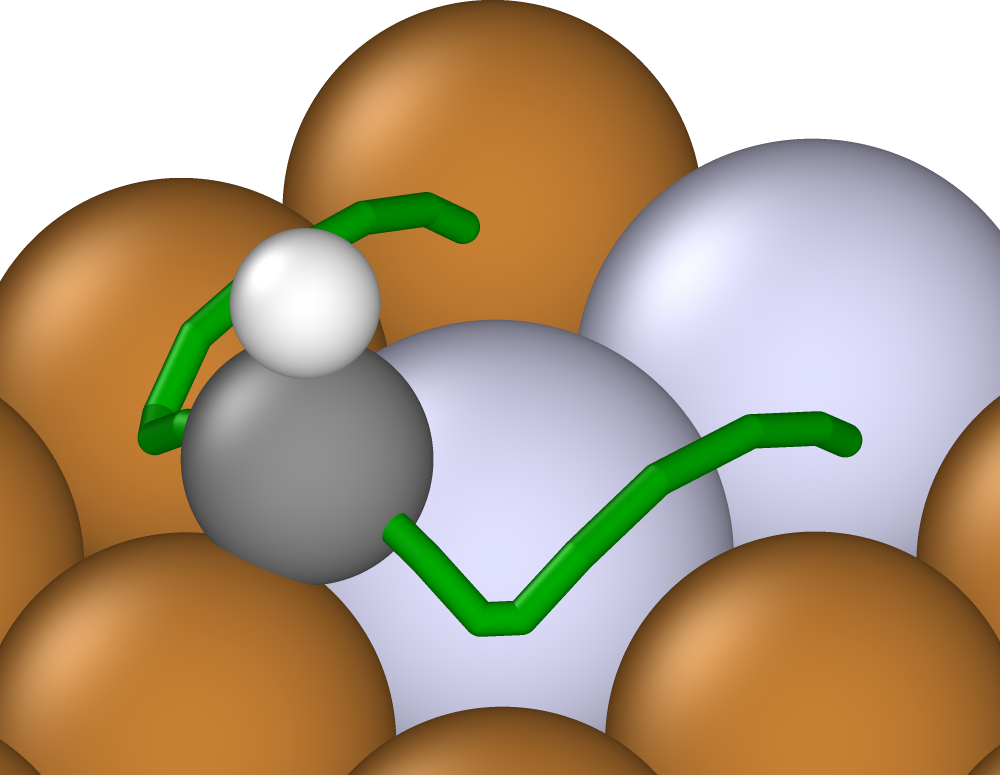}
    \vspace{-12pt}
    \caption{TS}
    \label{fig:sine_sella_ts}
  \end{subfigure}\hspace{0.001\linewidth}
  \begin{subfigure}[b]{0.19\linewidth}
    \centering
    \includegraphics[width=\linewidth]{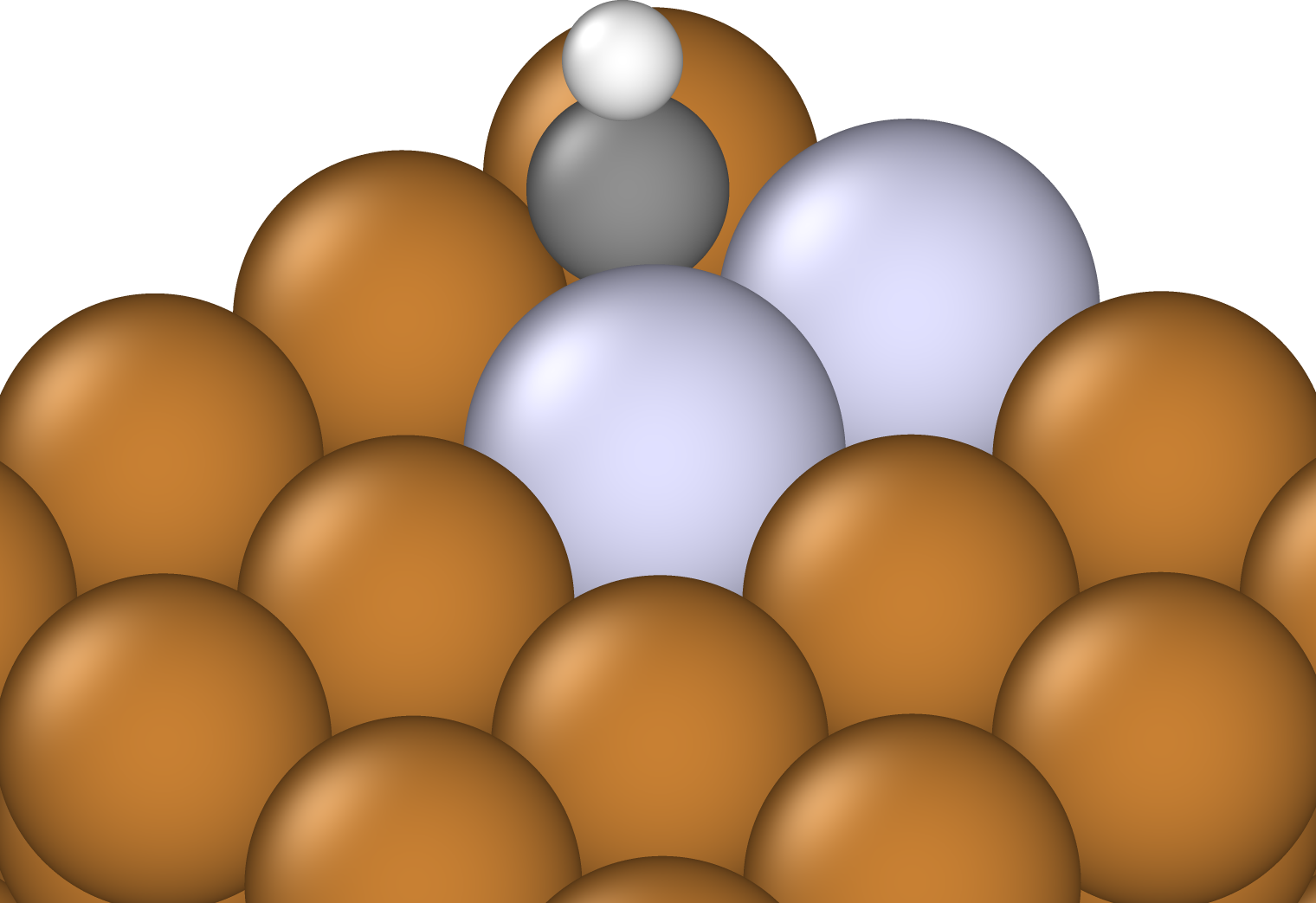}
    \vspace{-12pt}
    \caption{Final}
    \label{fig:sine_b}
  \end{subfigure}
  \\[0.5mm]
  \begin{subfigure}[b]{0.28\linewidth}
    \centering
    \includegraphics[width=\linewidth]{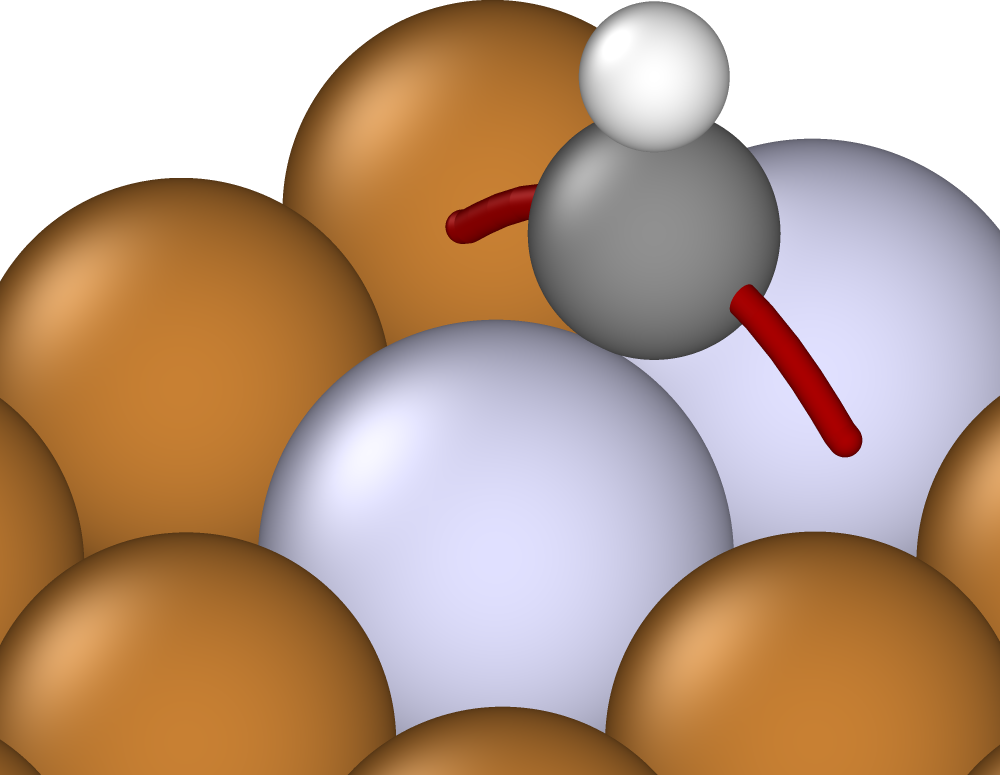}
    \vspace{-12pt}
    \caption{NEB TS \rcross}
    \label{fig:sine_neb_ts}
  \end{subfigure}\hspace{0.005\linewidth}
  \begin{subfigure}[b]{0.28\linewidth}
    \centering
    \includegraphics[width=\linewidth]{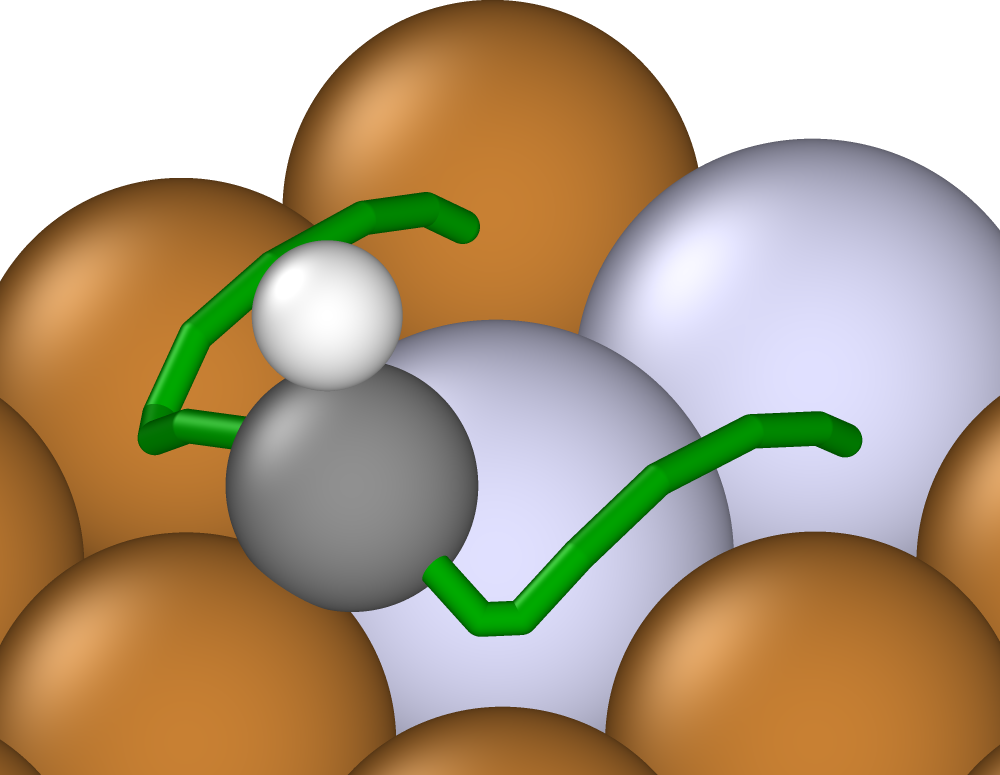}
    \vspace{-12pt}
    \caption{INR-GS TS \gcheck}
    \label{fig:sine_inr_ts}
  \end{subfigure}
  \vspace{-6pt}
  \caption{Diffusion of a methylidyne species on a \ce{Ag}/\ce{Cu} surface. NEB is trapped in the higher-barrier direct path. The INR (growing sampling) finds the lower-barrier route avoiding the \ce{Ag}--\ce{Ag} bridge.
  }
  \label{fig:sine}
\end{figure}
Another limitation of NEB is its tendency to become trapped in local minimum paths, illustrated earlier on the two-dimensional sine-n systems (Sec.~\ref{subsec:2d}). Here, we consider an atomistic setting where the same issue arises. The system involves the surface diffusion of an adsorbed methylidyne (*\ce{CH}) species on a heterogeneous catalyst surface composed of \ce{Cu} doped with \ce{Ag} atoms, shown in Figs.~\ref{fig:sine_a} and \ref{fig:sine_b}. It is periodic in the $x$ and $y$ directions, with six fixed layers of \ce{Cu}, and is modeled with the MACE-MP potential. A direct path crosses the \ce{Ag}--\ce{Ag} bridge site and is a valid MEP but encounters a larger barrier. An alternative, lower-barrier route avoids the \ce{Ag}--\ce{Ag} bridge, traversing the \ce{Cu}--\ce{Ag} sites instead. 

NEB gets trapped in the direct path as indicated by the estimated TS in Fig.~\ref{fig:sine_neb_ts} and the energy profile in Fig.~\ref{fig:sine_E}. On the other hand, applying the growing sampling idea from Sec.~\ref{subsec:sampling}, the INR identifies the lower-barrier path with (refined) TS energy $1.4142\ \text{eV}$, compared to $2.3539\ \text{eV}$ for the direct path. In Fig.~\ref{fig:sine_final}, we show more samples along the INR path, and in Appendix~\ref{app:sec:sine}, we present its evolution over iterations. Our results clearly highlight the versatility of the approach, showing how a simple adjustment to the sampling strategy can help escape local minima.
\begin{figure}[b]
  \centering
  \begin{subfigure}[b]{0.22\linewidth}
    \centering
    \includegraphics[width=\linewidth]{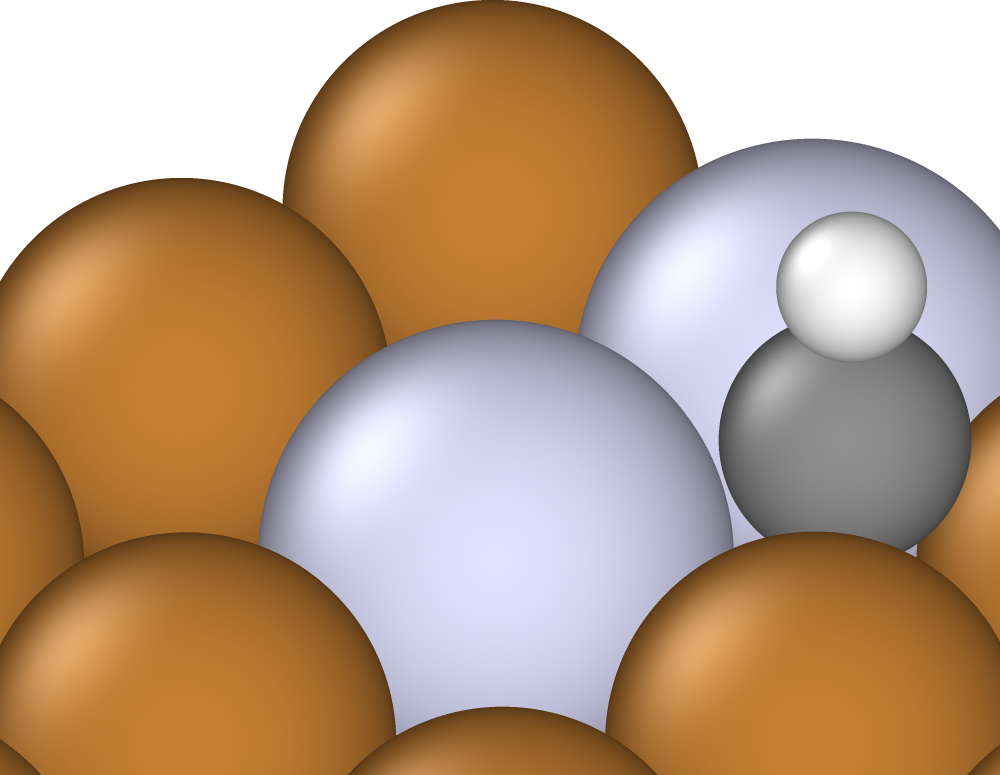}
    \vspace{-12pt}
    \caption{}
  \end{subfigure}\hspace{0.001\linewidth}
  \begin{subfigure}[b]{0.22\linewidth}
    \centering
    \includegraphics[width=\linewidth]{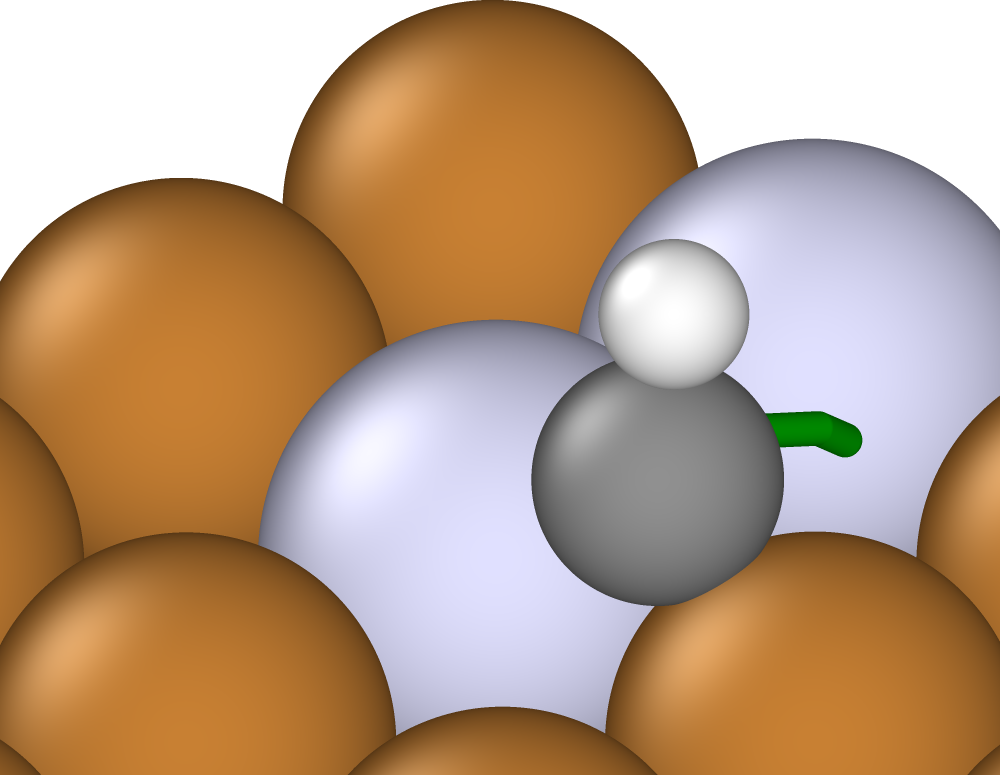}
    \vspace{-12pt}
    \caption{}
  \end{subfigure}\hspace{0.001\linewidth}
  \begin{subfigure}[b]{0.22\linewidth}
    \centering
    \includegraphics[width=\linewidth]{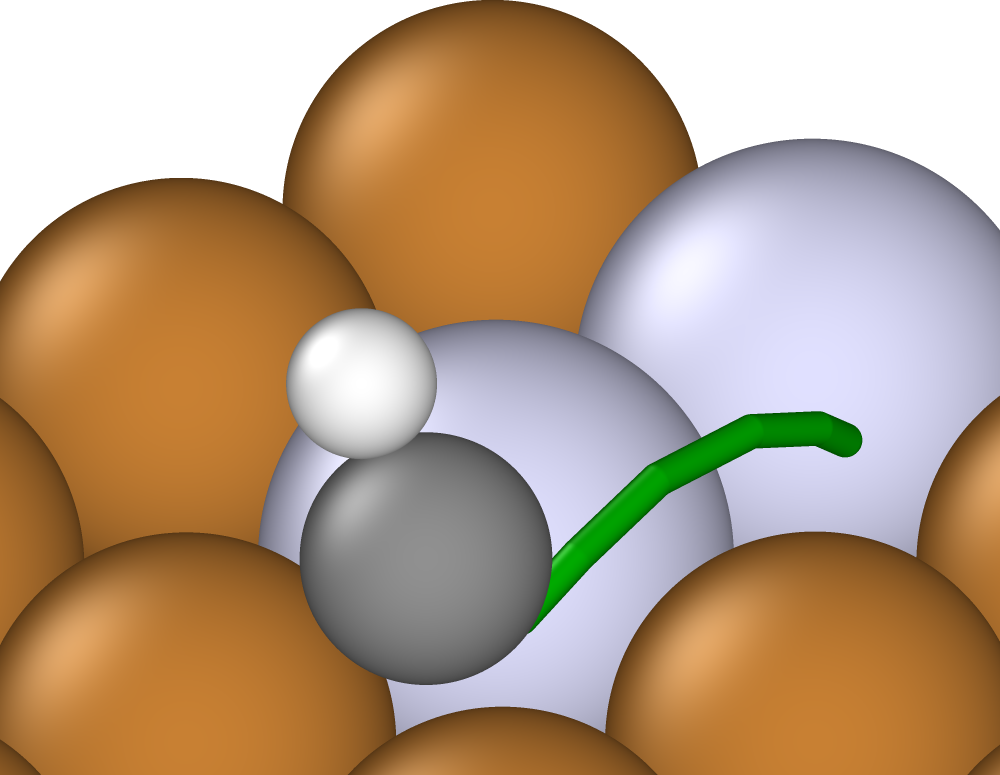}
    \vspace{-12pt}
    \caption{}
  \end{subfigure}\hspace{0.001\linewidth}
  \\[0.5mm]
  \begin{subfigure}[b]{0.22\linewidth}
    \centering
    \includegraphics[width=\linewidth]{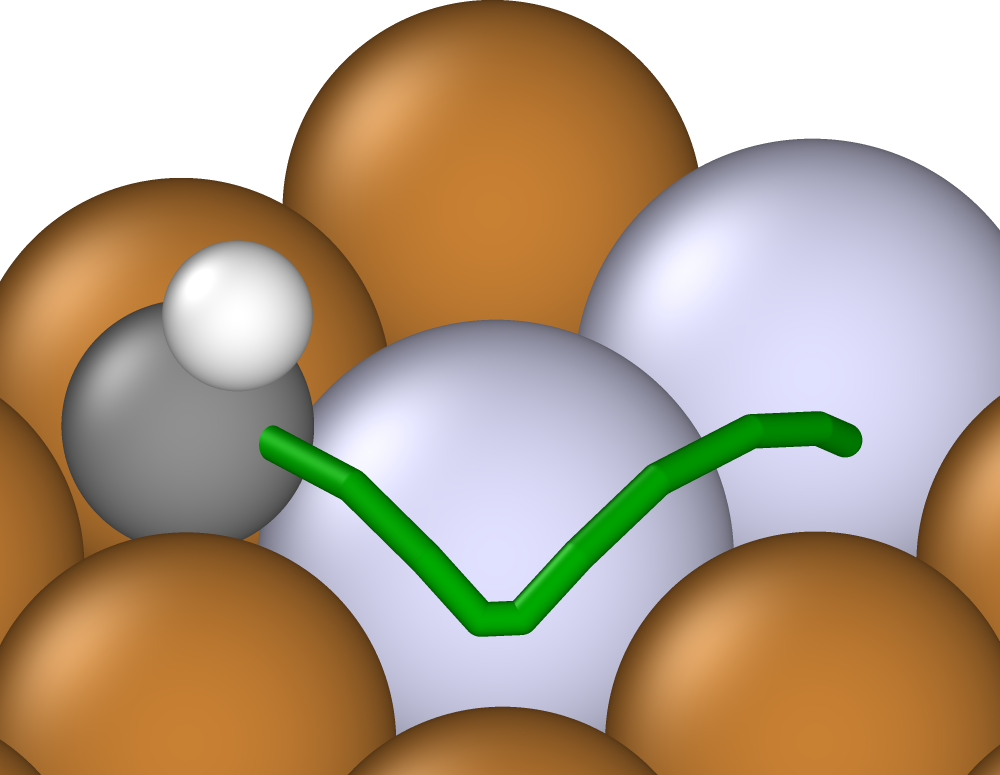}
    \vspace{-12pt}
    \caption{}
  \end{subfigure}\hspace{0.001\linewidth}
  \begin{subfigure}[b]{0.22\linewidth}
    \centering
    \includegraphics[width=\linewidth]{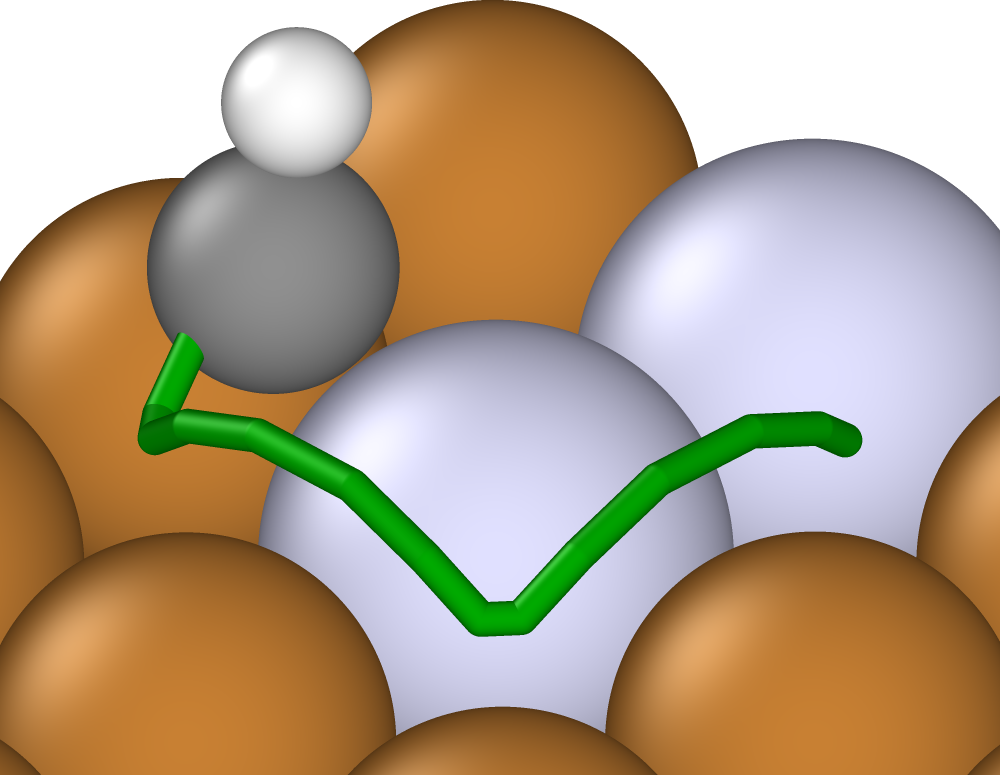}
    \vspace{-12pt}
    \caption{}
  \end{subfigure}\hspace{0.001\linewidth}
  \begin{subfigure}[b]{0.22\linewidth}
    \centering
    \includegraphics[width=\linewidth]{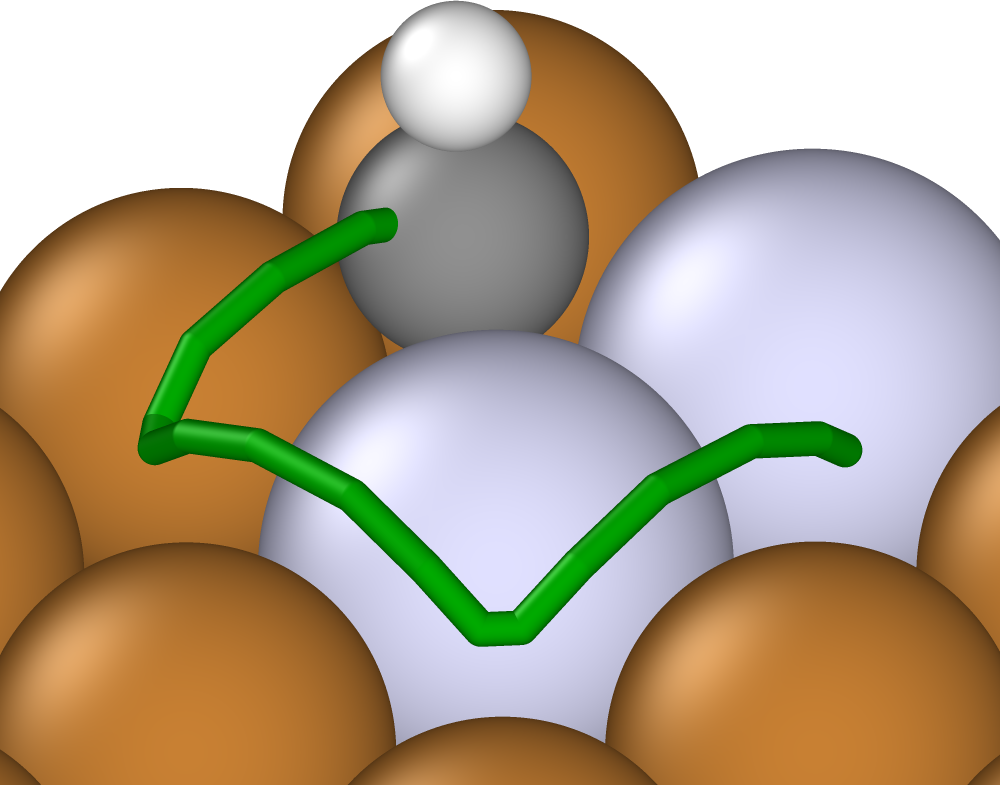}
    \vspace{-12pt}
    \caption{}
  \end{subfigure}
  \vspace{-8pt}
  \caption{Samples (a--f) along the INR-GS path for the methylidyne diffusion system. The adsorbed species passes over the \ce{Cu}--\ce{Ag} sites, avoiding the higher-barrier \ce{Ag}--\ce{Ag} bridge.}
  \label{fig:sine_final}
\end{figure}

\subsection{\textbf{\ce{N2 + H2}}}
\begin{figure}[h]
\vspace{-6pt}
  \centering
  \begin{subfigure}[b]{0.22\linewidth}
    \centering
    \includegraphics[width=\linewidth]{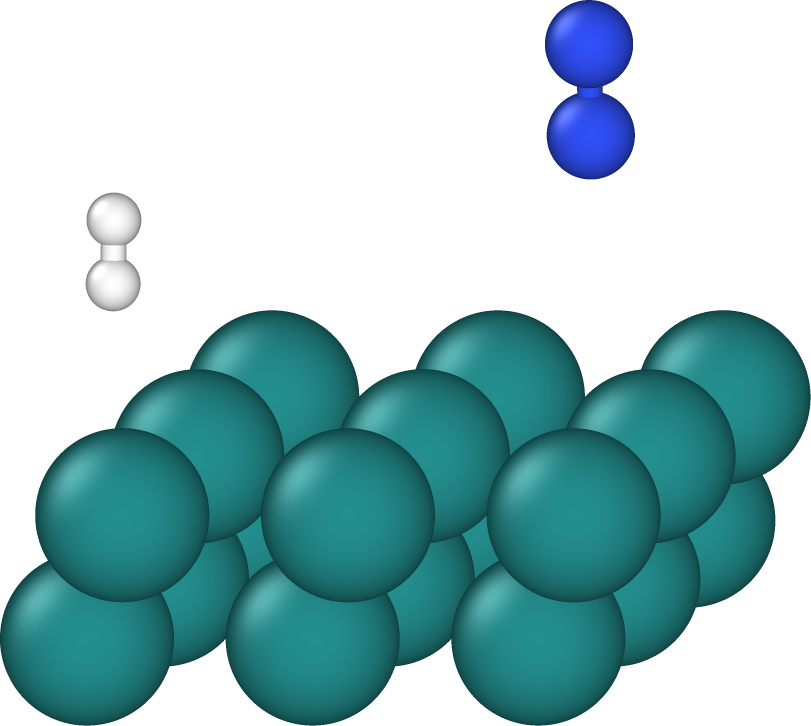}
    \vspace{-12pt}
    \caption{Initial}
    \label{fig:n2h2_a}
  \end{subfigure}
  \begin{subfigure}[b]{0.22\linewidth}
    \centering
    \includegraphics[width=\linewidth]{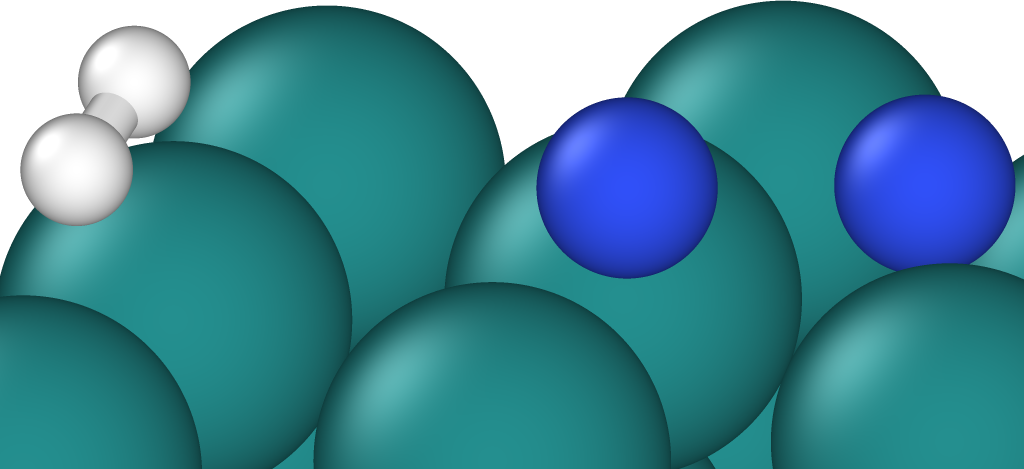}
    \vspace{-12pt}
    \caption{TS}
    \label{fig:n2h2_sella_ts}
  \end{subfigure}\hspace{0.001\linewidth}
  \begin{subfigure}[b]{0.22\linewidth}
    \centering
    \includegraphics[width=\linewidth]{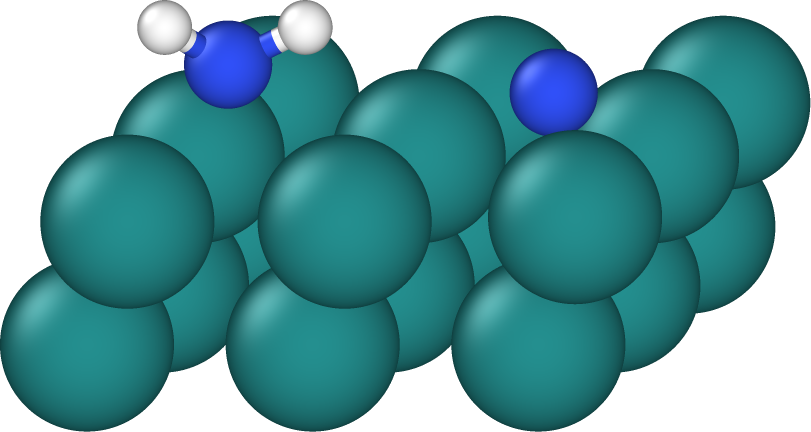}
    \vspace{-12pt}
    \caption{Final}
    \label{fig:n2h2_b}
  \end{subfigure}
  \\[0.5mm]
  \begin{subfigure}[b]{0.33\linewidth}
    \centering
    \includegraphics[width=\linewidth]{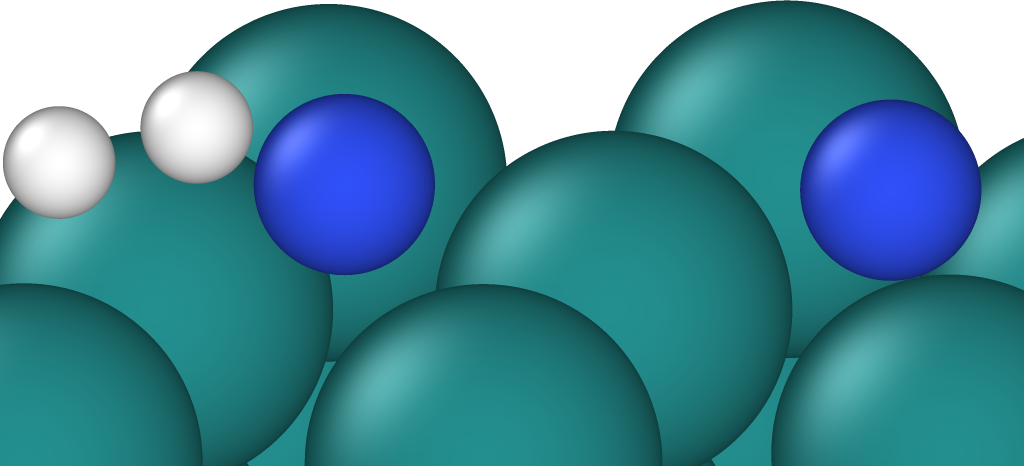}
    \vspace{-12pt}
    \caption{NEB TS \rcross}
    \label{fig:n2h2_neb_ts}
  \end{subfigure}\hspace{0.005\linewidth}
  \begin{subfigure}[b]{0.33\linewidth}
    \centering
    \includegraphics[width=\linewidth]{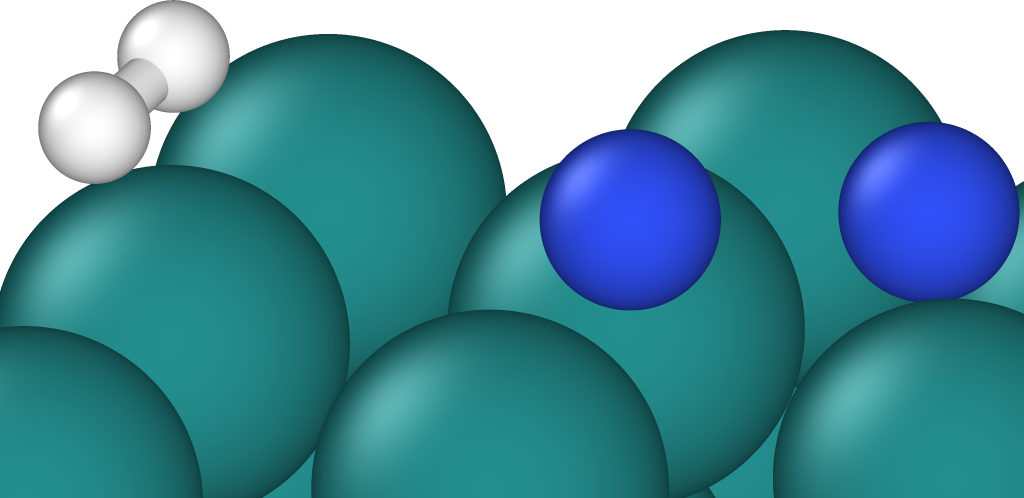}
    \vspace{-12pt}
    \caption{INR TS \gcheck}
    \label{fig:n2h2_inr_ts}
  \end{subfigure}
  \vspace{-8pt}
  \caption{Reaction of \ce{N2} and \ce{H2} to form adsorbed *\ce{NH2} on a \ce{Ru} surface. NEB fails to converge and produces an incorrect TS. The INR TS correctly represents the rate-limiting \ce{N2} dissociation step.}
  \label{fig:n2h2}
\end{figure}

\begin{figure}[b]
  \centering
  \begin{subfigure}[b]{0.24\linewidth}
    \centering
    \includegraphics[width=\linewidth]{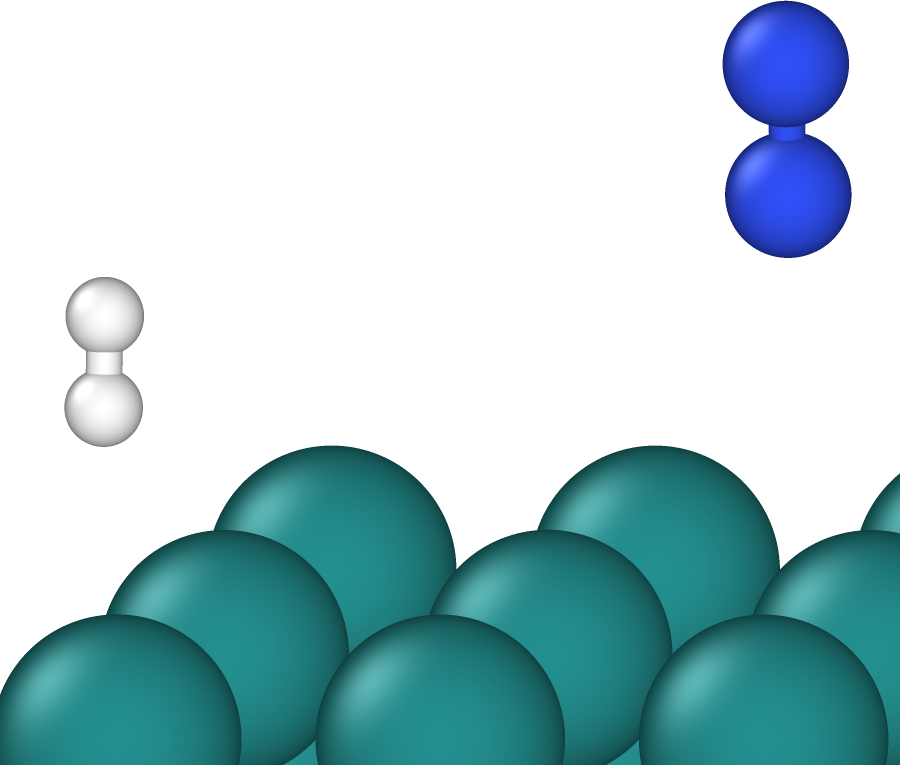}
    \vspace{-12pt}
    \caption{}
  \end{subfigure}\hspace{0.001\linewidth}
  \begin{subfigure}[b]{0.24\linewidth}
    \centering
    \includegraphics[width=\linewidth]{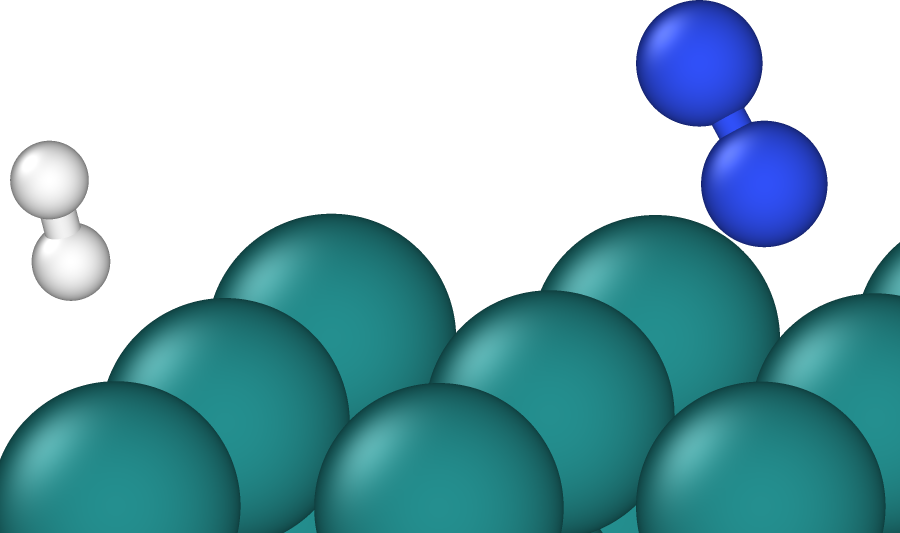}
    \vspace{-12pt}
    \caption{}
  \end{subfigure}\hspace{0.001\linewidth}
  \begin{subfigure}[b]{0.24\linewidth}
    \centering
    \includegraphics[width=\linewidth]{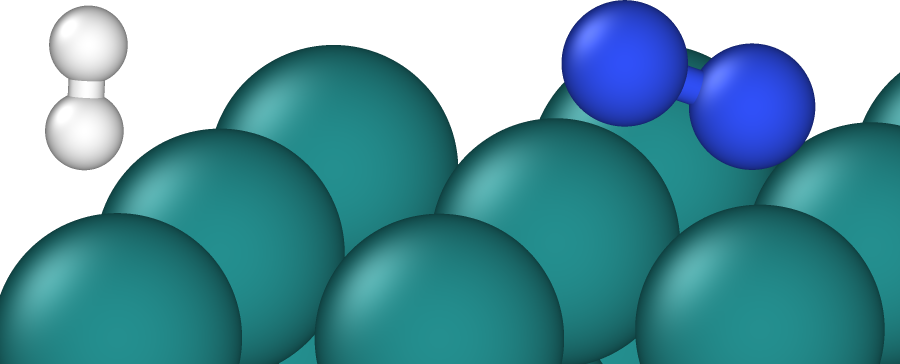}
    \vspace{-12pt}
    \caption{}
  \end{subfigure}\hspace{0.001\linewidth}
  \begin{subfigure}[b]{0.24\linewidth}
    \centering
    \includegraphics[width=\linewidth]{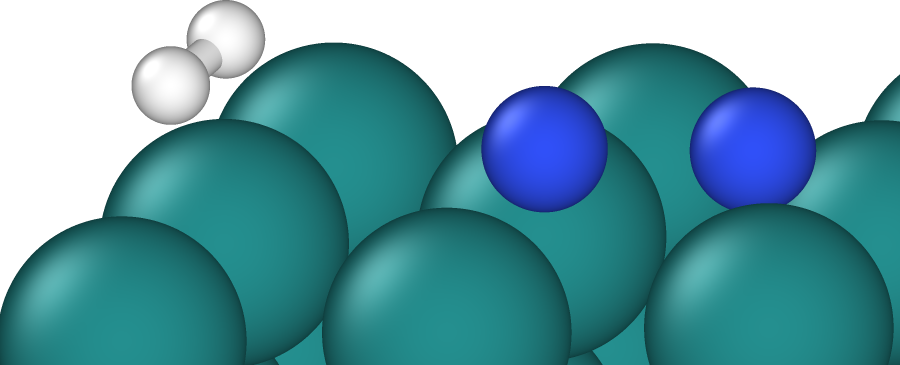}
    \vspace{-12pt}
    \caption{}
  \end{subfigure}
  \\[0.5mm]
  \begin{subfigure}[b]{0.24\linewidth}
    \centering
    \includegraphics[width=\linewidth]{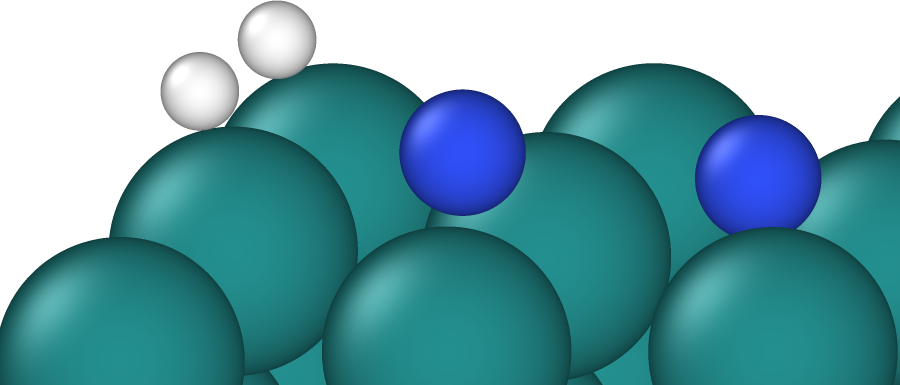}
    \vspace{-12pt}
    \caption{}
    \label{fig:hyd1}
  \end{subfigure}\hspace{0.001\linewidth}
  \begin{subfigure}[b]{0.24\linewidth}
    \centering
    \includegraphics[width=\linewidth]{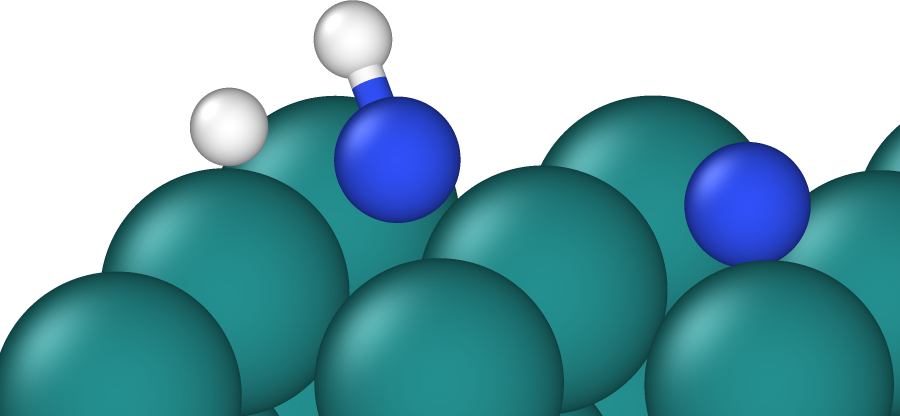}
    \vspace{-12pt}
    \caption{}
    \label{fig:hyd2}
  \end{subfigure}\hspace{0.001\linewidth}
  \begin{subfigure}[b]{0.24\linewidth}
    \centering
    \includegraphics[width=\linewidth]{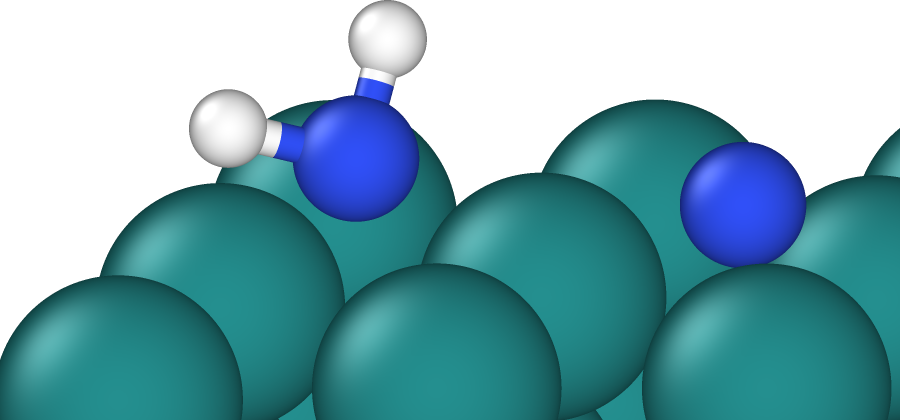}
    \vspace{-12pt}
    \caption{}
  \end{subfigure}\hspace{0.001\linewidth}
  \begin{subfigure}[b]{0.24\linewidth}
    \centering
    \includegraphics[width=\linewidth]{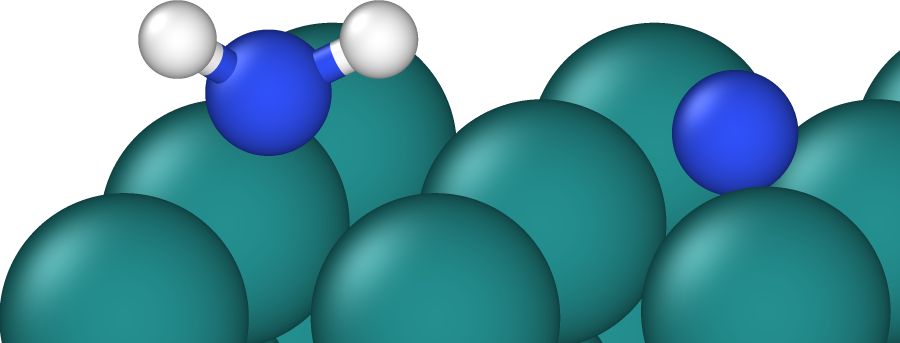}
    \vspace{-12pt}
    \caption{}
  \end{subfigure}
  \vspace{-18pt}
  \caption{Samples (a--h) along the INR path for the \ce{N2 + H2} system, showing \ce{N2} dissociation (d), \ce{H2} dissociation (e), and *\ce{N} hydrogenation (f--g).}
  \label{fig:n2h2_final}
\end{figure}

To test the ability of the INR to represent more complex reaction paths, we consider a reaction sequence in which \ce{N2} and \ce{H2} interact on a \ce{Ru} catalyst surface, leading to the formation of an adsorbed *\ce{NH2} intermediate. This marks the beginning of the dissociative pathway for ammonia synthesis, involving \ce{N2} and \ce{H2} dissociation followed by hydrogenation of the adsorbed *\ce{N} atoms\cite{nh3_o3}. The chosen initial and final states are shown in Figs.~\ref{fig:n2h2_a} and \ref{fig:n2h2_b}. The system is periodic in the $x$ and $y$ directions, with the bottom \ce{Ru} layer fixed. We model it with a MACE potential trained specifically for this system (see Appendix~\ref{app:sec:n2h2} for details). 

\begin{figure}[t]
  \centering
  \begin{subfigure}[b]{0.49\linewidth}
    \centering
    \includegraphics[width=\linewidth]{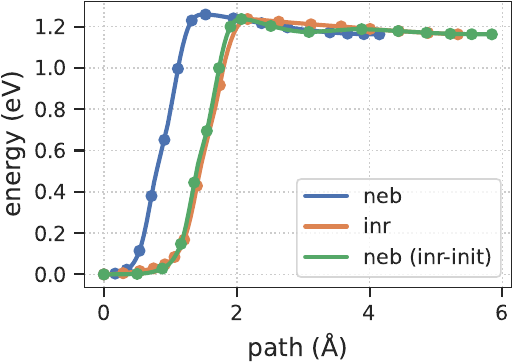}
    \vspace{-15pt}
    \caption{\ce{In2O3}}
    \label{fig:lars_E}
  \end{subfigure}
  \begin{subfigure}[b]{0.49\linewidth}
    \centering
    \includegraphics[width=\linewidth]{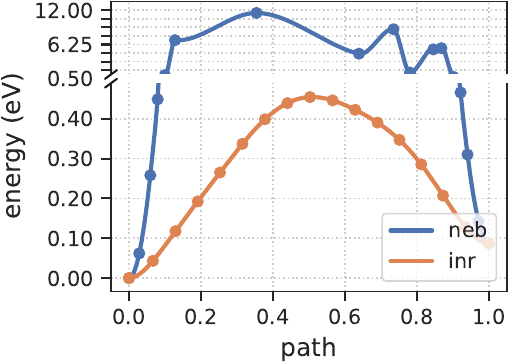}
    \vspace{-15pt}
    \caption{Alanine dipeptide}
    \label{fig:aladi_E}
  \end{subfigure}
  \\[0.5mm]
  \begin{subfigure}[b]{0.49\linewidth}
    \centering
    \includegraphics[width=\linewidth]{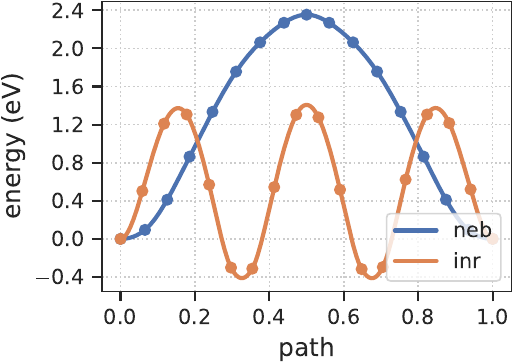}
    \vspace{-15pt}
    \caption{Methylidyne diffusion (INR-GS)}
    \label{fig:sine_E}
  \end{subfigure}
  \begin{subfigure}[b]{0.49\linewidth}
    \centering
    \includegraphics[width=\linewidth]{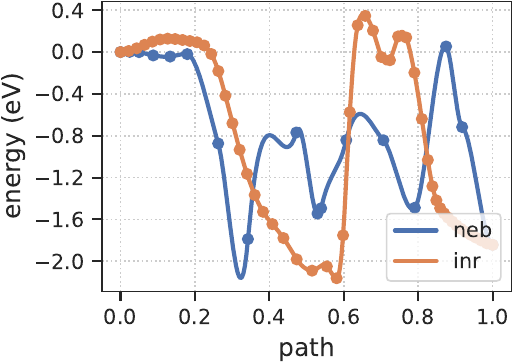}
    \vspace{-15pt}
    \caption{\ce{N2 + H2}}
    \label{fig:n2h2_E}
  \end{subfigure}
  \vspace{-8pt}
  \caption{Energy profiles for chemical systems where the INR (a) better approximates an MEP, (b) is resilient to a poor initial guess, (c) finds a lower-barrier solution, and (d) captures a complex path.}
  \label{fig:inr_profiles_3d}
\end{figure}

\begin{figure}[t]
  \centering
  \begin{subfigure}[b]{0.49\linewidth}
    \centering
    \includegraphics[width=\linewidth]{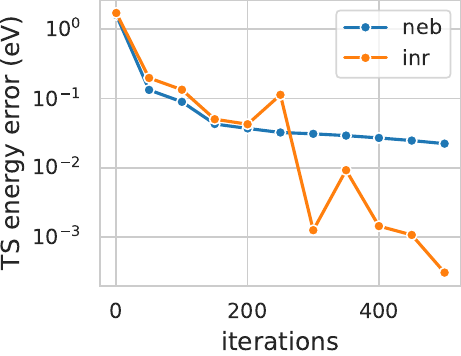}
    \vspace{-15pt}
    \caption{\ce{In2O3}}
    \label{fig:lars_E_error}
  \end{subfigure}
  \begin{subfigure}[b]{0.49\linewidth}
    \centering
    \includegraphics[width=\linewidth]{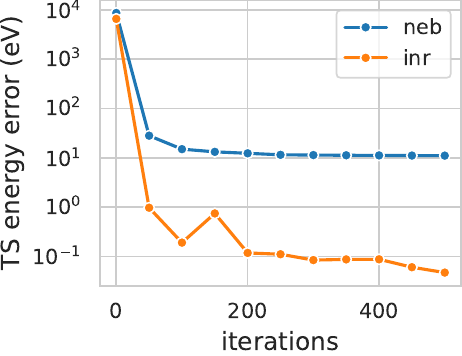}
    \vspace{-15pt}
    \caption{Alanine dipeptide}
    \label{fig:aladi_E_error}
  \end{subfigure}
  \\[0.5mm]
  \begin{subfigure}[b]{0.49\linewidth}
    \centering
    \includegraphics[width=\linewidth]{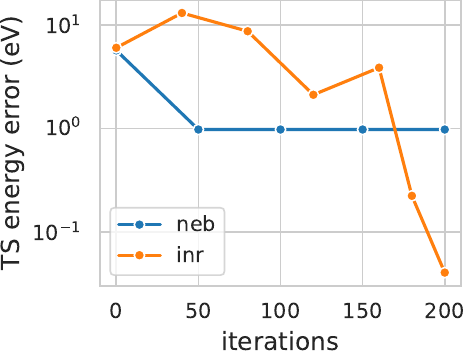}
    \vspace{-15pt}
    \caption{Methylidyne diffusion (INR-GS)}
    \label{fig:sine_E_error}
  \end{subfigure}
  \begin{subfigure}[b]{0.49\linewidth}
    \centering
    \includegraphics[width=\linewidth]{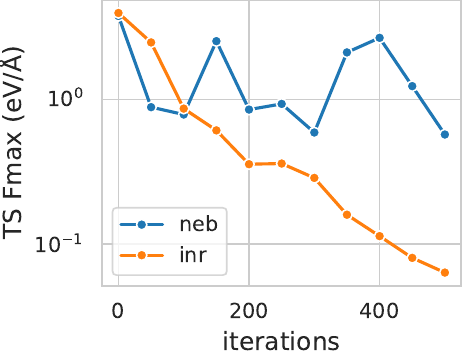}
    \vspace{-15pt}
    \caption{\ce{N2 + H2}}
    \label{fig:n2h2_fmax}
  \end{subfigure}
  \vspace{-8pt}
  \caption{Training progress for chemical systems where the INR (a) approaches a TS faster, (b) recovers from a bad start, (c) avoids a suboptimal path, and (d) predicts the TS for a complex system.}
  \label{fig:progress}
\end{figure}

Fig.~\ref{fig:n2h2_final} shows samples along the path learned by the INR. Qualitatively, the INR identifies the elementary steps of this complex reaction within a single optimization, consistent with the established mechanism (for an \ce{Fe} surface\cite{nh3_o3}) given by
\begin{align}
\ce{N2 + 2\text{*} &-> 2N\text{*}},\nonumber\\
\ce{H2 + 2\text{*} &<=> 2H\text{*}},\nonumber\\
\ce{N\text{*} + H\text{*} &<=> NH\text{*} + \text{*}},\nonumber\\
\ce{NH\text{*} + H\text{*} &<=> NH2\text{*} + \text{*}},
\end{align}
where * indicates an empty site on the surface. The estimated TS, shown in Fig.~\ref{fig:n2h2_inr_ts}, correctly corresponds to the rate-limiting\cite{nh3_o3,nh3_science} \ce{N2} dissociation step. This appears as the highest point on the energy profile in Fig.~\ref{fig:n2h2_E} (showing $50$ samples for visualization only). The next peak (following the barrierless \ce{H2} dissociation\cite{nh3_o3}) visually corresponds to the first hydrogenation (Figs.~\ref{fig:hyd1}-\ref{fig:hyd2}), while no separate peak is observed for the second hydrogenation. Note that we sample $15$ inputs for training, prioritizing just the highest energy point (Eq.~\ref{eq:cl}). A more refined sampling strategy may better capture additional complexities. In contrast, NEB fails to converge for this multi-step reaction, as evident from the $F_\text{max}$ plot in Fig.~\ref{fig:n2h2_fmax}, resulting in an incorrect TS estimate. NEB would likely require a separate run for each manually defined elementary step\cite{dneb}. Thus, our results highlight the INR's ability to automatically express a complex reaction path, significantly reducing the need for manual supervision.

\begin{table*}[t]
\centering
\setlength{\tabcolsep}{2.9pt} 
\renewcommand{\arraystretch}{1.2} 
\caption{Quantitative results. ``No refinement'' runs just the method for $500$ iterations ($200$ for INR-GS). Best energies are in \textbf{bold}.}
\vspace{-6pt}
\label{tab:numbers}
\begin{tabular}{cc|c|cc|cc}
\hline
\multirow{2}{*}{System} & \multirow{2}{*}{Method} & (no refinement) & \multicolumn{4}{c}{(with refinement)} \\
                        &                         & TS energy (eV)    & Method iters  & Sella energy evals & Total energy evals & TS energy (eV) \\ 
\hline
\multirow{2}{*}{\ce{In2O3}} & NEB                     & 1.2585     & 50          & 66          & 818        & 1.2362 \\ 
                        & INR                     & \textbf{1.2364}     & 50          & 83          & 835        & 1.2362 \\ 
\hline
\multirow{2}{*}{Alanine dipeptide}  & NEB                     & 11.5239    & 500      &  561       & 8063       & 11.0014 \\ 
                        & INR                     & \textbf{0.4544}     & 200         & 44          & 3046       & \textbf{0.4070} \\ 
\hline
\multirow{2}{*}{Methylidyne diffusion} & NEB                     & 2.3539     & 100         & 7           & 1509       & 2.3539 \\ 
                        & INR-GS                  & \textbf{1.3720}     & 200         & 28          & 3030       & \textbf{1.4142} \\ 
\hline
\multirow{2}{*}{\ce{N2 + H2}}   & NEB                     & 0.0533     &  500         & 38          & 7540       & -0.4379 \\ 
                        & INR                     & \textbf{0.4131}     & 200         & 64          & 3066       & \textbf{0.3415} \\ 
\hline
\end{tabular}
\end{table*}

\section{Path Generalization}
\label{sec:general}

\begin{figure*}[t]
  \centering
  \begin{subfigure}[b]{0.287\linewidth}
    \centering
    \includegraphics[width=\linewidth]{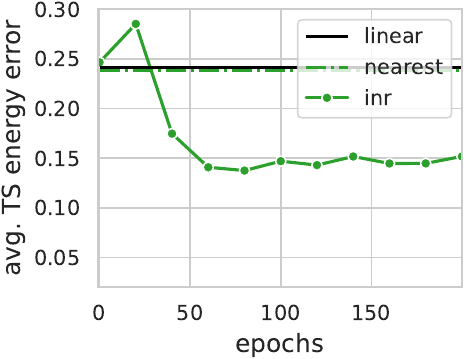}
    \vspace{-15pt}
    \caption{$8$ training systems}
    \label{fig:errors_8}
  \end{subfigure}\hspace{0.5mm}
  \begin{subfigure}[b]{0.287\linewidth}
    \centering
    \includegraphics[width=\linewidth]{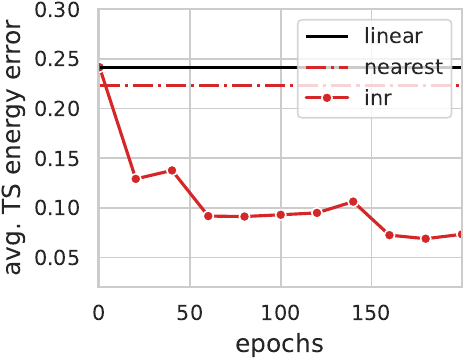}
    \vspace{-15pt}
    \caption{$16$ training systems}
    \label{fig:errors_16}
  \end{subfigure}\hspace{0.5mm}
  \begin{subfigure}[b]{0.287\linewidth}
    \centering
    \includegraphics[width=\linewidth]{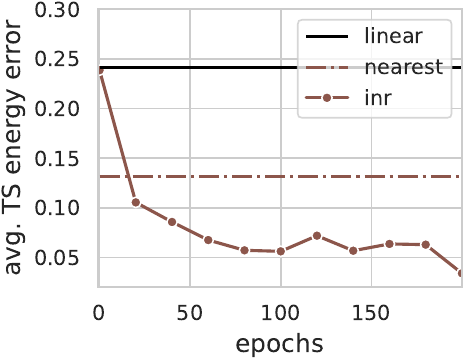}
    \vspace{-15pt}
    \caption{$32$ training systems}
    \label{fig:errors_32}
  \end{subfigure}
  \vspace{-8pt}
  \caption{Average TS energy error on unseen systems vs.~training epochs. The INR improves with more training examples and outperforms the nearest neighbor baseline, demonstrating its ability to generalize to new systems.}
  \label{fig:path_errors}
\vspace{-6pt}
\end{figure*}

\begin{figure*}[t]
  \centering
  \begin{subfigure}[b]{0.9\linewidth}
    \centering
    \includegraphics[width=\linewidth]{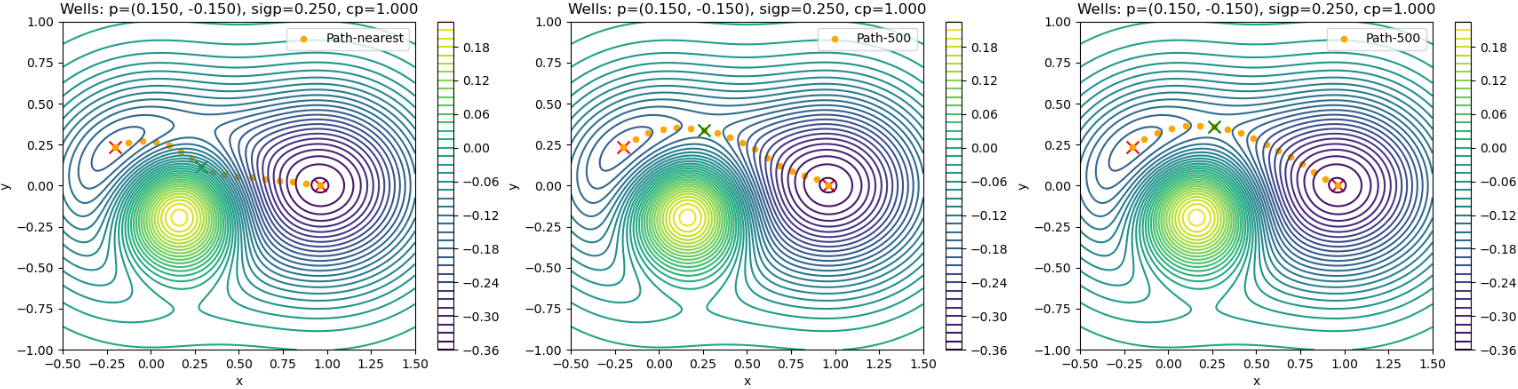}
    \vspace{-15pt}
    \caption{}
  \end{subfigure}
  \\[1mm]
  \begin{subfigure}[b]{0.9\linewidth}
    \centering
    \includegraphics[width=\linewidth]{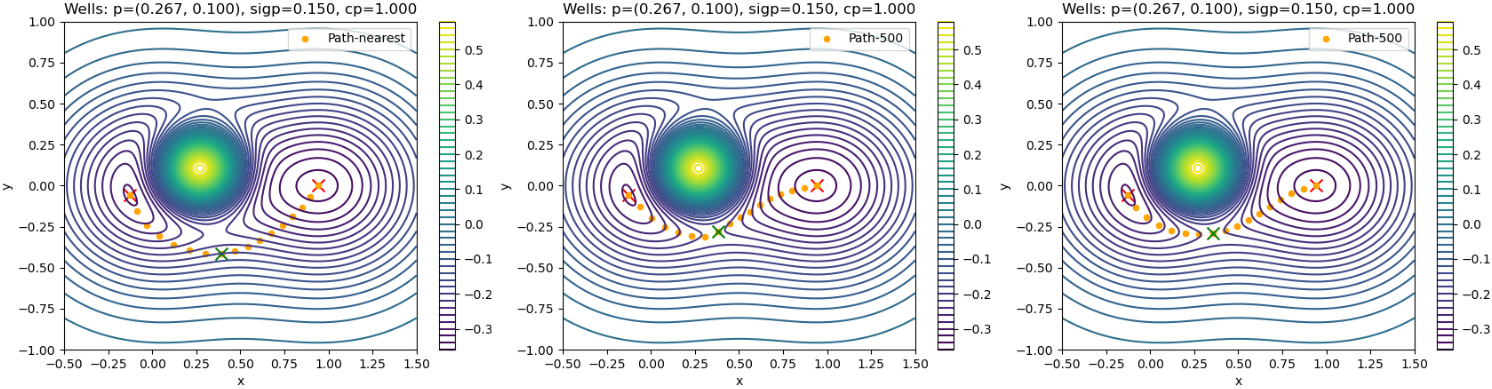}
    \vspace{-15pt}
    \caption{}
  \end{subfigure}
  \vspace{-8pt}
  \caption{Nearest neighbor (left), INR (middle), and true (right) paths for two unseen systems (a--b). The INR learns from existing paths to instantly approximate MEPs for new systems, aligning more closely with the true path compared to the nearest neighbor baseline.}
  \label{fig:paths12}
\vspace{-10pt}
\end{figure*}

Sec.~\ref{sec:exp} showed that neural networks can learn to represent reaction paths when optimized \emph{separately} for individual systems. We now ask the following question: could a single neural network learn from existing MEPs to instantly predict paths for multiple \emph{unseen} systems? This would represent a considerable leap over all existing methods, which start for each system from scratch. Assuming that a dataset of known MEPs for related systems is available, a neural network could be trained on this data to be applied to unseen systems where evaluating the potential energy surface is prohibitively expensive. Such a network must also take as inputs the initial and final states $\{\vec{A}, \vec{B}\}$ and, optionally, parameters $\phi$ characterizing the potential energy of the system. The reaction path would then take the following form:
\begin{equation}
    \vec{x}_\theta(t; \ {\small \vec{A}, \vec{B}, \phi}) = \vec{b}(t; \ {\small \vec{A}, \vec{B}}) + t(1-t)\ \vec{g}_\theta(t; \ {\small \vec{A}, \vec{B}, \phi}).
\end{equation}
The path and TS estimated by the network for an unseen system may serve as a quality initial guess for energy-based optimization. In this paper, we test this \emph{path generalization} idea on a set of two-dimensional systems as a proof of concept, leaving the challenging extension to atomistic systems as future work. Our chosen potential energy surface has the following form, where $\mathcal{N}(x, y;\ \vec{\mu}, \sigma)$ is the normal density with mean $\vec{\mu}$ and standard deviation $\sigma$:
\begin{align}
U_\text{Wells}(x, y) = -&\mathcal{N}(x, y; \ \vec{\mu}=(0, 0), \sigma=0.4)\nonumber \\
                       -&\mathcal{N}(x, y; \ \vec{\mu}=(0, 1), \sigma=0.4)\nonumber \\
                       +c_p &\mathcal{N}(x, y; \ \vec{\mu}=\vec{\mu}_p, \sigma=\sigma_p).
\end{align}
The first two terms are fixed negative wells centered at $(0, 0)$ and $(0, 1)$, and the third term is a variable peak centered at $\vec{\mu}_p$ with a scaling factor $c_p$. We consider the set of systems formed by varying the parameter vector $\phi = (\mu_{px}, \mu_{py}, \sigma_p, c_p)$ over the following range, chosen so the peak has a meaningful effect on the MEP: $\mu_{px}\in [0.15, 0.85]$, $\mu_{py}\in [-0.15, 0.15]$, $\sigma_p\in [0.05, 0.25]$, and $c_p\in [0.2, 1.0]$. For each system, the endpoints $\vec{A}$ and $\vec{B}$ are the local minima of the potential near the wells $(0, 0)$ and $(0, 1)$, respectively.

To create training and testing datasets, we first form a grid by discretizing each dimension of $\phi$ into seven equidistant values and then randomly sample $n_\text{train}$ and $n_\text{test}$ disjoint systems from this grid. We use the L-BFGS optimizer for each system to locate the two endpoints, followed by climbing NEB with the FIRE optimizer to compute the MEP (see Sec.~\ref{subsec:implementation} for details). The neural network $\vec{g}_\theta(t; \ {\small \vec{A}, \vec{B}, \phi})$ inherits the architecture from Sec.~\ref{subsec:implementation}, with the addition of the system-specific inputs, and is trained to minimize the error between the predicted and actual paths (discretized) for each system. We use the following loss function at iteration $k$:
\begin{equation}
L_\text{Path}^k(\theta) = \mathbb{E}_t \left[\|\vec{x}_\theta(t; \ {\small \vec{A}^k, \vec{B}^k, \phi^k}) - \vec{x}^k(t)\|^2\right],
\end{equation}
where $(\vec{A}^k, \vec{B}^k, \phi^k)$ specifies the $k$th training system, $\vec{x}^k(t)$ is the true (NEB) path, and $\mathbb{E}_t$ is the average over equidistant samples $t_i = i/n$, $i\in\{0, 1, \cdots, n\}$. We train the network for $200$ epochs using the Adam optimizer with a $5e{-4}$ learning rate. Our evaluation metric is the average TS energy error over the test systems.

Fig.~\ref{fig:path_errors} plots the test error vs.~epochs for INRs trained on $n_\text{train} = 8$, $16$, and $32$ systems (with $n_\text{test} = 100$). In each case, we compare with two baselines: (i) linear interpolation, which predicts a straight line connecting the endpoints of the test system, and (ii) nearest neighbor, which predicts a path by transforming the MEP of the training system having parameters closest to those of the test system. Specifically, for a test system specified by $(\vec{A}, \vec{B}, \phi)$, we first identify the \emph{nearest} training system $n = \arg\min_j \|\phi^j - \phi\|$ and its MEP $\vec{x}^n(t)$ and then affine-transform this path so its endpoints match $(\vec{A}, \vec{B})$,
\begin{equation}
\tilde{x}(t) = \vec{A} + \frac{\|\vec{B}\ -\ \vec{A}\|}{\|\vec{B}^n - \vec{A}^n\|}\ \bm{R}\cdot\left(\vec{x}^n(t) - \vec{A}^n\right),
\end{equation}
where $\bm{R}$ is the rotation matrix that aligns the direction $(\vec{B}^n - \vec{A}^n)$ with $(\vec{B} - \vec{A})$. Note that $\tilde x(0) = \vec{A}$ and $\tilde{x}(1) = \vec{B}$ by construction. We evaluate this baseline to ensure that the network's apparent generalization is not simply a result of memorizing the training data. From Fig.~\ref{fig:path_errors}, we see that the INR outperforms both baselines at locating the TS, generalizing to new systems despite being trained on limited data (e.g., when $n_\text{train} = 16$). While the INR's advantage over the nearest neighbor baseline reduces when provided with $32$ vs.~$16$ training systems in our low-dimensional setting, we anticipate a persistent improvement in high-dimensional, atomistic settings, where simple parameterizations of systems will not be feasible. Future work should investigate the potential benefits of neural networks in such scenarios and whether learning a universal path representation is possible. Fig.~\ref{fig:paths12} shows two example predicted paths on unseen systems (for $n_\text{train} = 32$) along with the nearest neighbor and true paths, indicating that the INR learns to produce a good approximation of the MEP. More examples are in Appendix~\ref{app:sec:general}.

\section{Conclusion}
We presented a method that optimizes neural networks to smoothly represent reaction paths while permitting direct estimation of the transition state. Through experiments on challenging material and molecular systems, we demonstrated that it overcomes some key limitations of Nudged Elastic Band (NEB). In particular, our approach (i) shows resilience to unnatural states arising from poor initial guesses, (ii) is easily adapted to escape local-minimum solutions, and (iii) automatically learns a complex multi-step reaction path with no manual supervision. We also showed that by conditioning the network on initial and final states, it has the potential to learn from existing paths and generalize to multiple unseen reactions, paving the way for a generalizable reaction path representation. 

We end with some suggestions for future work. First, since our approach allows flexibility in the loss function, base path, and sampling strategy for optimization, different choices guided by chemical intuition may further improve performance. This equally applies to inductive biases in the neural network architecture. For example, instead of predicting the Cartesian coordinates of atoms, the network could be designed to operate in the internal coordinate\cite{ics} space. Second, while we used the same number of samples at training and inference to compare with NEB directly, one might improve efficiency by amortizing training with fewer, randomly sampled inputs and using stochastic gradient descent, following standard practice in deep learning. On a related note, in our growing sampling technique, the sampled region expands according to a pre-defined iteration count, limiting a full coverage of the input space to the last few iterations. This explains the slow initial progress in Fig.~\ref{fig:sine_E_error}, suggesting room for improvement. Third, it would be helpful to understand the effect of the choice of optimizer to train the network. While we used Adam, a standard optimizer for neural networks, integrating ideas from physics-based\cite{fire} or ODE solver-based\cite{precon} optimizers is an interesting alternative. Finally, training networks that can directly predict paths and transition states for unseen systems in high-dimensional, atomistic settings could be an impactful direction for future work. While our preliminary evidence supports this idea, building a practical, generalizable reaction path predictor would require both a large-scale dataset\cite{transition1x} and an efficient model design that supports variable-sized outputs. Incorporating known symmetries into the network architecture through group equivariant graph neural networks\cite{egnn,mace} represents a promising direction toward a scalable and universal reaction path representation.

\begin{acknowledgments}
\noindent
KR is supported by the EPSRC Center for Doctoral Training in Autonomous Intelligent Machines and Systems EP/S024050/1. LLS is supported by the EPSRC Center for Doctoral Training in Chemical Synthesis EP/S024220/1 and Wolfson College, University of Cambridge.
\end{acknowledgments}

\section*{Data Availability Statement}
\noindent
The data that support the findings of this study are available from the corresponding author upon reasonable request.

\appendix


\section{Two-dimensional potentials}
\label{app:sec:2d}

\subsection{LEPS}
\vspace{-16pt}
\begin{equation}
U_\text{LEPS}(r_{ab}, r_{bc}) = E_\text{Coulomb} - E_\text{exchange},
\end{equation}
where
\begin{equation}
E_\text{Coulomb} = \frac{q_{ab}}{1+a} + \frac{q_{bc}}{1+b} + \frac{q_{ac}}{1+c}\nonumber
\end{equation}
and 
\begin{align}
E_\text{exchange}^2 &= \frac{j_{ab}^2}{(1+a)^2} + \frac{j_{bc}^2}{(1+b)^2} + \frac{j_{ac}^2}{(1+c)^2}\nonumber \\
&-\frac{j_{ab} j_{bc}}{(1+a)(1+b)} - \frac{j_{bc} j_{ac}}{(1+b)(1+c)} - \frac{j_{ab} j_{ac}}{(1+a)(1+c)}.\nonumber
\end{align}
Here, $r_{ac} = r_{ab} + r_{bc}$, and the functions $q = Q(r, d)$ and $j = J(r, d)$ are defined as
\begin{align}
Q(r, d) &= \frac{d}{2}\left(\frac{3}{2}e^{-2\alpha_0 (r-r_0)} - e^{-\alpha_0 (r-r_0)}\right),\nonumber \\
J(r, d) &= \frac{d}{4}\left(e^{-2\alpha_0 (r-r_0)} -6 e^{-\alpha_0 (r-r_0)}\right),\nonumber
\end{align}
with parameters $a=0.05$, $b=0.3$, $c=0.05$, $d_{ab}=4.746$, $d_{bc}=4.746$, $d_{ac}=3.445$, $r_0 = 0.742$, and $\alpha_0=1.942$. The initial and final minima are $(0.75, 4.0)$ and $(4.0, 0.75)$, respectively.

\begin{figure}[t]
  \centering
  \begin{subfigure}[b]{0.5\linewidth}
    \centering
    \includegraphics[width=\linewidth]{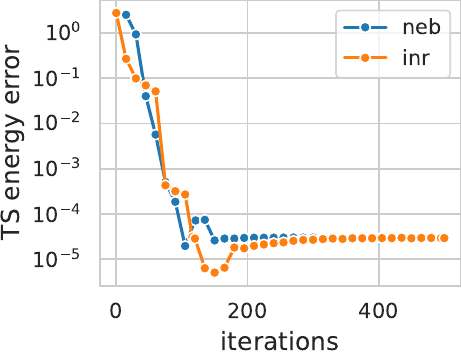}
  \end{subfigure}
  \vspace{-8pt}
  \caption{Training progress for the standard LEPS potential. Both methods approach the TS at a similar rate.}
  \label{fig:leps_progress}
\end{figure}


\vspace{-8pt}
\subsection{M\"uller--Brown}
\vspace{-16pt}
\begin{equation}
U_\text{MB}(x, y) = \sum_{i=1}^4 A_i\ e^{a_i(x-x_{0,i})^2 + b_i(x-x_{0,i})(y-y_{0,i}) + c_i(y-y_{0,i})^2},
\end{equation}
with parameters 
\begin{align}
a &= (-1, -1, -6.5, 0.7),\nonumber \\
b &= (0, 0, 11, 0.6),\nonumber \\
c &= (-10, -10, -6.5, 0.7),\nonumber \\
x_0 &= (1, 0, -0.5, -1),\nonumber \\
y_0 &= (0, 0.5, 1.5, 1),\nonumber \\
A &= (-200, -100, -170, 15).\nonumber
\end{align}
The initial and final minima are $(-0.5582, 1.4417)$ and $(0.6235, 0.0280)$, respectively.

\section{Hyperparameters}
\label{app:sec:hyper}

\begin{figure}[t]
  \centering
  \begin{subfigure}[b]{0.45\linewidth}
    \centering
    \includegraphics[width=\linewidth]{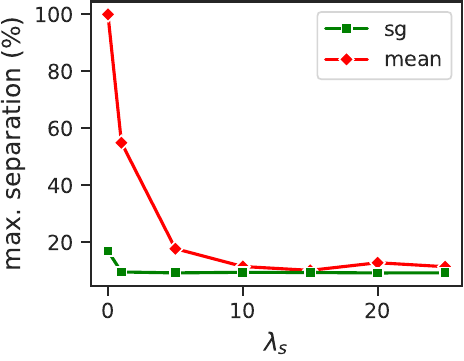}
    \vspace{-15pt}
    \caption{Sample separation vs.~$\lambda_s$}
    \label{fig:separation_vs_S}
  \end{subfigure}\hspace{0.5mm}
  \begin{subfigure}[b]{0.45\linewidth}
    \centering
    \includegraphics[width=\linewidth]{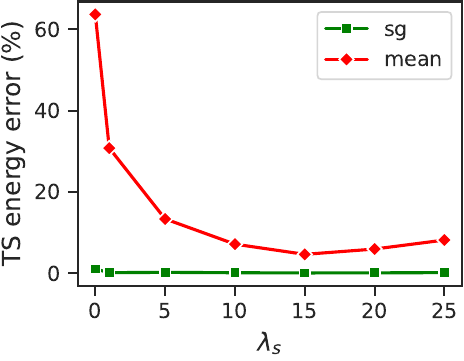}
    \vspace{-15pt}
    \caption{TS energy error vs.~$\lambda_s$}
    \label{fig:E_error_vs_S}
  \end{subfigure}
  \\[0.5mm]
  \begin{subfigure}[b]{0.45\linewidth}
    \centering
    \includegraphics[width=\linewidth]{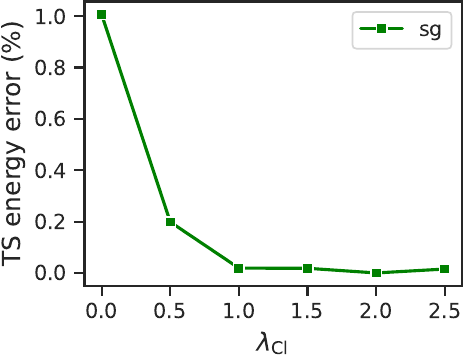}
    \vspace{-15pt}
    \caption{TS energy error vs.~$\lambda_\text{Cl}$}
    \label{fig:E_error_vs_C}
  \end{subfigure}
  \vspace{-8pt}
  \caption{Loss function and hyperparameter analysis. The stop gradient loss is much less sensitive to $\lambda_s$ with respect to sample spacing (a) and TS estimation (b). The climbing loss improves accuracy (c).}
  \label{fig:loss_hyper}
\end{figure}

To better understand the effect of the loss function and hyperparameters $(\lambda_s, \lambda_\text{Cl})$ on the path learned by a smooth representation like the INR, we run some experiments on the M\"uller--Brown potential. 

First, we compare the mean-energy loss function (Eq.~\ref{eq:naive}), which retains the full energy gradient, against its stop-gradient counterpart (Eq.~\ref{eq:updated}), which discards tangential components. Specifically, we calculate the largest separation between adjacent sampled points $\max_i \| \vec{x}_\theta(t_{i+1}) - \vec{x}_\theta(t_i)\|$ along the paths resulting from either loss. We plot this as a percentage of the overall displacement $\| \vec{B} - \vec{A}\|$ in Fig.~\ref{fig:separation_vs_S}. The mean loss (``mean'') incurs a considerable separation for low values of the spacing loss coefficient $\lambda_s$. For $\lambda_s = 0$, the separation is at $100\%$, meaning that all samples have departed from the high energy interior, settling at one of the two end minima $(\vec{A}, \vec{B})$. While less severe at larger $\lambda_s$, this phenomenon stifles sampling-based learning and produces incorrect TS estimates, as evident from the TS energy error plot in Fig.~\ref{fig:E_error_vs_S}. Moreover, $\lambda_s$ must be re-tuned to neutralize tangential gradients for a different potential energy surface. In contrast, by discarding tangential components, the stop gradient loss (``sg'') is significantly less sensitive to $\lambda_s$, estimating the TS reliably. We next examine the climbing loss coefficient $\lambda_\text{Cl}$ in Eq.~\ref{eq:cl} by plotting the TS energy error (with $\lambda_s = 0$) in Fig~\ref{fig:E_error_vs_C}. Setting $\lambda_\text{Cl}$ to around $1.0$ or higher improves the accuracy of the TS while incurring no additional cost in the optimization. Thus, these simple experiments demonstrate the benefit of incorporating nudging and climbing mechanisms into the continuous representation.

\vspace{-16pt}
\textcolor{black}{
\section{Theoretically justifying the modified energy $\tilde{U}$}
\label{app:sec:theory}
To understand the effect of the stop gradient in theory, we analyze the loss functions involving $\tilde{U}(\vec{x})$ vs.~$U(\vec{x})$ using the Euler--Lagrange theorem, which we informally state here for completeness.
\begin{quote}
\underline{Theorem (Euler--Lagrange)}: Let the functional
\[
    S[\vec{x}(\cdot)] = \int_a^b \mathcal{L}\left(t,\ \vec{x}(t),\ \vec{x}\ '(t)\right)\ dt
\]
be defined for smooth functions $\vec{x}(\cdot): [a, b]\rightarrow \mathbb{R}^d$ satisfying $\vec{x}(a) = \vec{A}$ and $\vec{x}(b) = \vec{B}$. If $\vec{x}_{*}(t)$ is a minimizer of $S$, then 
\[
    \delta S\left(\vec{x}_{*}(\cdot)\right)(t) = 0\quad \forall t\in [a, b],
\]
where $\delta S$ is the functional derivative of $S$,
\[
    \delta S\left(\vec{x}(\cdot)\right) = \left. \frac{\partial\mathcal{L}}{\partial \vec{x}} - \frac{d}{dt}\left(\frac{\partial\mathcal{L}}{\partial\vec{x}\ '}\right) \right|_{\vec{x}(\cdot)}.
\]
\end{quote}
\noindent
In our case, assuming uniform sampling and no spacing or climbing terms ($\lambda_s = \lambda_\text{Cl} = 0$), the stop gradient loss functional (Eq.~\ref{eq:updated}) is
\[
    S_{\tilde{U}}[\vec{x}(\cdot)] = \int_0^1 \tilde{U}(\vec{x}(t))\ dt,
\]
with functional derivative equal to the orthogonal energy gradient at $\vec{x}(t)$ (as shown in Eq.~\ref{eq:discard}),
\[
    \delta S_{\tilde{U}}\left(\vec{x}(\cdot)\right)(t) = \nabla_x \tilde{U}(\vec{x}(t)) = \nabla_x U(\vec{x}(t))_{\perp}.
\]
This implies that the minimizer $\vec{x}_{*}(\cdot)$ satisfies $\nabla_x U(\vec{x}_{*}(t))_{\perp} = 0\ \ \forall t\in [0, 1]$, which is precisely the definition of the MEP. Moreover, since $\delta S_{\tilde{U}}$ is always orthogonal to the current path, analogous to NEB, gradient-based optimization nudges the path at each update (see Fig.~\ref{fig:larsx2_evolve} for a visual example).
}

\textcolor{black}{
On the other hand, the true energy loss functional (Eq.~\ref{eq:naive}) is 
\[
    S_{U}[\vec{x}(\cdot)] = \int_0^1 U(\vec{x}(t))\ dt,
\]
with functional derivative equal to the full energy gradient at $\vec{x}(t)$,
\[
    \delta S_{U}\left(\vec{x}(\cdot)\right)(t) = \nabla_x U(\vec{x}(t)).
\]
If a minimizer $\vec{x}_{*}(\cdot)$ exists, it must satisfy $\nabla_x U(\vec{x}_{*}(t)) = 0\ \ \forall t\in [0, 1]$, which is not possible unless there is a constant-energy path connecting the endpoints $\vec{A}$ and $\vec{B}$. Hence, while $S_{U}$ does not have a minimizing path in general, it is easy in practice to \emph{nearly} satisfy the zero-gradient condition by ``wasting time'' at local minima of $U$. An example path with this property (that wastes time at the endpoints) is
\[
    \vec{y}(t) = 
    \begin{cases}
        \vec{A} & \text{if}\ \ t \in [0,\ t_1]\\
        \vec{x}_c(t) & \text{if}\ \ t \in [t_1,\ t_1 + \epsilon]\\
        \vec{B} & \text{if}\ \ t \in [t_1 + \epsilon,\ 1],
    \end{cases}
\]
where $t_1 \in (0, 1)$, $0 < \epsilon \ll 1$, and $\vec{x}_c(t)$ is a smooth sub-path connecting $\vec{A}$ and $\vec{B}$. Empirically, a similar path results from optimizing $S_{U}$ for the M\"uller--Brown potential, where we observe all sampled points to eventually settle at either $\vec{A}$ or $\vec{B}$ (see Appendix~\ref{app:sec:hyper} for details). In this case, since $\delta S_U$ includes components tangential to the current path, unlike NEB, gradient-based optimization attempts to avoid higher-energy regions on it at each update.
}

\textcolor{black}{
To summarize, (i) the stop gradient operator produces nudging updates in a continuous setting; (ii) with $\tilde{U}(\vec{x})$, the optimal path, in theory, has no orthogonal energy gradient and is thus an MEP; (iii) with $U(\vec{x})$, the resulting path in practice is generally not an MEP and may even converge to a degenerate solution as above; and (iv) the empirical benefit of $\tilde{U}$ over $U$ (despite balancing the latter objective with a spacing loss term) is further described in Appendix~\ref{app:sec:hyper}.
}

\section{Sine-n systems}
\label{app:sec:larsxn}
\begin{figure}[t]
  \centering
  \begin{subfigure}[b]{0.45\linewidth}
    \centering
    \includegraphics[width=\linewidth]{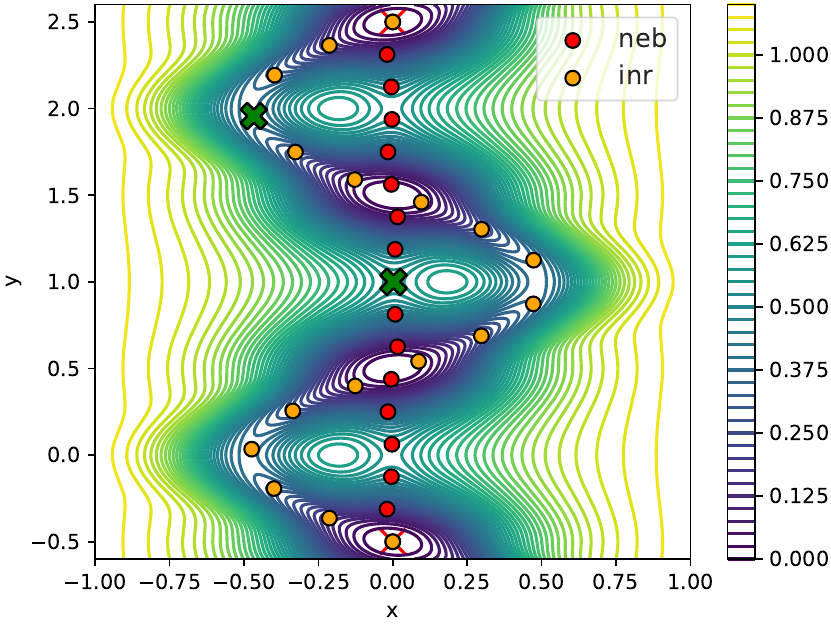}
    \vspace{-15pt}
    \caption{NEB and INR-GS paths}
    \label{fig:larsx3_neb+inr}
  \end{subfigure}\hspace{0.001\linewidth}
  \begin{subfigure}[b]{0.45\linewidth}
    \centering
    \includegraphics[width=\linewidth]{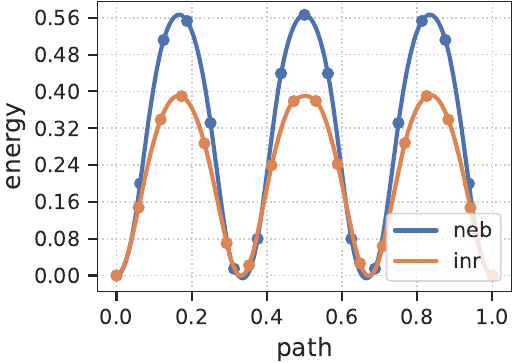}
    \vspace{-15pt}
    \caption{Energy profiles}
    \label{fig:larsx3_E}
  \end{subfigure}
  \vspace{-8pt}
  \caption{Sine-3 system results. NEB gets stuck in the direct path (red). The INR (growing sampling) finds the lower-barrier curved path (orange).}
  \label{fig:larsx3}
\end{figure}

\begin{figure*}
  \centering
  \begin{subfigure}[b]{0.245\linewidth}
    \centering
    \includegraphics[width=\linewidth]{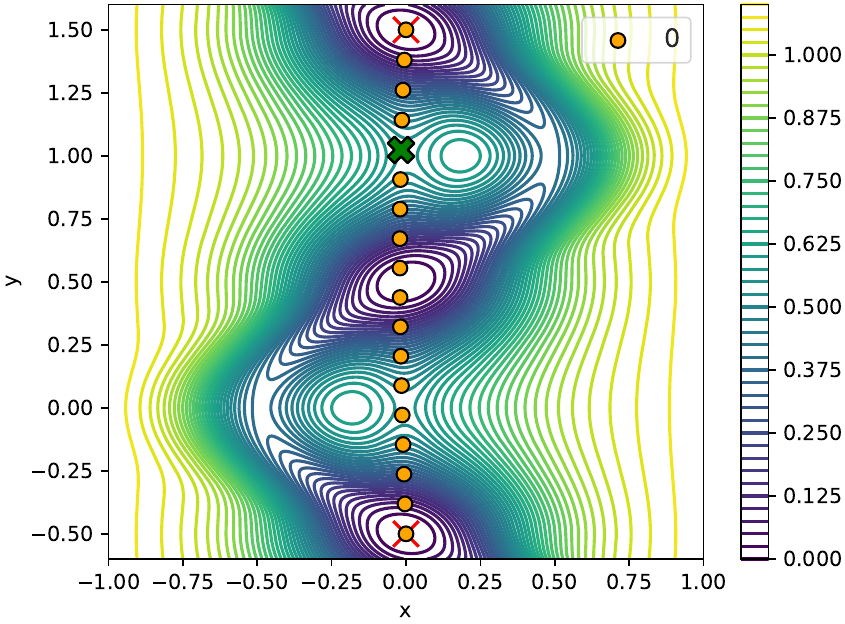}
    \vspace{-15pt}
    \caption{Iteration $0$}
    \label{fig:larsx2_0}
  \end{subfigure}
  \begin{subfigure}[b]{0.245\linewidth}
    \centering
    \includegraphics[width=\linewidth]{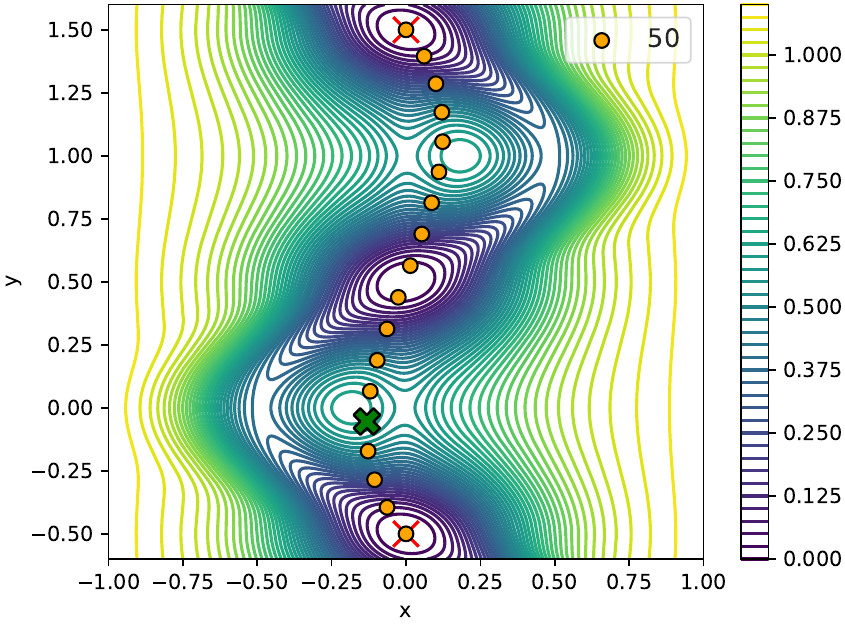}
    \vspace{-15pt}
    \caption{Iteration $50$}
    \label{fig:larsx2_50}
  \end{subfigure}
  \begin{subfigure}[b]{0.245\linewidth}
    \centering
    \includegraphics[width=\linewidth]{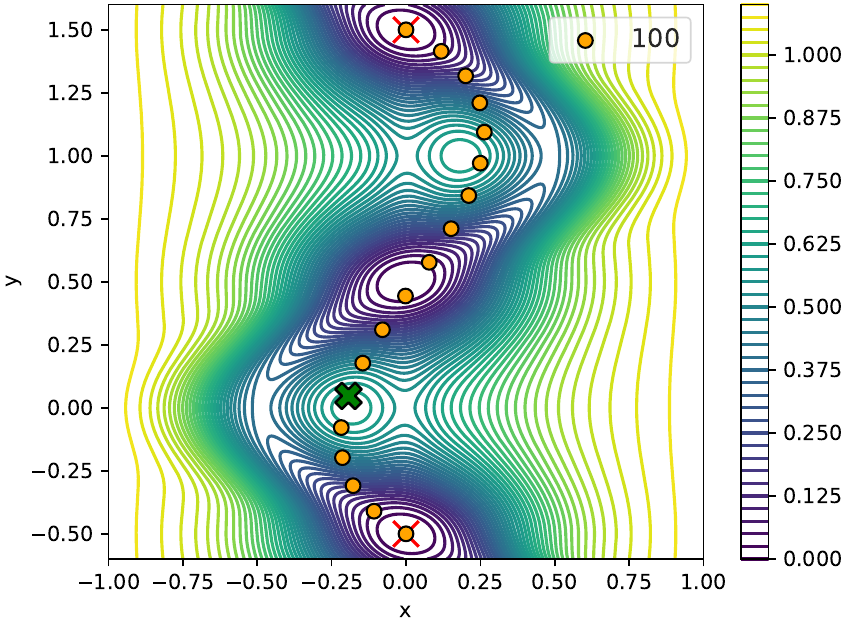}
    \vspace{-15pt}
    \caption{Iteration $100$}
    \label{fig:larsx2_100}
  \end{subfigure}
  \begin{subfigure}[b]{0.245\linewidth}
    \centering
    \includegraphics[width=\linewidth]{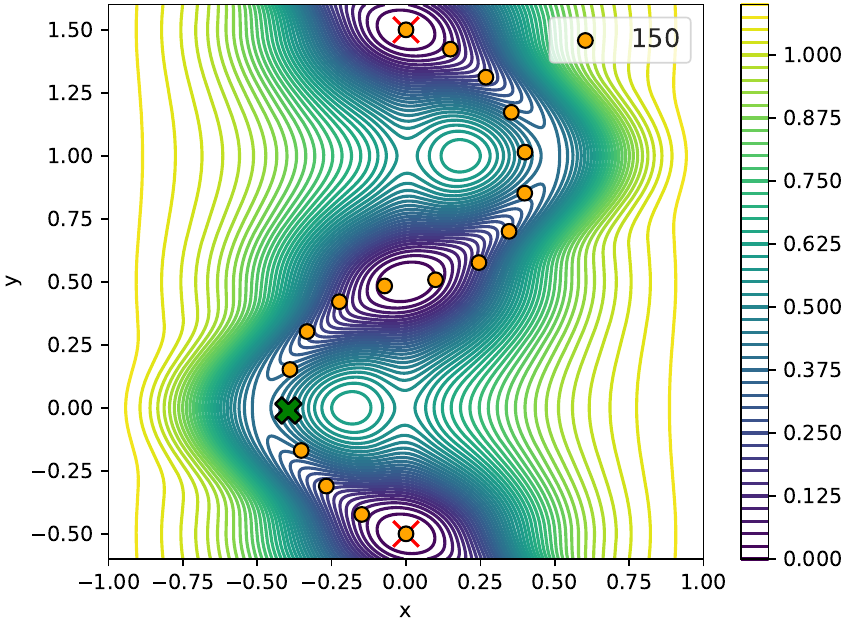}
    \vspace{-15pt}
    \caption{Iteration $150$}
    \label{fig:larsx2_150}
  \end{subfigure}
  \\[0.5mm]
  \begin{subfigure}[b]{0.245\linewidth}
    \centering
    \includegraphics[width=\linewidth]{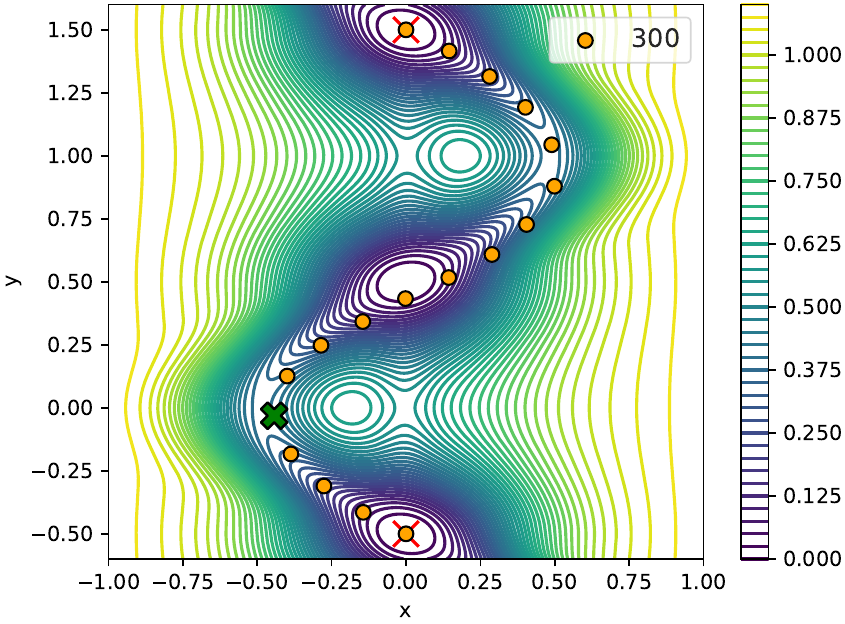}
    \vspace{-15pt}
    \caption{Iteration $300$}
    \label{fig:larsx2_300}
  \end{subfigure}
  \begin{subfigure}[b]{0.245\linewidth}
    \centering
    \includegraphics[width=\linewidth]{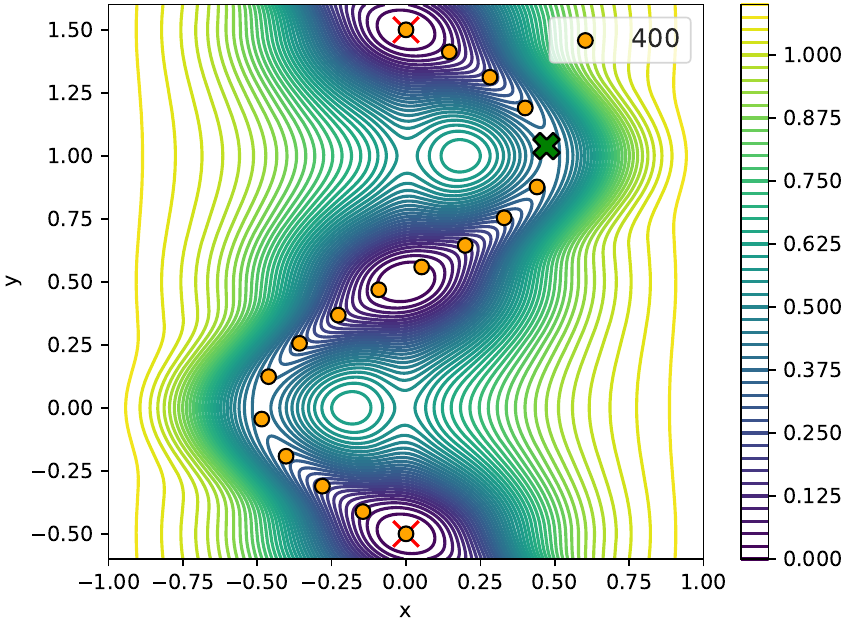}
    \vspace{-15pt}
    \caption{Iteration $400$}
    \label{fig:larsx2_400}
  \end{subfigure}
  \begin{subfigure}[b]{0.245\linewidth}
    \centering
    \includegraphics[width=\linewidth]{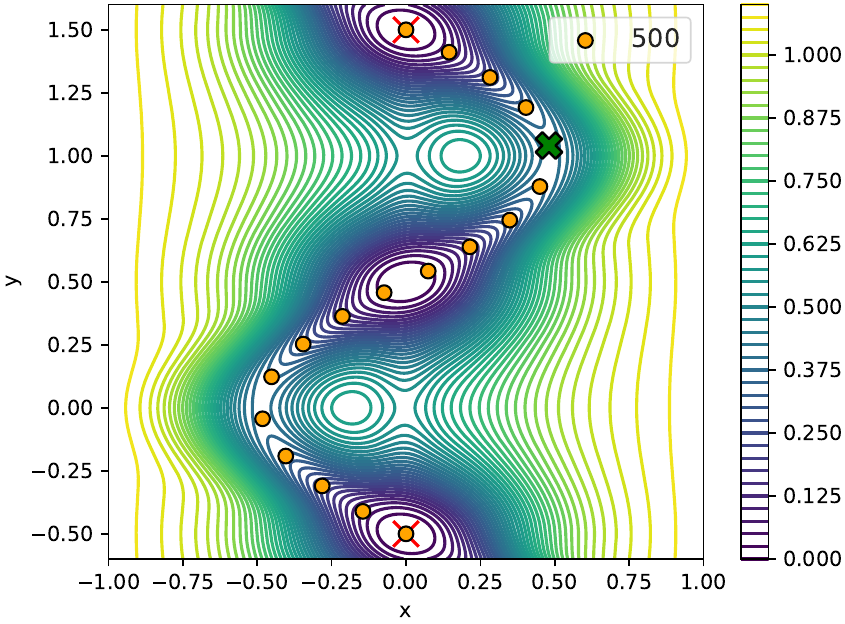}
    \vspace{-15pt}
    \caption{Iteration $500$}
    \label{fig:larsx2_500}
  \end{subfigure}  
  \vspace{-5pt}
  \caption{INR-GS paths at different iterations for the sine-2 system.}
  \label{fig:larsx2_evolve}
\end{figure*}

\noindent
Fig.~\ref{fig:larsx3} shows results for the sine-3 system. Fig.~\ref{fig:larsx2_evolve} shows the evolution of the INR-GS path with iterations for sine-2.

\section{Methylidyne diffusion system}
\label{app:sec:sine}
\begin{figure*}
  \centering
  \begin{subfigure}[t]{\linewidth}
    \centering
    \begin{subfigure}[b]{0.161\linewidth}
      \centering
      \includegraphics[width=\linewidth]{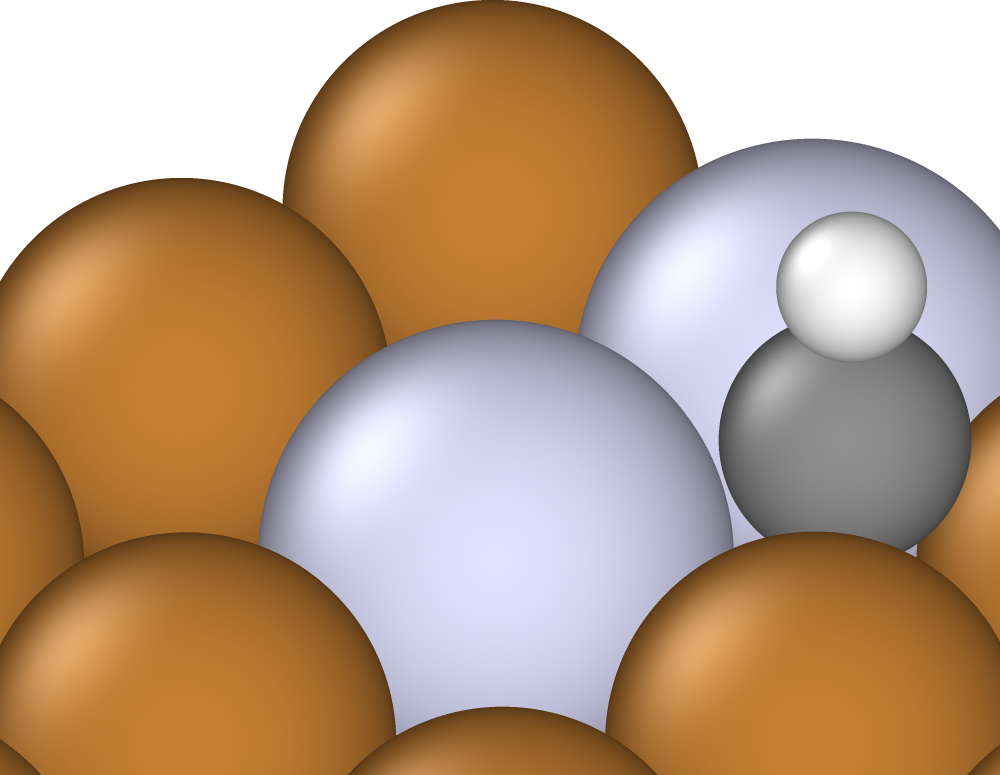}
    \end{subfigure}\hspace{0.001\linewidth}
    \begin{subfigure}[b]{0.161\linewidth}
      \centering
      \includegraphics[width=\linewidth]{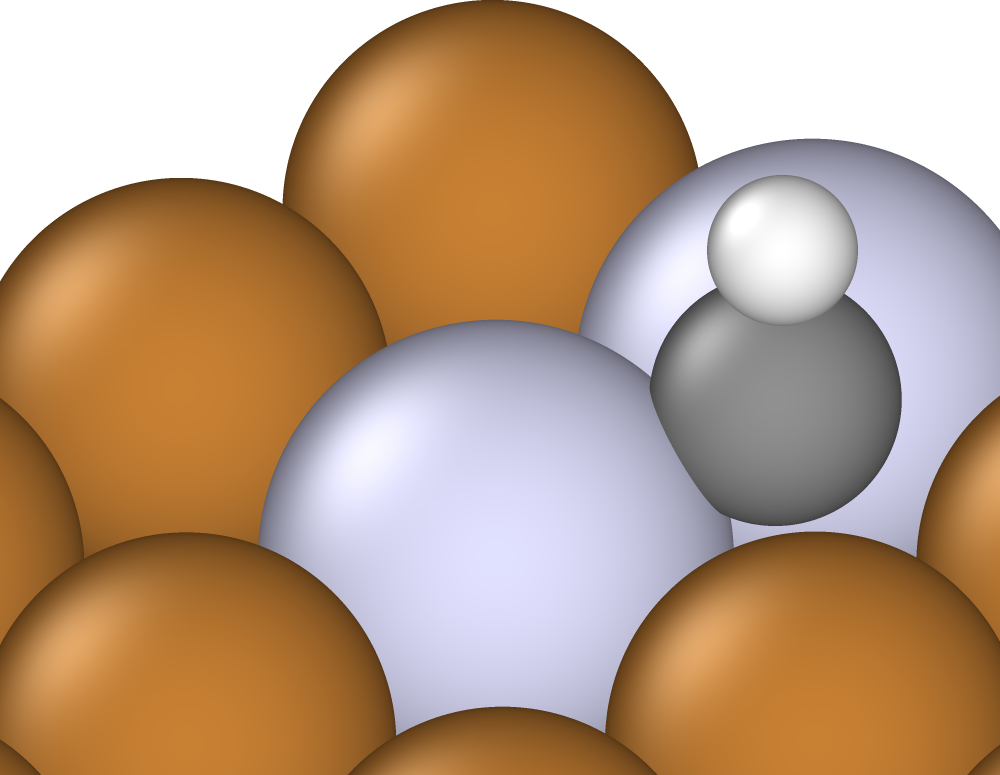}
    \end{subfigure}\hspace{0.001\linewidth}
    \begin{subfigure}[b]{0.161\linewidth}
      \centering
      \includegraphics[width=\linewidth]{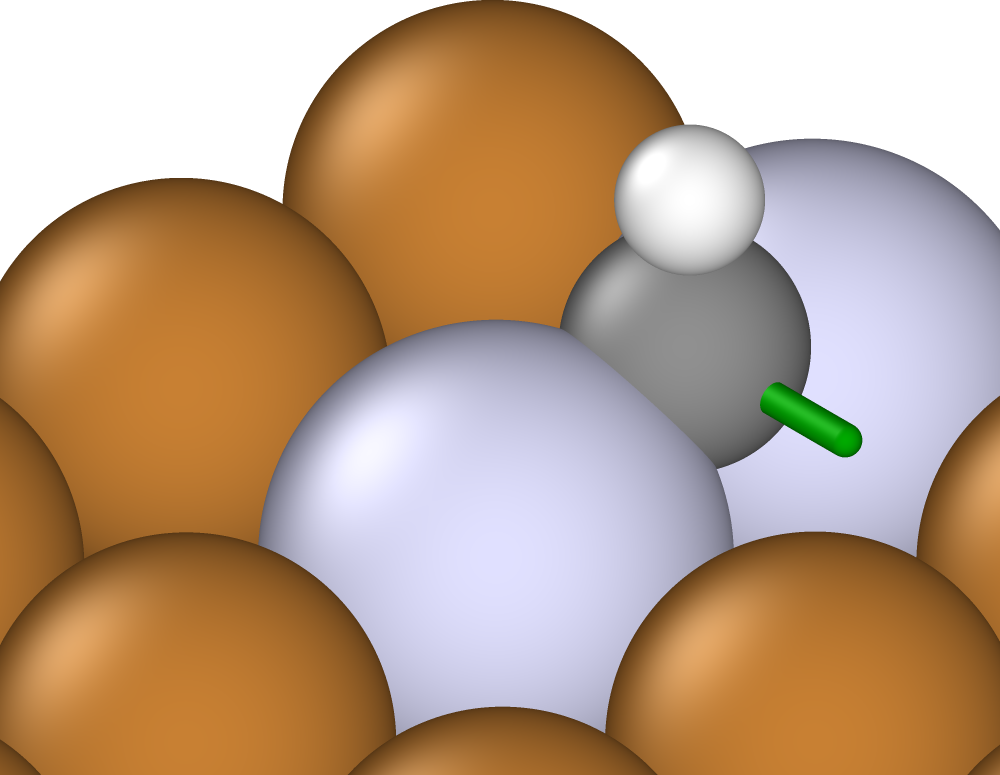}
    \end{subfigure}\hspace{0.001\linewidth}
    \begin{subfigure}[b]{0.161\linewidth}
      \centering
      \includegraphics[width=\linewidth]{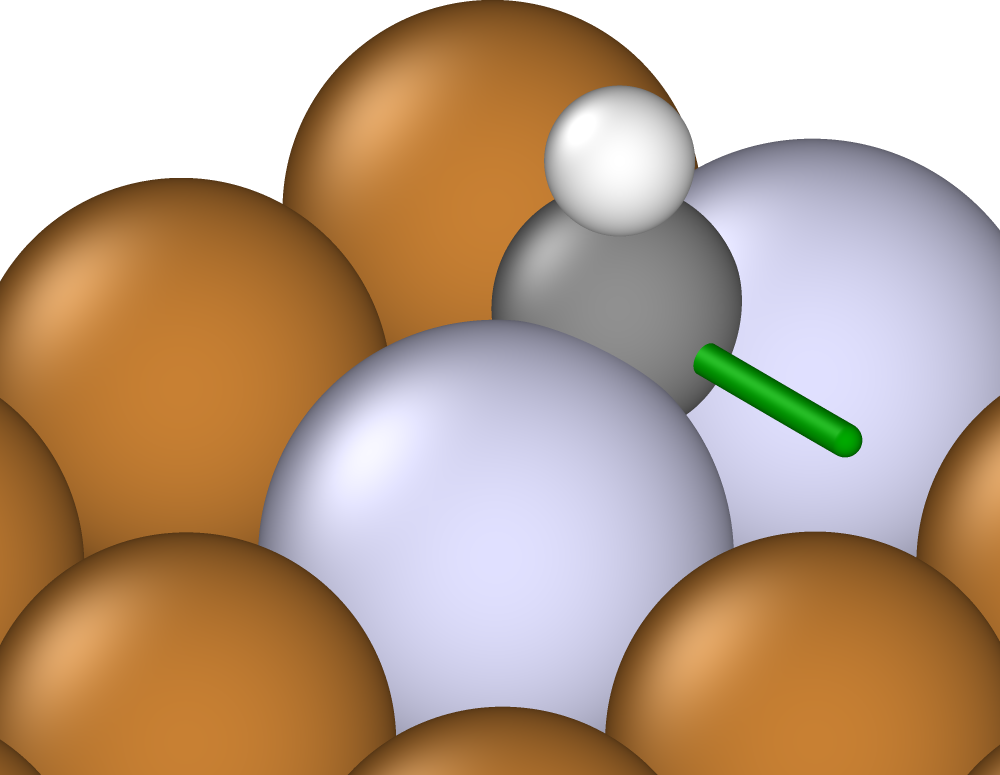}
    \end{subfigure}\hspace{0.001\linewidth}
    \begin{subfigure}[b]{0.161\linewidth}
      \centering
      \includegraphics[width=\linewidth]{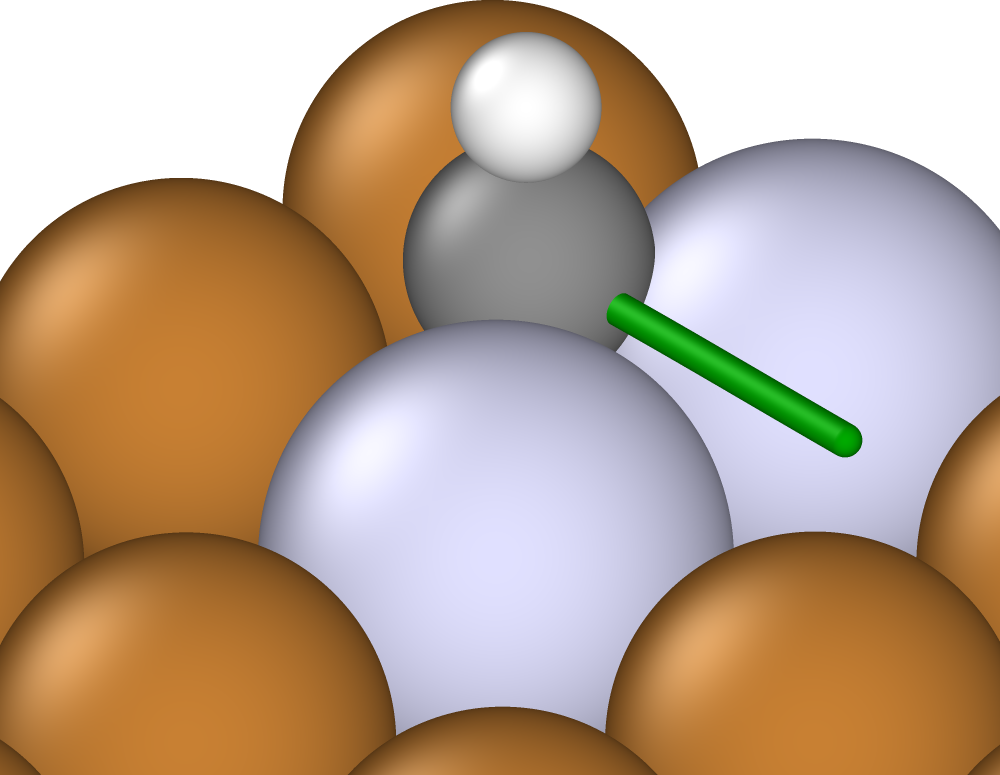}
    \end{subfigure}\hspace{0.001\linewidth}
    \begin{subfigure}[b]{0.161\linewidth}
      \centering
      \includegraphics[width=\linewidth]{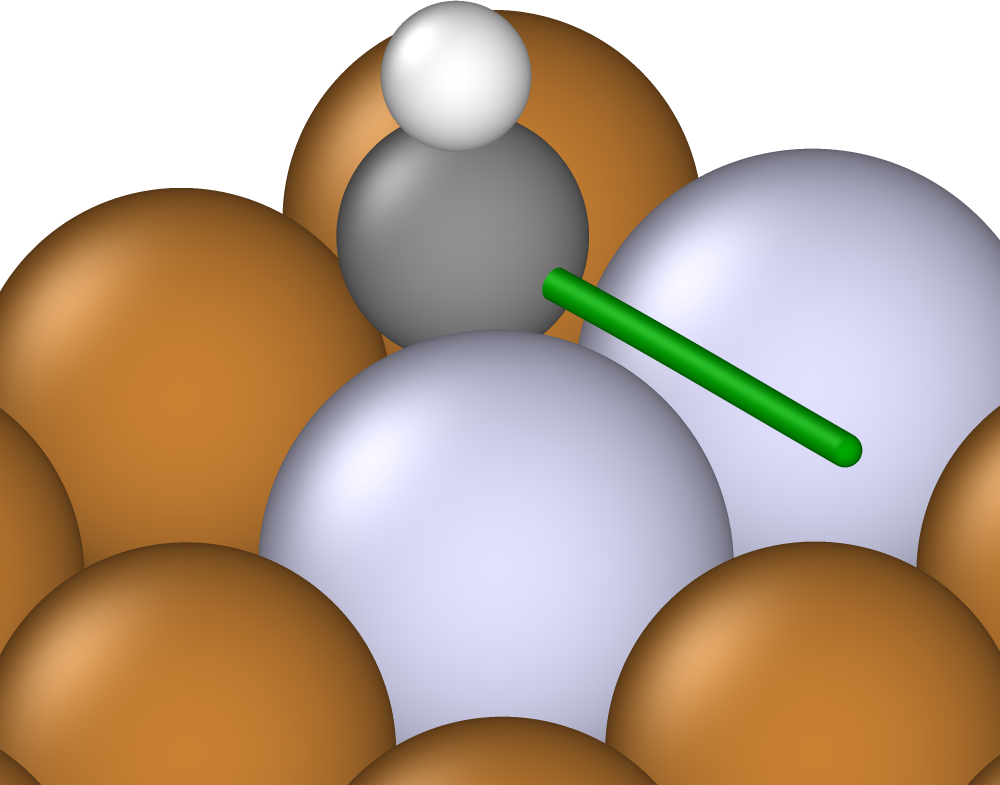}
    \end{subfigure}
    \vspace{-3pt}
    \caption{Iteration $0$}
  \end{subfigure}
  \\[0.5mm]  
  \begin{subfigure}[t]{\linewidth}
    \centering
    \begin{subfigure}[b]{0.161\linewidth}
      \centering
      \includegraphics[width=\linewidth]{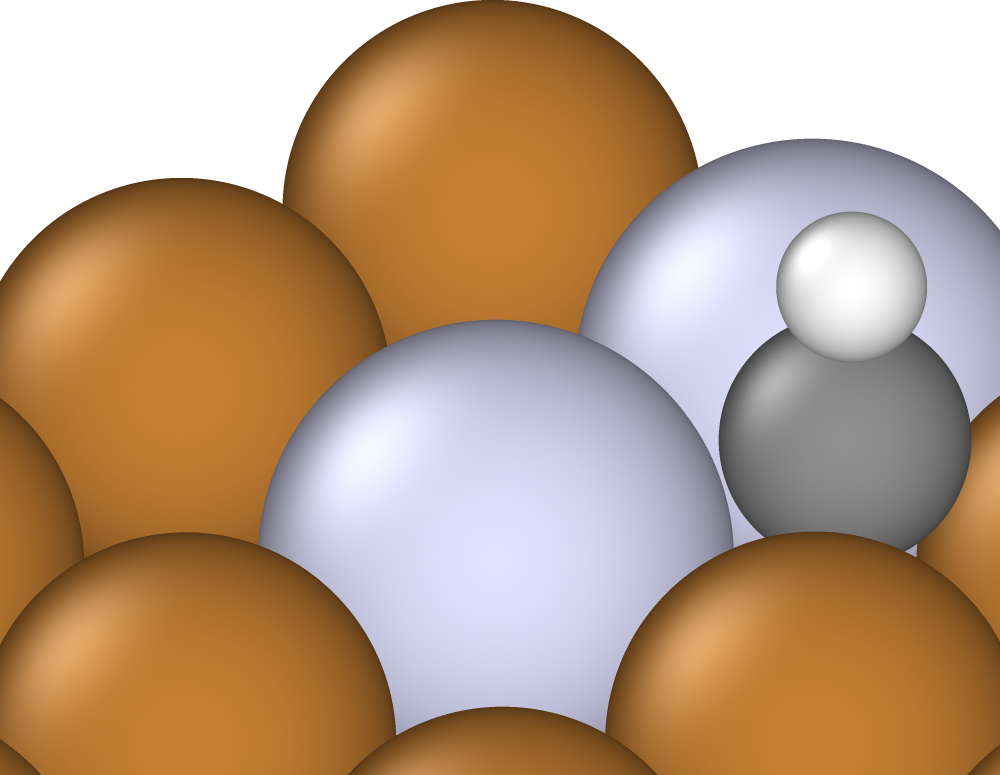}
    \end{subfigure}\hspace{0.001\linewidth}
    \begin{subfigure}[b]{0.161\linewidth}
      \centering
      \includegraphics[width=\linewidth]{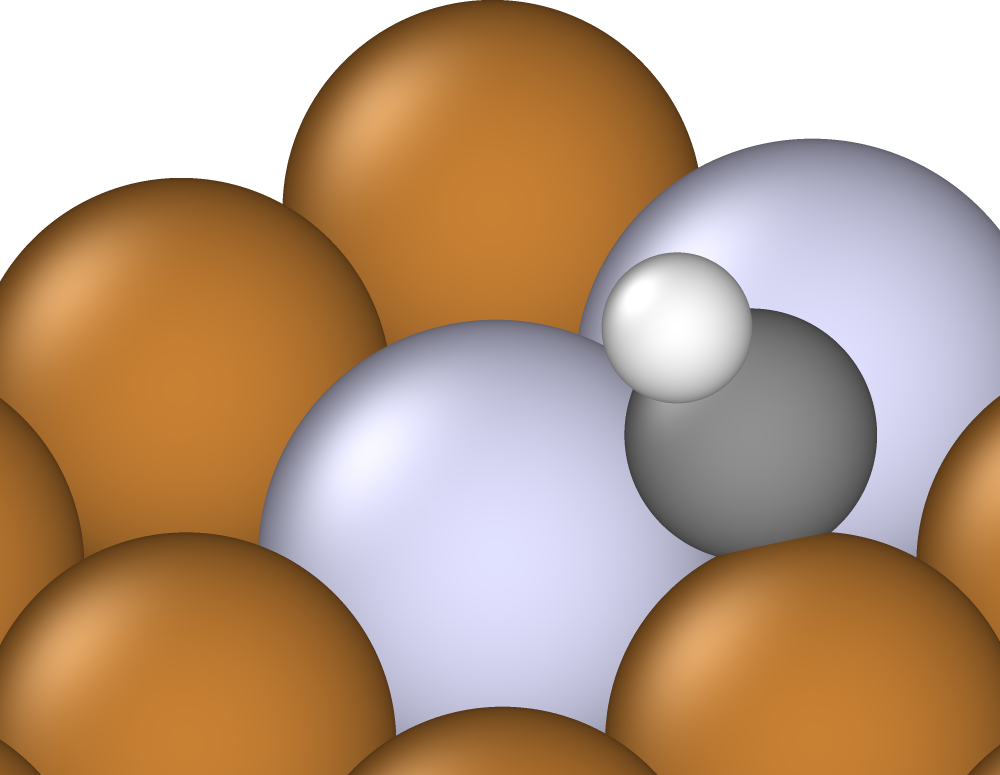}
    \end{subfigure}\hspace{0.001\linewidth}
    \begin{subfigure}[b]{0.161\linewidth}
      \centering
      \includegraphics[width=\linewidth]{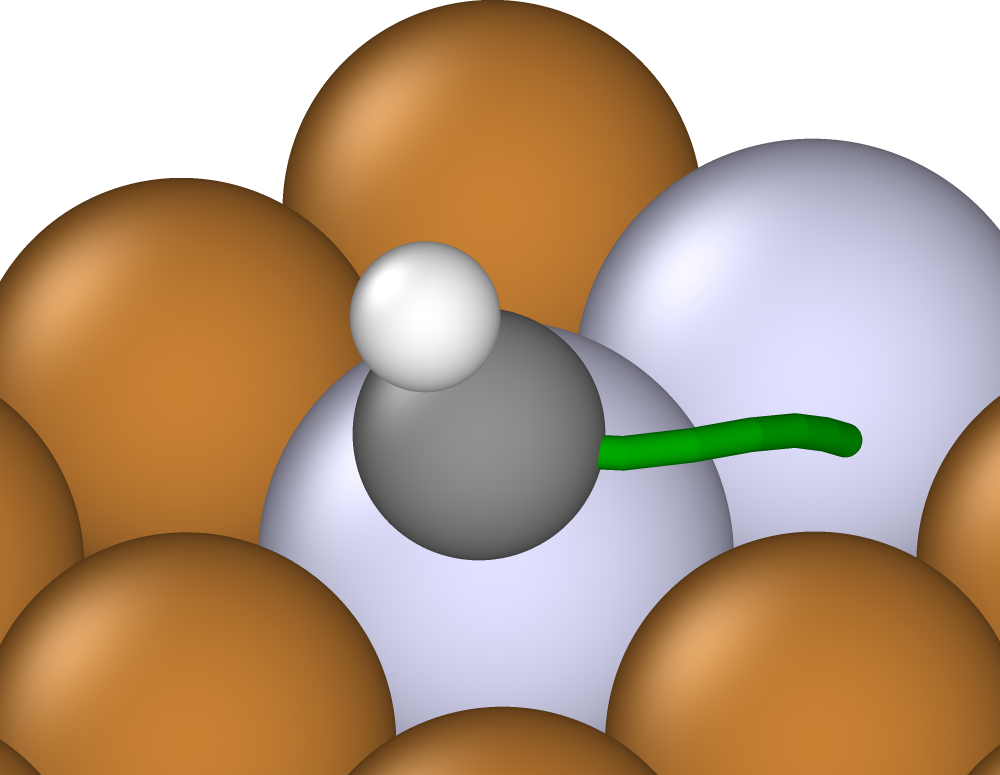}
    \end{subfigure}\hspace{0.001\linewidth}
    \begin{subfigure}[b]{0.161\linewidth}
      \centering
      \includegraphics[width=\linewidth]{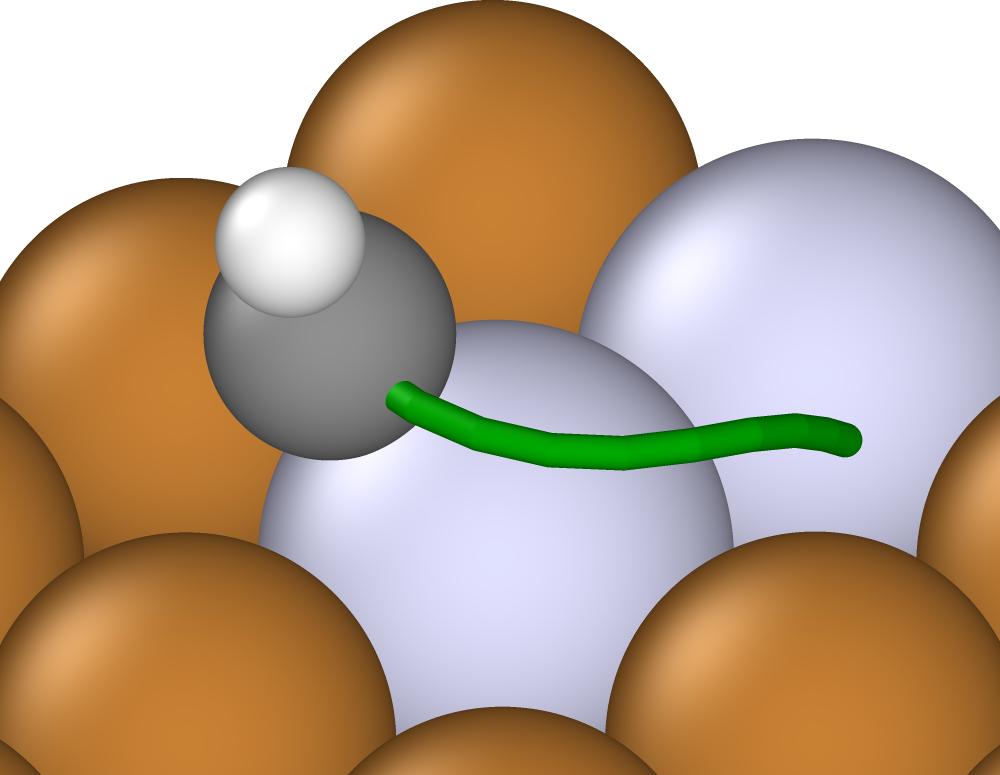}
    \end{subfigure}\hspace{0.001\linewidth}
    \begin{subfigure}[b]{0.161\linewidth}
      \centering
      \includegraphics[width=\linewidth]{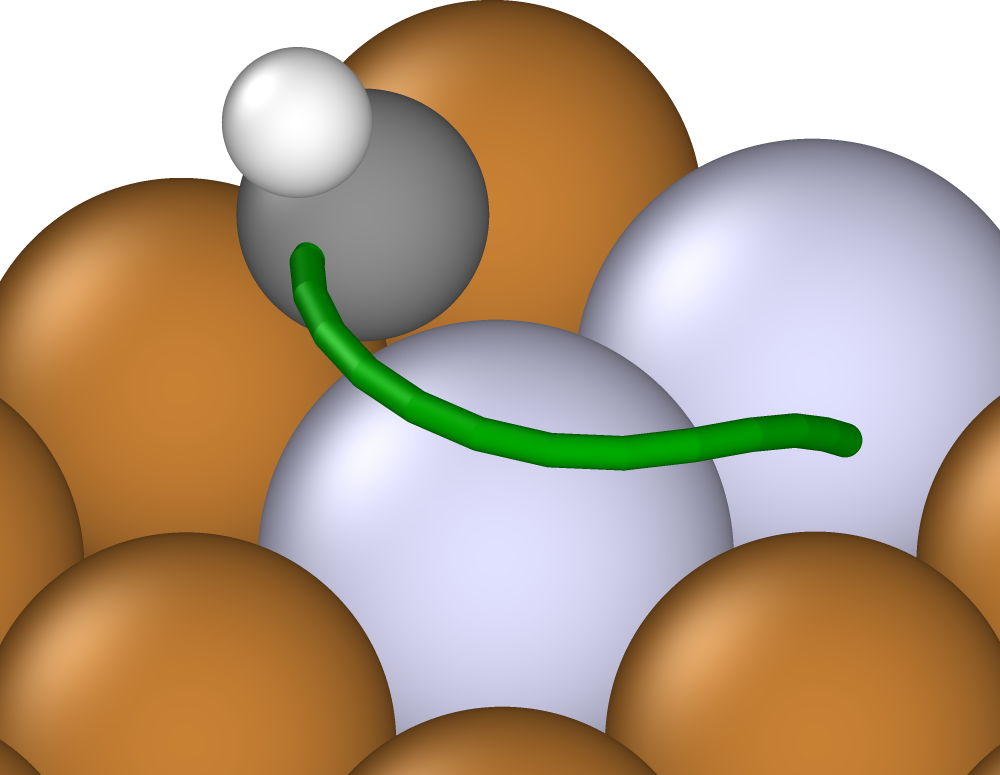}
    \end{subfigure}\hspace{0.001\linewidth}
    \begin{subfigure}[b]{0.161\linewidth}
      \centering
      \includegraphics[width=\linewidth]{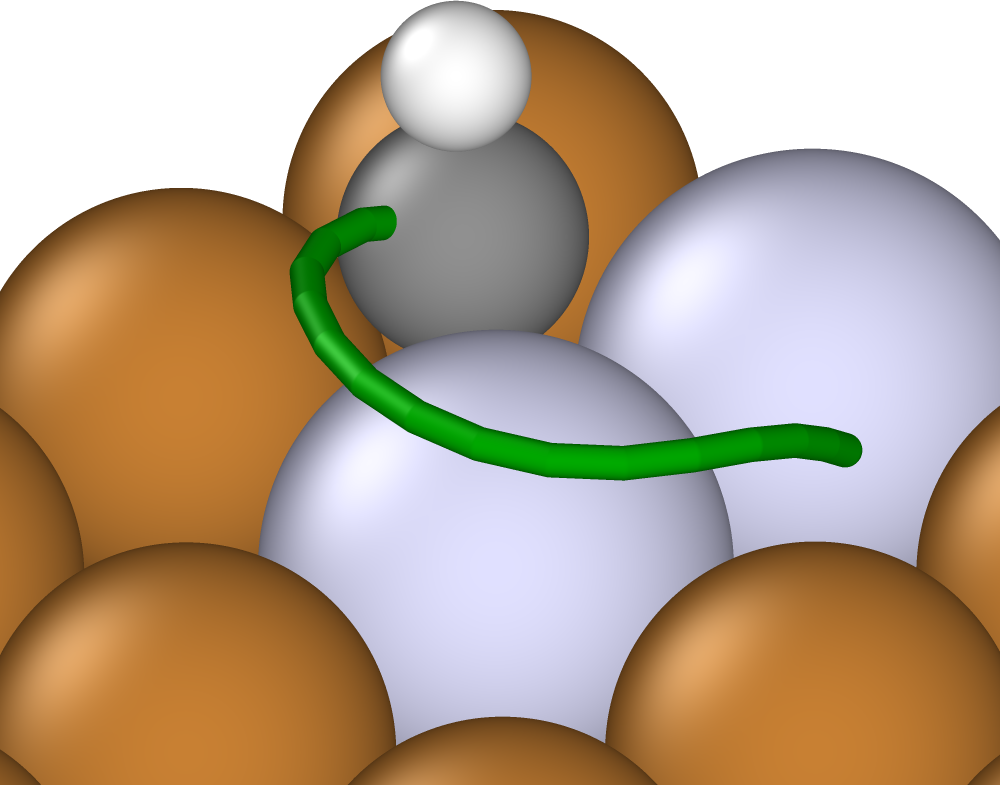}
    \end{subfigure}
    \vspace{-3pt}
    \caption{Iteration $250$}
  \end{subfigure}
  \\[0.5mm]  
  \begin{subfigure}[t]{\linewidth}
    \centering
    \begin{subfigure}[b]{0.161\linewidth}
      \centering
      \includegraphics[width=\linewidth]{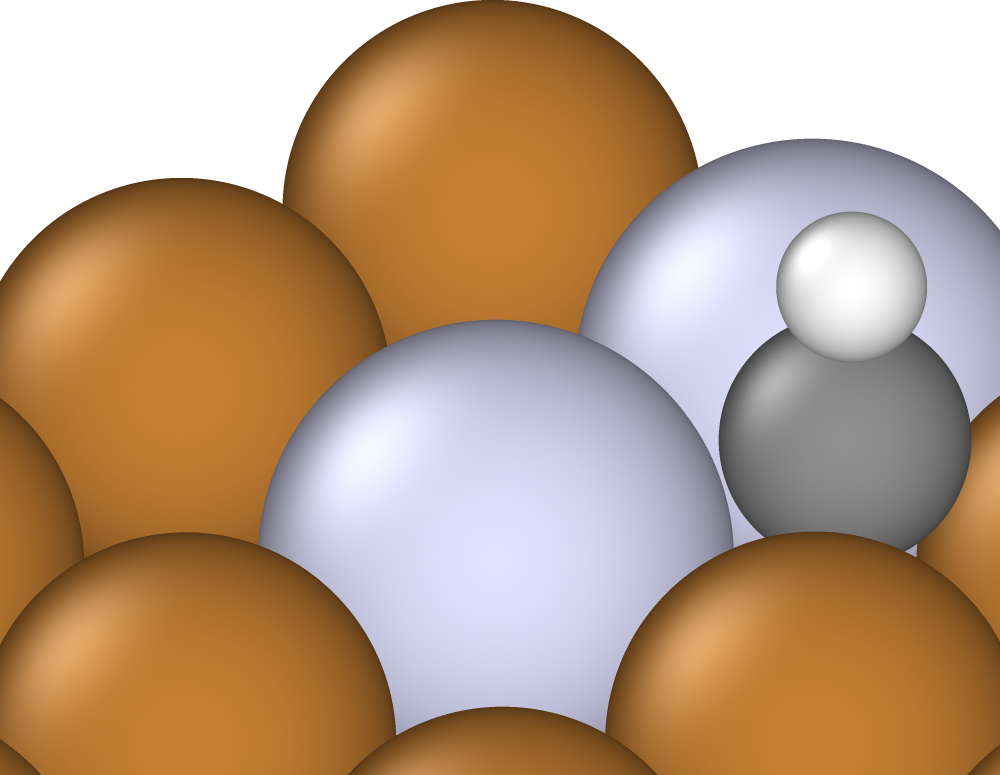}
    \end{subfigure}\hspace{0.001\linewidth}
    \begin{subfigure}[b]{0.161\linewidth}
      \centering
      \includegraphics[width=\linewidth]{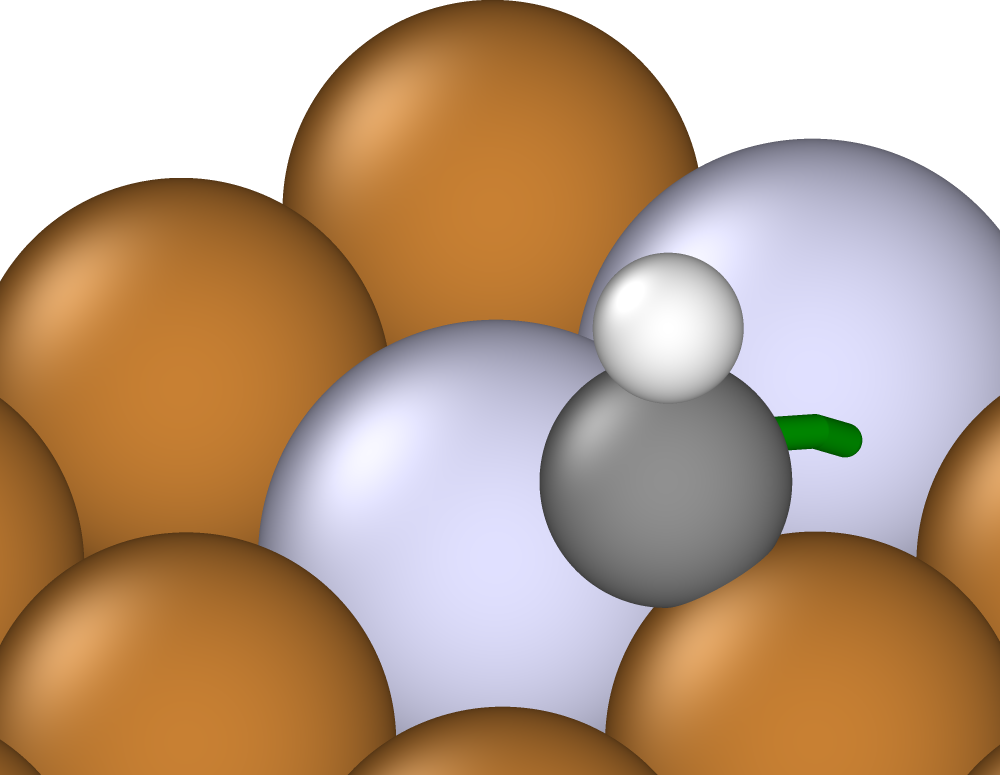}
    \end{subfigure}\hspace{0.001\linewidth}
    \begin{subfigure}[b]{0.161\linewidth}
      \centering
      \includegraphics[width=\linewidth]{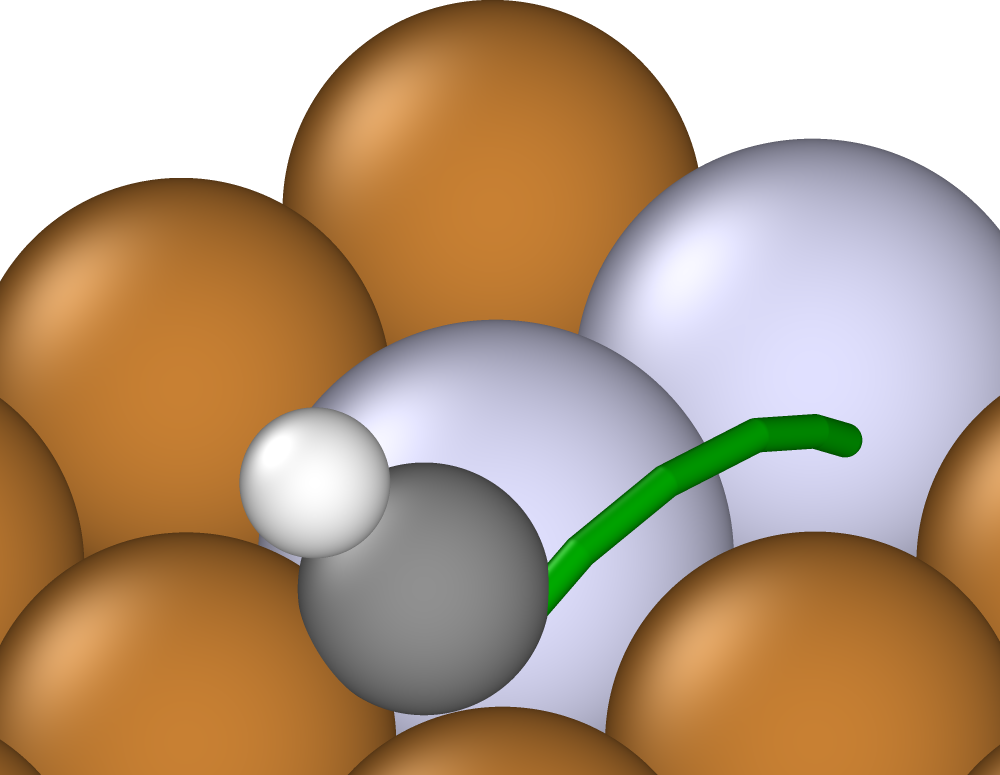}
    \end{subfigure}\hspace{0.001\linewidth}
    \begin{subfigure}[b]{0.161\linewidth}
      \centering
      \includegraphics[width=\linewidth]{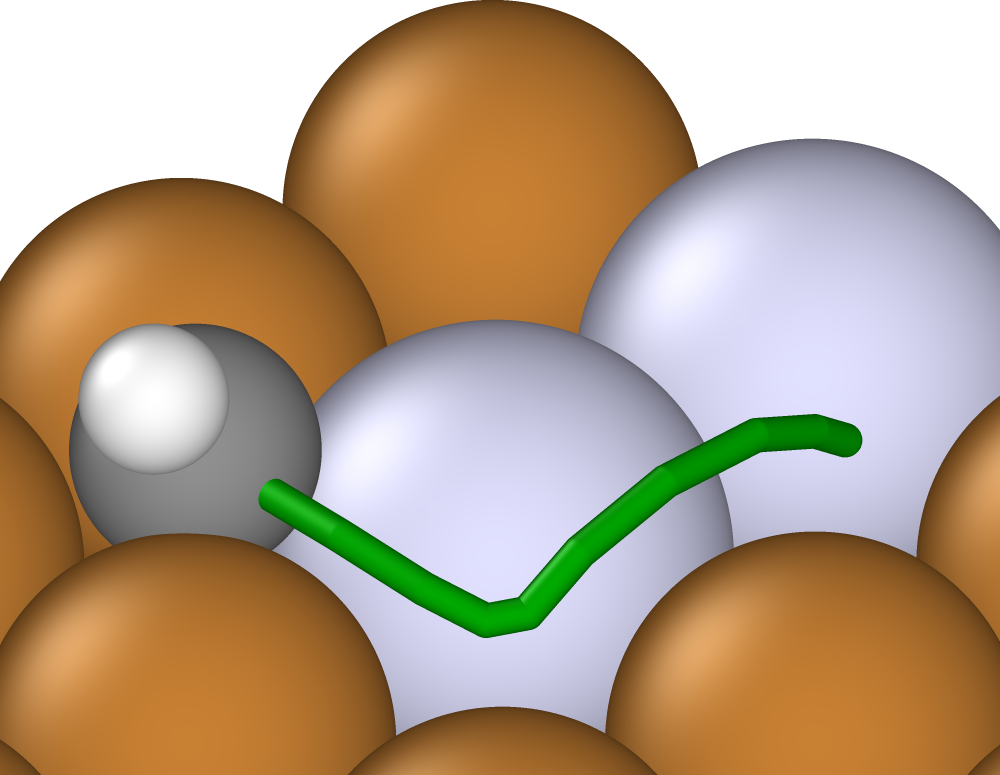}
    \end{subfigure}\hspace{0.001\linewidth}
    \begin{subfigure}[b]{0.161\linewidth}
      \centering
      \includegraphics[width=\linewidth]{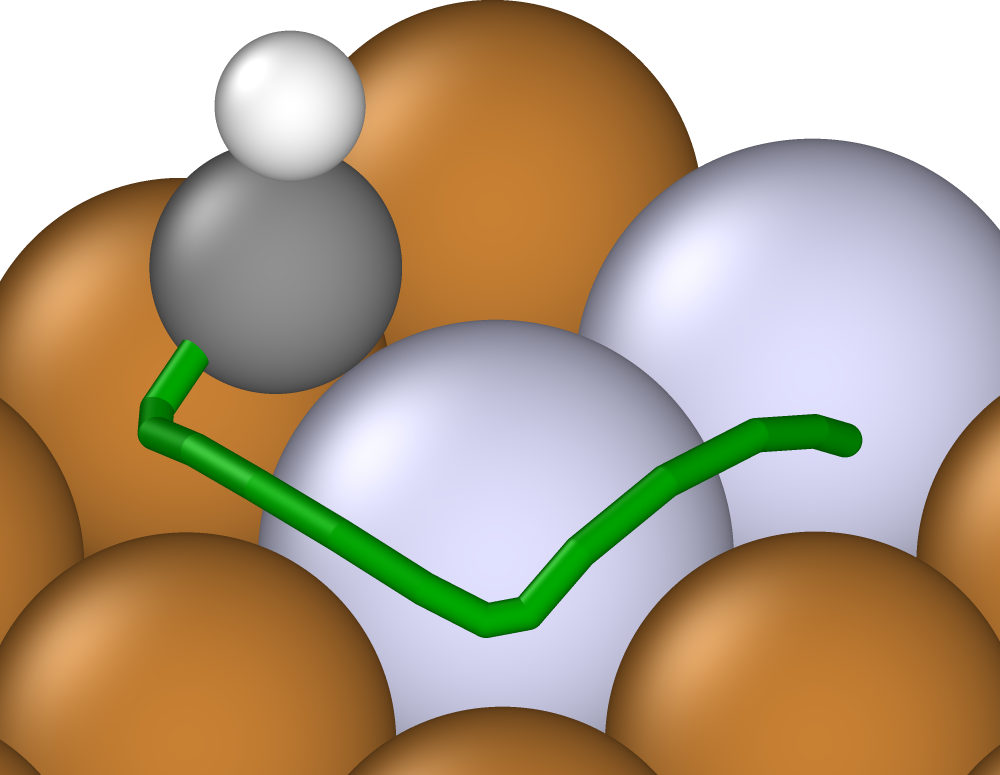}
    \end{subfigure}\hspace{0.001\linewidth}
    \begin{subfigure}[b]{0.161\linewidth}
      \centering
      \includegraphics[width=\linewidth]{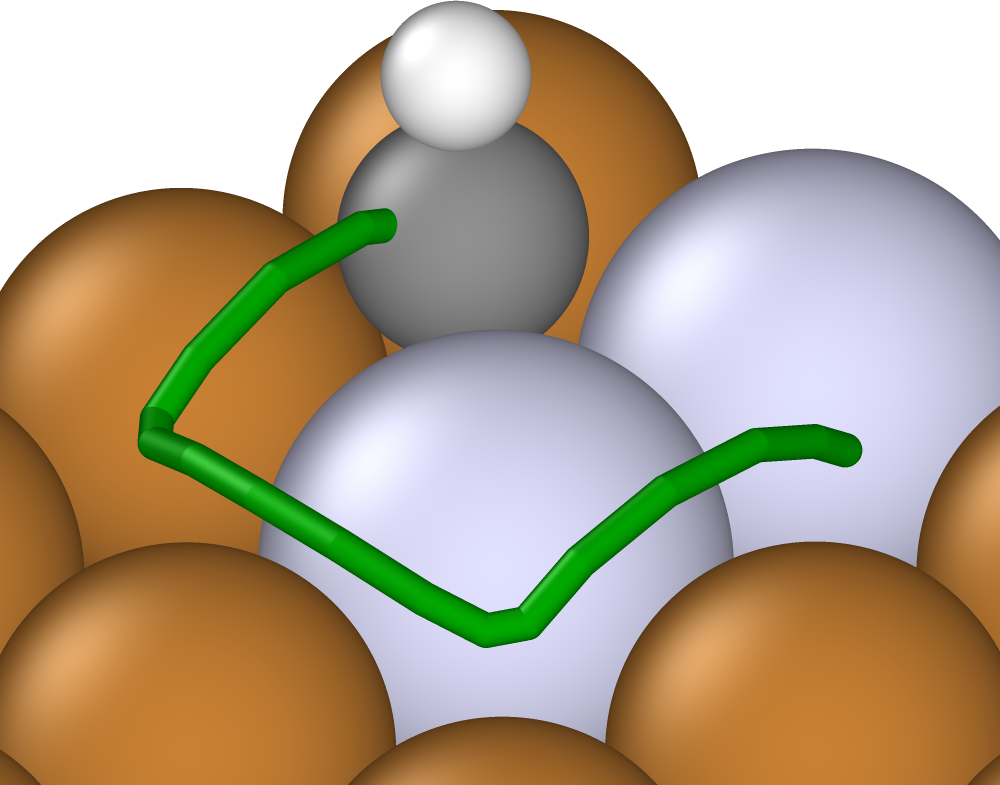}
    \end{subfigure}
    \vspace{-3pt}
    \caption{Iteration $400$}
  \end{subfigure}
  \\[0.5mm]  
  \begin{subfigure}[t]{\linewidth}
    \centering
    \begin{subfigure}[b]{0.161\linewidth}
      \centering
      \includegraphics[width=\linewidth]{figs/3d/sine/ovito/traj500/0.png}
    \end{subfigure}\hspace{0.001\linewidth}
    \begin{subfigure}[b]{0.161\linewidth}
      \centering
      \includegraphics[width=\linewidth]{figs/3d/sine/ovito/traj500/3.png}
    \end{subfigure}\hspace{0.001\linewidth}
    \begin{subfigure}[b]{0.161\linewidth}
      \centering
      \includegraphics[width=\linewidth]{figs/3d/sine/ovito/traj500/7.png}
    \end{subfigure}\hspace{0.001\linewidth}
    \begin{subfigure}[b]{0.161\linewidth}
      \centering
      \includegraphics[width=\linewidth]{figs/3d/sine/ovito/traj500/10.png}
    \end{subfigure}\hspace{0.001\linewidth}
    \begin{subfigure}[b]{0.161\linewidth}
      \centering
      \includegraphics[width=\linewidth]{figs/3d/sine/ovito/traj500/14.png}
    \end{subfigure}\hspace{0.001\linewidth}
    \begin{subfigure}[b]{0.161\linewidth}
      \centering
      \includegraphics[width=\linewidth]{figs/3d/sine/ovito/traj500/17.png}
    \end{subfigure}
    \vspace{-3pt}
    \caption{Iteration $500$}
  \end{subfigure}
  \vspace{-8pt}
  \caption{INR-GS paths at different iterations for the methylidyne diffusion system.}
  \label{fig:sine_evolve}
\end{figure*}

\noindent
Fig.~\ref{fig:sine_evolve} shows the evolution of the INR-GS path for the methylidyne diffusion system when optimized over $500$ iterations.

\section{\ce{N2 + H2} system}
\label{app:sec:n2h2}
We train the MACE potential against energies and forces from Density Functional Theory (DFT) to improve its accuracy on the system. For this, we first construct a dataset of configurations by randomly perturbing points along the interpolated path connecting the initial and final states. The catalyst is not perturbed, and the interpolation is repeated for all permutations of like atoms (\ce{N} and \ce{H} only) in the initial state for a better coverage of the configuration space. Examples containing clashing atoms are filtered out. We then collect reference energies and per-atom forces for all examples using GPU-accelerated DFT calculations implemented in Quantum Espresso\cite{qe}. The self‐consistent field (SCF) simulations use a plane wave basis set with kinetic energy cutoff $600\ \text{eV}$. The Brillouin zone is sampled with a $4\times4\times1$ Monkhorst--Pack grid, and the Perdew--Burke--Ernzerhof (PBE) exchange-correlation functional is adopted. The SCF procedure uses mixing parameter $0.7$ and Davidson diagonalization and is run for a maximum of $300$ iterations at an energy tolerance $1e{-6}$ eV. All other parameters are set to their defaults. We source pseudopotentials from Materials Cloud\footnote{See \url{https://www.materialscloud.org/discover/sssp/table/efficiency} for pseudopotentials.}\textsuperscript{,}\cite{mp}. The final dataset consists of $856$ examples with converged energies and forces. We use these data points to train MACE over $1500$ epochs with distance cutoff $6$ \r{A}, stochastic weight averaging\cite{swa} starting at $1200$ epochs, and exponentially moving average decay $0.99$. All other parameters are set to their defaults.

\section{Path generalization examples}
\label{app:sec:general}

\begin{figure*}
  \centering
  \begin{subfigure}[b]{0.9\linewidth}
    \centering
    \includegraphics[width=\linewidth]{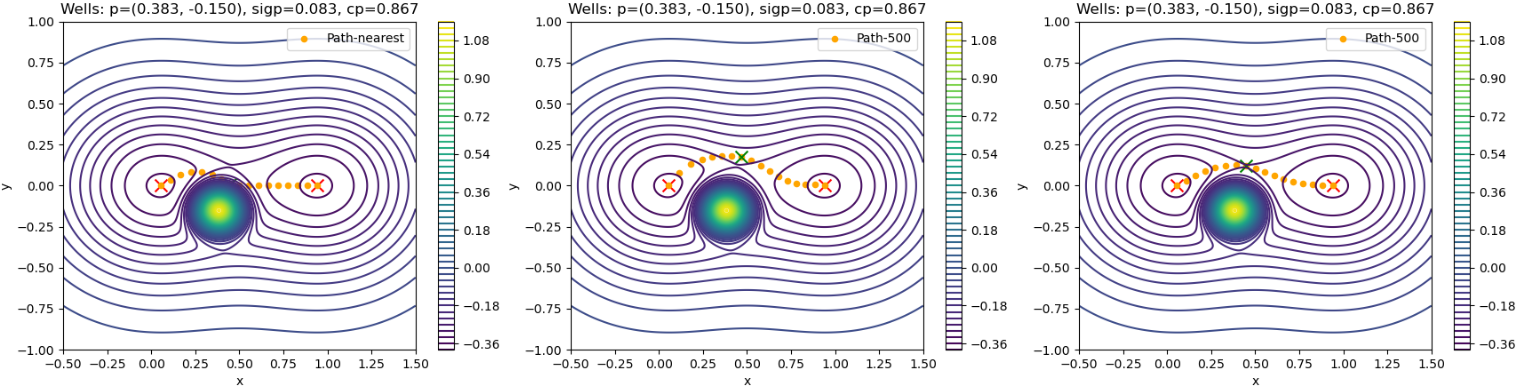}
    \vspace{-15pt}
    \caption{}
  \end{subfigure}
  \\[0.5mm]
  \begin{subfigure}[b]{0.9\linewidth}
    \centering
    \includegraphics[width=\linewidth]{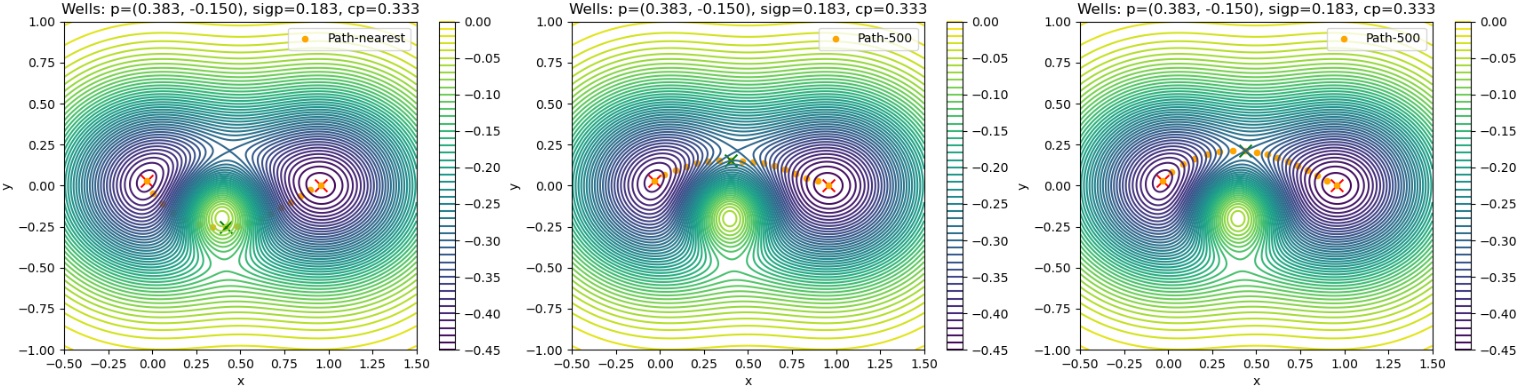}
    \vspace{-15pt}
    \caption{}
  \end{subfigure}
  \\[0.5mm]
  \begin{subfigure}[b]{0.9\linewidth}
    \centering
    \includegraphics[width=\linewidth]{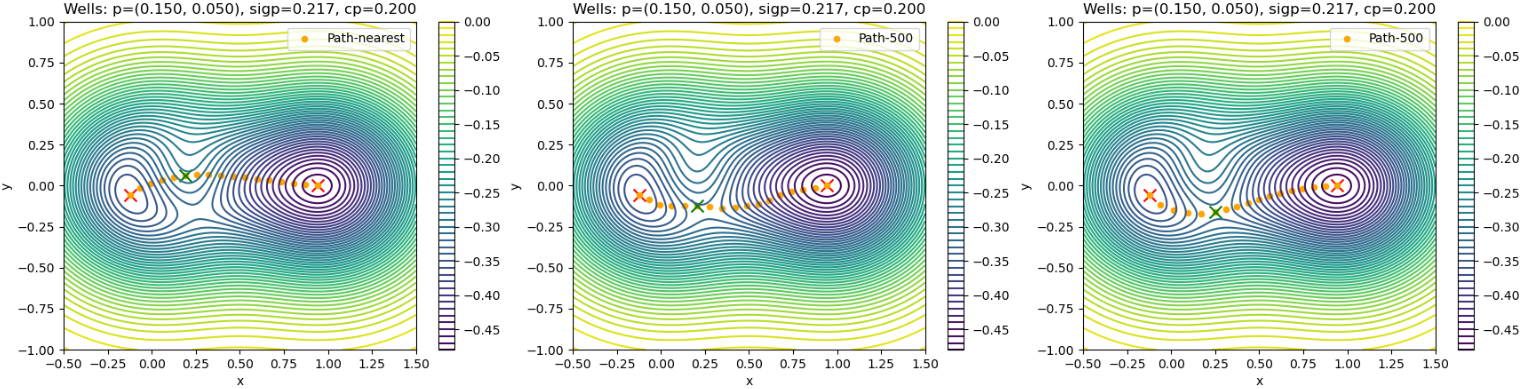}
    \vspace{-15pt}
    \caption{}
  \end{subfigure}
  \\[0.5mm]
  \begin{subfigure}[b]{0.9\linewidth}
    \centering
    \includegraphics[width=\linewidth]{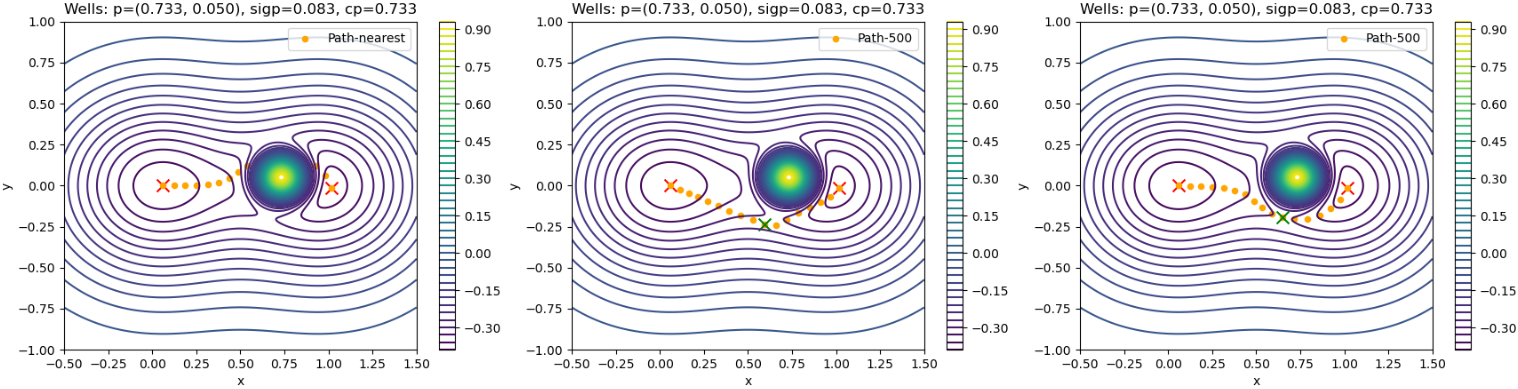}
    \vspace{-15pt}
    \caption{}
  \end{subfigure}
  \\[0.5mm]
  \begin{subfigure}[b]{0.9\linewidth}
    \centering
    \includegraphics[width=\linewidth]{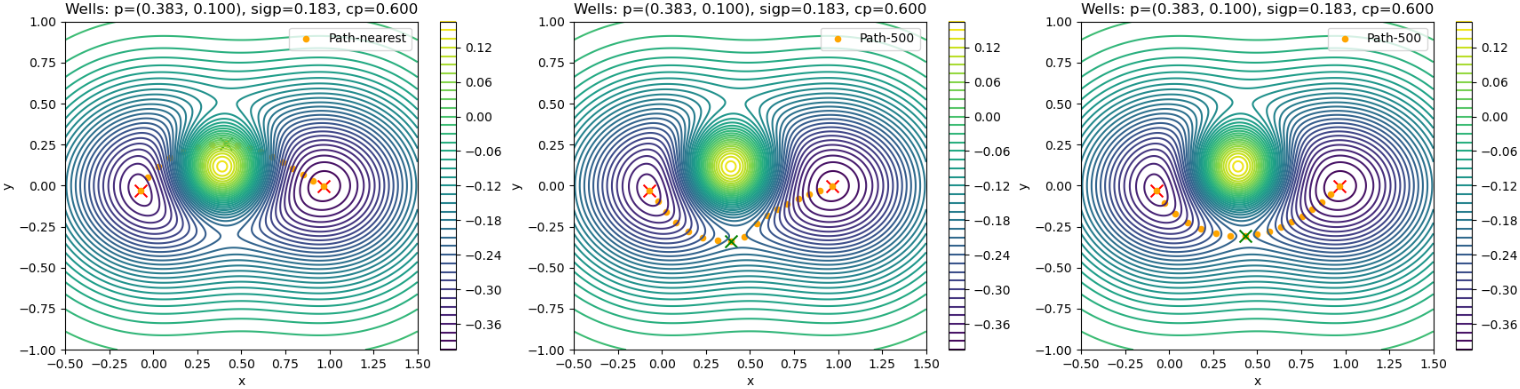}
    \vspace{-15pt}
    \caption{}
  \end{subfigure}
  \\[0.5mm]
  \vspace{-8pt}
  \caption{Nearest neighbor (left), INR (middle), and true (right) paths for five unseen systems (a--e).}
  \label{fig:paths3-7}
\end{figure*}

Fig.~\ref{fig:paths3-7} shows some example predicted paths on unseen systems along with the nearest neighbor and true paths.

\clearpage

\section*{References}

\nocite{*}
\bibliography{aipsamp}

\end{document}